%% file: main.tex
\documentclass[10pt,twocolumn,letterpaper]{article}

\usepackage[pagenumbers]{iccv} 

\usepackage{svg}
\usepackage{multirow}
\usepackage{tabularx}
\usepackage{stackengine}
\usepackage{makecell}
\usepackage{cellspace}
\usepackage{algorithm}
\usepackage{algpseudocode}
\usepackage{afterpage}
\input{preamble}

\definecolor{iccvblue}{rgb}{0.21,0.49,0.74}
\usepackage[pagebackref,breaklinks,colorlinks,allcolors=iccvblue]{hyperref}

\def\paperID{3} 
\def\confName{ICCV AIGENS}
\def\confYear{2025}

\title{LSSGen: Leveraging Latent Space Scaling in Flow and Diffusion for Efficient Text to Image Generation}

\author{
  Jyun-Ze Tang\textsuperscript{1}\quad
  Chih-Fan Hsu\textsuperscript{1}\quad
  Jeng-Lin Li\textsuperscript{1}\quad
  Ming-Ching Chang\textsuperscript{2}\quad
  Wei-Chao Chen\textsuperscript{1}
  \and 
  \textsuperscript{1}Inventec Corporation, Taipei, Taiwan
  \and
  \textsuperscript{2}University at Albany, State University of New York, NY, USA
  \and
  {\tt\small \textsuperscript{1}\{tang.nickct, hsu.chih-fan, li.johncl, chen.wei-chao\}@inventec.com, \textsuperscript{2}mchang2@albany.edu}
}

\begin{document}
\twocolumn[
  {
    \maketitle
    \begin{center}
    \centering
    \includegraphics[width=\textwidth]{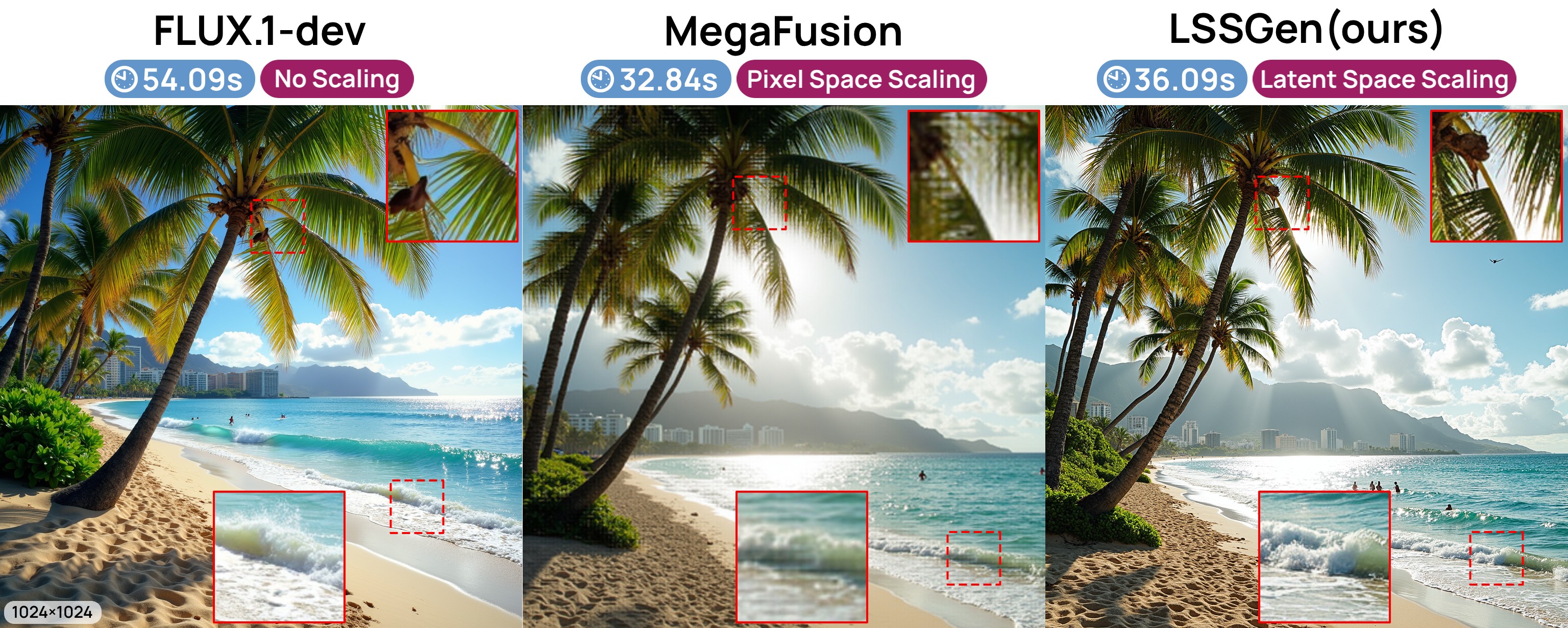}
    \captionsetup{hypcap=false}\vspace{-6mm}
    \captionof{figure}{Comparison of image synthesis results at $1024 \times 1024$ resolution: Our proposed method, LSSGen, leverages latent space scaling to achieve SOTA speed-quality optimality. 
    LSSGen generates significantly higher-quality images than MegaFusion~\cite{wu2024megafusion} and provides much faster inference than FLUX.1-dev~\cite{blackforest2024flux}.
    }
    \label{fig:teaser}
\end{center}
  }
]

\input{sec/0_abstract}    
\input{sec/1_intro}

\input{sec/2_related}
\input{sec/3_preliminary}
\input{sec/4_method}

\input{sec/5_experiments}
\input{sec/6_conclusion}
{
    \small
    \bibliographystyle{ieeenat_fullname}
    \bibliography{main}
}

%
%

\newpage
\appendix
\input{sec/_supplementary}

\end{document}

%% file: preamble.tex
\usepackage[acronym]{glossaries}
\newacronymstyle{long-short-paren}  
{%
  \GlsUseAcrEntryDispStyle{long-short}%
}
{%
  \GlsUseAcrStyleDefs{long-short}%
}
\setacronymstyle{long-short-paren}

\input{Utility/parameter}

\usepackage{enumitem}

%% file: Utility/parameter.tex
\newacronym{DynaFlow}{LSSGen}{Latent Space Scaling Generation}
\newcommand\DynaFlow{\gls*{DynaFlow}}
\newcommand\LSSGen{\gls*{DynaFlow}}
\newacronym{LSSGen}{LSSGen}{Latent Space Scaling Generation}
\newacronym{VAR}{VAR}{Visual AutoRegressive}

\newacronym{LDM}{LDM}{Latent Diffusion Model}
\newcommand\LDM{\gls*{LDM}}
\newacronym{DiT}{DiT}{Diffusion Transformer}
\newcommand\DiT{\gls*{DiT}}
\newacronym{CNN}{CNN}{Convolutional Neural Network}

\newacronym{RF}{RF}{Rectified Flow}
\newcommand\RF{\gls*{RF}}
\newacronym{FM}{FM}{Flow Matching}
\newcommand\FM{\gls*{FM}}
\newacronym{DM}{DM}{Diffusion Model}
\newcommand\DM{\gls*{DM}}
\newacronym{CLIP}{CLIP}{Contrastive Language-Image Pretraining}

\newcommand\InitSigma{$\sigma_\text{init}$}


%% file: sec/0_abstract.tex
\begin{abstract}
Flow matching and diffusion models have shown impressive results in text-to-image generation, producing photorealistic images through an iterative denoising process. A common strategy to speed up synthesis is to perform early denoising at lower resolutions. However, traditional methods that downscale and upscale in pixel space often introduce artifacts and distortions. These issues arise when the upscaled images are re-encoded into the latent space, leading to degraded final image quality.
To address this, we propose {\bf Latent Space Scaling Generation (LSSGen)}, a framework that performs resolution scaling directly in the latent space using a lightweight latent upsampler. Without altering the Transformer or U-Net architecture, LSSGen improves both efficiency and visual quality while supporting flexible multi-resolution generation. Our comprehensive evaluation covering text-image alignment and perceptual quality shows that LSSGen significantly outperforms conventional scaling approaches. When generating $1024^2$ images at similar speeds, it achieves up to 246\% TOPIQ score improvement.
\end{abstract}

%% file: sec/1_intro.tex
\section{Introduction}
\label{sec:intro}


Recent advances in {\FM} and {\DM} have revolutionized text-to-image synthesis, achieving unprecedented image quality. Despite their differing mathematical foundations, both are implemented as continuous denoising processes based on probabilistic path learning~\cite{domingo2024adjoint}. However, this iterative nature leads to slow generation~\cite{salimansprogressive}, and the computational cost grows quadratically with image resolution~\cite{FMBoost}.

A common approach to this challenge is a progressive, coarse-to-fine generation pipeline~\cite{wu2024megafusion, kim2025diffusehigh, du2024demofusion}, which generates a base-resolution image and then upscales it beyond the training resolution in pixel space. While this helps preserve semantic structure at higher resolutions, pixel-space upscaling often introduces artifacts like blurriness (Fig.~\ref{fig:teaser}). These artifacts arise because re-encoding the upscaled image back into the latent space distorts the underlying latent features. Without a clear understanding of these distortions, it remains difficult to resolve the speed-quality trade-off in FM and DM models.

Both {\DM} and {\FM} generate images by gradually transforming Gaussian noise into a clear image. This denoising process can be viewed as predicting a velocity field that moves the sample from a blurry state toward a sharp one. From an information-theoretic perspective, this progression aligns with a process that first captures low-frequency components and then incrementally adds high-frequency details~\cite{qian2024boosting,falck2025fourier,wu2024freediff}. We hypothesize that this natural blurry-to-sharp progression can be efficiently approximated by a {\em resolution-based} approach: starting with a low-resolution image that captures the scene’s structure, and progressively increasing the resolution to introduce fine details and improve visual clarity.

Building on this principle, we propose the {\bf \LSSGen} framework, which replaces the early stages of image generation with lower-resolution processing. Rather than scaling images in pixel space, LSSGen progressively upsamples features directly in the latent space using a dedicated latent upsampler. This approach avoids the artifacts commonly introduced by pixel-space resolution changes. To further address the shift in noise characteristics during scaling, we propose a novel {\em noise compensation} and {\em rescheduling} strategy. This ensures consistency between noise and data across stages and dynamically adjusts the denoising process for improved stability and image quality.

Notably, our latent upsampler is VAE-dependent, making it reusable across different flow and diffusion models that share the same VAE. This architectural decoupling allows the upsampler to work purely on VAE-derived latent features, making the upsampler agnostic to the underlying generative architecture ({\em e.g.}, U-Net or Transformer). It provides a versatile, {\em train-once}, {\em use-across-models} solution.

Extensive experiments show that {\LSSGen} achieves a $1.5 \times$ speedup for $1024^2$ image generation while maintaining comparable quality across four standard evaluation metrics: GenEval~{\cite{ghosh2023geneval}}, CLIP-IQA~\cite{clipiqa:wang2023exploring}, TOPIQ~\cite{chen2024topiq}, and NIQE~\cite{niqe:mittal2012making}. At $2048^2$ resolution, the benefits are even greater due to the quadratic increase in computational cost with image size. These results establish  LSSGen as a practical and scalable solution for efficient high-resolution image synthesis. Additionally, our experiments show that LSSGen offers superior global semantic preservation compared to previous methods, particularly when generating images beyond the original training resolution.

Our main contributions are summarized as follows:
\begin{itemize}[leftmargin=10pt] \itemsep -.1em

\item We introduce {\LSSGen}, a latent space scaling framework applicable to a wide range of generative models. It includes a novel latent upsampling method and a noise compensation and rescheduling strategy.

\item We demonstrate through extensive experiments that {\LSSGen} achieves a strong balance between computational efficiency and image quality, outperforming baseline methods in both areas.

\item We present a detailed analysis comparing pixel-space and latent-space transformations, offering fresh insights into the dynamics of multi-resolution image generation.

\end{itemize}

%% file: sec/2_related.tex
\section{Related Work}
\label{sec:related}


Acceleration methods for {\DM} and {\FM} generally fall into two main categories: architecture optimization and model distillation.

\medskip
\noindent
{\bf Architecture optimization:}
Dynamic transformer~\cite{dydit:zhao2024dynamic} addresses computation redundancy across timesteps by introducing a Timestep-wise Dynamic Width strategy, which reduces computation by adaptively adjusting the model's width during generation.
SANA~\cite{xie2024sana} employs a deep compression autoencoder to map images into a smaller feature domain, and replaces standard attention in {\DiT}~\cite{dit:peebles2023scalable} with linear attention for efficiency.
CLEAR~\cite{liu2024clear} introduce a convolution-like mechanism to reduce the complexity of the attention module. These approaches improve inference efficiency by optimizing the attention mechanisms in Transformer architectures~\cite{vaswani2017attention}. However, they typically require training from scratch or additional fine-tuning to accommodate architectural changes.

\medskip
\noindent
{\bf Model Distillation:}
An alternative approach to acceleration is reducing the number of inference timesteps. Salimans and Ho~\cite{salimansprogressive} propose progressive distillation by merging timesteps, while Luo et al.~\cite{lcm:luo2023latent} introduce consistency models that directly map noisy inputs to clean samples. Although these methods achieve significant speedups, they often compromise high-resolution fidelity by limiting the gradual error correction offered by iterative denoising. Additionally, distillation methods typically require extensive computational resources and complex training procedures. In contrast, our framework employs progressive resolution upscaling to reduce computational cost while maintaining image quality. It remains compatible with existing distillation techniques, offering complementary acceleration and quality gains when combined.




\medskip
\noindent
{\bf Latent Space Modeling:}
Direct sampling in pixel space poses significant computational challenges, especially for high-resolution image generation.  To address this, Rombach et al.~\cite{ldm:rombach2022high} propose {\LDM}, which perform denoising in the compressed latent space of a pre-trained autoencoder~\cite{kingma2013auto,rezende2014stochastic,esser2021taming}. This approach significantly reduces computational costs by leveraging dimensionality reduction. Latent-space techniques have also been adopted in flow matching~\cite{dao2023flow}, highlighting the general applicability of this paradigm. Recent work by Esser et al.~\cite{sd3:esser2024scaling} further demonstrates the scalability of {\RF} by integrating it into the {\LDM} framework. However, a key challenge remains: the mismatch between pixel-space and latent-space representations complicates operations such as interpolation and scaling~\cite{park2023understanding, guo2024smooth}, making latent manipulation less intuitive and harder to control.

\medskip
\noindent
{\bf Progressive Upscaling:}
Cascaded Diffusion~\cite{cascade:ho2022cascaded} adopts a multi-resolution generation strategy to reduce the cost of high-resolution synthesis but relies on training resolution-specific super-resolution models. Self-Cascade Diffusion~\cite{selfcascade:guo2024make} integrates ResNet-based plugins into UNet architectures, conditioned on low-resolution latent features. While it improves generation quality, it does not reduce computational cost and requires training resolution-specific components, adding memory overhead and limiting generalizability across different architectures.
MegaFusion~\cite{wu2024megafusion} and DiffuseHigh~\cite{kim2025diffusehigh} introduce progressive upsampling pipelines via pixel-space scaling and refinement.
FMBoost~\cite{FMBoost} accelerates high-resolution generation by upscaling small diffusion outputs in pixel space before applying a flow model to refine blurry regions.
I-Max~{\cite{du2024imax}} modifies both the attention mechanism and inference pipeline of FLUX.1-dev~\cite{blackforest2024flux} to support 4K image synthesis.
However, these methods share two key limitations:
(1) The encode-decode cycle often leads to detail degradation~\cite{avae:zhu2023designing}, and
(2) Resolution scaling in pixel space introduces systematic bias into the latent space, resulting in persistent blur artifacts, even when additional noise is injected.

Our method progressively upsamples features in the latent space and uses the original diffusion or flow model to refine the final image. The upsampler is a VAE-dependent ResNet that can be reused across different generative models. This design reduces computational cost and allows for easy integration with various diffusion and flow-based architectures.

\begin{figure*}[t]
\centerline{
  \includegraphics[width=\linewidth]{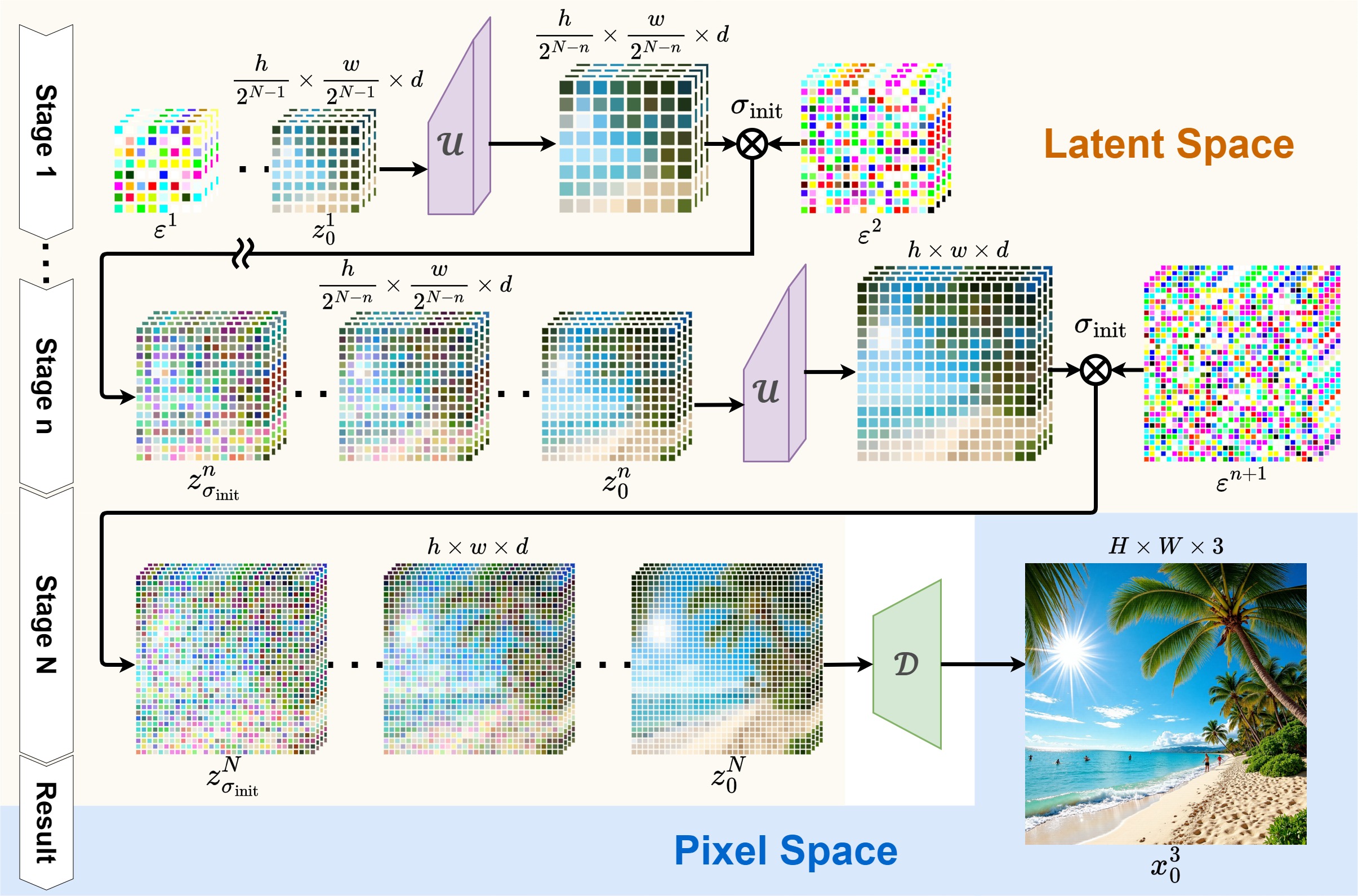}
}
\caption{Overview of our Latent Space Scaling Generation (LSSGen) framework. The inference process is divided into multiple stages, implementing a progressive upscaling methodology. This approach utilizes latent space upsampling with a pretrained upsampler $\mathcal{U}$ and incorporates a noise compensation strategy with noise timestep initialization $\sigma_\text{init}$ at each stage (1 $\to$ N). Variables $h$ and $w$ correspond to latent image dimensions, $d$ represents latent channel count, while $H$ and $W$ denote pixel space image dimensions.}
\label{fig:flow_chart}	
\end{figure*}

%% file: sec/3_preliminary.tex
\section{Preliminary}
\label{sec:Preliminaries}

\noindent
{\bf Diffusion Models (DM)} generate images by iteratively denoising Gaussian noise into a target image. Let $x_t$ denote noisy image at timestep $t$, $\epsilon \sim \mathcal{N}(0,1)$ denote Gaussian noise,  $\beta_t \in (0,1)$ denote a predefined noise schedule controlling noise variance at time $t$, and $\bar{\alpha}$ denote the cumulative signal retention. The forward diffusion process adds noise to the input image $x_0$ over $t$ timesteps, defined as:
\begin{equation}
\label{eq:diffusion_function}
x_t = \sqrt{\bar{\alpha}_t}x_0 + \sqrt{1-\bar{\alpha}_t}\epsilon, \; \text{where} \; \bar{\alpha}_t = \prod_{i=1}^{t}(1-\beta_i).
\end{equation}
This closed-form allows direct sampling at timestep $t$, improving efficiency without simulating the full Markov chain.


During training, the forward process $q(x_{1:T}|x_0)$ corrupts the clean image $x_0$ using Gaussian noise:
$q(x_t|x_{0}) := \mathcal{N}(x_t; \sqrt{\bar{\alpha}_t}x_0, (1-\bar{\alpha})\mathbf{I})$.
The reverse process, parameterized by a neural network with model parameters $\theta$, learns to reconstruct the image through denoising:
%
$p_\theta(x_{t-1}|x_t) := \mathcal{N}(x_{t-1}; \mu_\theta(x_t, t), \Sigma_\theta(x_t, t))$,
%
where $\mu_\theta$ and $\Sigma_\theta$ are the predicted mean and variance. The model is trained to minimize the difference between the predicted noise $\epsilon_\theta$ and the true noise $\epsilon$ via the pixel-space loss:
\begin{equation}
\label{eq:diffusion_loss}
\mathcal{L}_\text{pixel} := \mathbb{E}_{x,\epsilon\sim\mathcal{N}(0,1),t}\left[\|\epsilon - \epsilon_\theta(x_t,t)\|_2^2\right].
\end{equation}
This formulation, introduced by Ho et al.~\cite{ddpm:ho2020denoising} and extended by Song et al.~\cite{ddim:songdenoising}, enables high-quality image generation through iterative refinement, with denoising accuracy improving at each step.

\medskip
\noindent
{\bf Flow Matching (FM)}~{\cite{chen2018neural}} formulates generation as a linear interpolation between data $x$ and Gaussian noise $\epsilon$: 
\begin{equation}
x_t := (1-t)x + t \epsilon, 
\label{eq:flow_function}
\end{equation}
where $t \in [0, 1]$ represents the interpolation time.

\medskip
\noindent
\textbf{Rectified Flow (RF)}~{\cite{lipmanflow, liuflow}} further specifies the flow trajectory using a simple velocity field $v_\theta(x_t, t) = \epsilon - x_0$. This is modeled by a neural network parameterized by $\theta$, and trained using the \textit{velocity matching objective}:
\begin{equation}
\label{eq:flow_loss}
\mathcal{L}_\text{flow} := \mathbb{E}_{x,\epsilon\sim\mathcal{N}(0,1),t}\left[\|v_\theta(x_t, t) - (x_1 - x_0)\|_2^2\right].
\end{equation}
At inference, image samples are generated by solving the learned \textit{flow ODE}, which constructs images progressively along the linear path defined by $x_t$. This approach offers efficient image synthesis by learning a direct transport path from noise to data, avoiding the complexity of iterative score-matching. However, accurately modeling this path remains challenging, especially when targeting complex, high-resolution distributions that require fine-grained detail and structure.

%% file: sec/4_method.tex

\section{Methodology}
\label{sec:method}

To describe our proposed {\LSSGen} framework, we adopt the unified formulation of diffusion and flow models introduced by Patel et al.~\cite{patel2024exploring}. In this setting, the generative process is defined by a parameterized ground-truth Markov generator $F_t(x)$,  with the training objective: 
\begin{equation}
\mathcal{L}_\theta=\mathbb{E}_{t,x}[D_{X_t}(F_t(x), F_t^\theta(x))],
\end{equation}
where $F_t^\theta(x)$ is the neural network with learning parameter $\theta$, and $D$ denotes the Bregman Divergence. The loss functions of both diffusion and flow models, described in Eqs.~\eqref{eq:diffusion_loss} and \eqref{eq:flow_loss}, can be expressed under this unified objective. Similarly, the sampling processes in Eq.~\eqref{eq:diffusion_function} and \eqref{eq:flow_function} can be unified by setting $\sqrt{\bar{\alpha}_t} = 1-t$~\cite{song2024unraveling}. In practice, most diffusion and flow models encode pixel space data $x$ into a latent representation $z$ during training and perform inference in the latent space. Thus, we re-express the generative trajectory in terms of $z$ as:
\begin{equation}
\label{eq:unified_function}
z_t = (1 - \sigma_t) z_0 + \sigma_t \epsilon,
\end{equation}
where $\sigma_t$ controls the noise intensity over time.


Our proposed {\LSSGen} framework builds on this foundation by introducing three key components:
(1) a probabilistic perspective for resolution-autoregressive scaling ($\S$\ref{subsec:method:resolutionAR}),
(2) a latent space upsampler for resolution transitions ($\S$\ref{subsec:method:scaling}), and
(3) a timestep rescheduling mechanism to accelerate generation ($\S$\ref{subsec:method:shifting}).  



\subsection{Resolution-Autoregressive Scaling}
\label{subsec:method:resolutionAR}

From an information-theoretic perspective, the generative path from blurry to sharp images mirrors a process that reconstructs low-frequency components first, followed by high-frequency details~\cite{qian2024boosting,falck2025fourier,wu2024freediff}. Our goal in high-resolution image synthesis follows this principle: preserve low-frequency structure and progressively add high-frequency content to enhance detail and realism.

In Eq.~\eqref{eq:unified_function}, the noise coefficient $\sigma_t$ controls the trade-off between signal and noise ratio (SNR)~\cite{kingma2021variational}:
\begin{equation}
SNR(t)=\frac{(1-\sigma_t)^2}{\sigma_t^2},
\end{equation}
which decreases over time as noise increases. Under the variance-preserving formulation~\cite{hang2024improved}, $\sigma_t$ is recovered as:
\begin{equation}
\sigma_t=\frac{1}{1+\sqrt{SNR(t)}}.
\end{equation}
 
While diffusion and flow models usually define noise schedules using $\bar{\alpha}_t$, we reinterpret upsampling as a transformation of $SNR$. For example, scaling the resolution by a factor of $s=2$ (doubling both width and height) increases the pixel count by $4 \times$, which we approximate as reducing $SNR(t)$ to $\frac{1}{4}SNR(t)$. This aligns with the view of bilinear interpolation as a smoothing process that effectively adds noise.
To reflect this in the generative path, we scale the noise coefficient after upsampling as $\sigma'_t = \frac{3}{4}\sigma_t$, providing a more suitable initialization for denoising and better aligning with the underlying probabilistic trajectory.

The resolution scaling process is divided into $N$ stages to better preserve image quality during upsampling.
At each stage $n\in \{1, ..., N\}$, the resolution is initialized as $\frac{h}{2^{(N-1)}} \times \frac{w}{2^{(N-1)}}$ (see Fig.~\ref{fig:flow_chart}). At this scale, the signal-to-noise ratio is approximately $\frac{1}{4}SNR$, corresponding to a noise coefficient of $\sigma_\text{init} \approx 0.75$, assuming a linear relationship between $SNR$ and $\sigma$.   
We adapt the probabilistic trajectory in Eq.~\eqref{eq:unified_function} at each stage as:
\begin{equation}
z_{\sigma_{\text{init}}}^{n} \approx (1-\sigma_\text{init}) \mathcal{U}(z_0^{n-1}) + \sigma_\text{init} \epsilon^{n},
\label{eq:relation_sigma_resolution}
\end{equation}
where $\mathcal{U}(z)$ is the latent upsampler from $\S$\ref{subsec:method:scaling}, and $\epsilon^n \sim {\cal N}(0,1)$ is Gaussian noise added at stage $n$.

This process iterates across $N$ stages, with each step refining the previous output. The final high-resolution latent $z_0^N$ is obtained by denoising the upsampled signal $\mathcal{U}(z_0^{N-1})$. 
Our progressive scaling approach avoids the cost of direct high-resolution generation at every stage. 
Since self-attention scales quadratically with image size, {\em i.e.}, $O\left(\left(H \times W\right)^2\right)$, starting generation at low resolution and gradually scaling up provides substantial computational savings. These savings are proportional to the square of the scaling factor. A detailed analysis is provided in the supplementary material.

\begin{figure}[t]
\centerline{
  \includegraphics[width=1.1\linewidth]{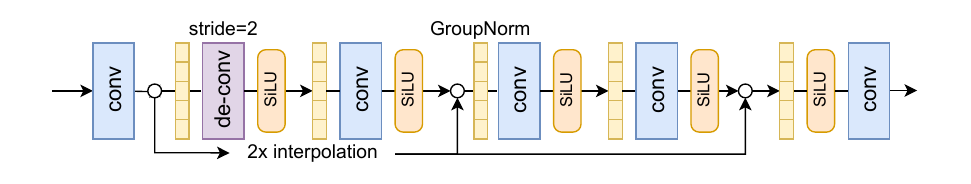}
\vspace{-3mm}
}
\caption{Upsampler architecture.}
\label{fig:upsampler_architecture}
\vspace{-5mm}
\end{figure}

\subsection{Latent Space Upsampler}
\label{subsec:method:scaling}

We use a lightweight ResNet Upsampler $\mathcal{U}$ (Fig.~\ref{fig:upsampler_architecture}), with roughly 500K parameters, trained for 3 epochs on the COCO dataset\cite{lin2014coco} using a frozen, pre-trained VAE. The design supports broad compatibility with both diffusion and flow models, based on two key observations:
(1) Many latent diffusion models share the same VAE, and (2) latent transformations should ideally be independent of the specific generative backbone (U-Net or {\DiT}). This architectural decoupling allows the upsampler to operate directly on VAE-derived latent features, making it easily reusable. Since our method consistently applies $2 \times$ resolution scaling, we use ResNet~\cite{resnet:he2016deep} blocks with stride-based deconvolution~\cite{zeiler2010deconvolutional}, for efficient and effective upsampling. This design ensures the upsampler can be applied across a wide range of VAE-based generative models.

\subsection{Timestep Schedule Shifting}
\label{subsec:method:shifting}

To improve computational efficiency, we adopt a timestep shifting strategy inspired by Stable Diffusion 3~\cite{sd3:esser2024scaling}, which emphasizes the importance of early {\RF} steps in generating high-resolution images. In our multi-stage pipeline, more denoising steps are allocated to the early, low-resolution stages where computation is cheaper. Given two resolutions, with the higher resolution $m$ at timestep $t_m$ and a lower resolution $n$ at timestep $t_n$, we apply the following shift to align the schedules:
\begin{equation}
    t_m = \frac{\sqrt{\frac{m}{n}}t_n}{1+(\sqrt{\frac{m}{n}}-1)t_n}.
\end{equation}
This mapping increases the number of steps when spatial resolution is low, leading to a significant reduction in overall computation. The shift factor $\sqrt{\frac{m}{n}}$ also updates the signal-to-noise ratio, allowing the noise coefficient $\sigma$ to remain consistent with the change in resolution.


%% file: sec/5_experiments.tex
\begin{table*}[htp]
    \centering
    \caption{Quality and efficiency comparison on $1024^2$ and $2048^2$ resolutions with various generative architectures. The gray rows represent the baseline methods without a scaling strategy.} \vspace{-2mm}
\resizebox{1.0\linewidth}{!}{
    \setlength{\tabcolsep}{3pt}
    \begin{tabular}{c|c|c|c|c|c|c|c|c|c|c}
        \hline
        Model & Scaling &  Resolution &Stage Steps& Time/img{$\downarrow$} & Speed{$\uparrow$} & TFLOPs{$\downarrow$} & GenEval{$\uparrow$} & CLIP-IQA{$\uparrow$} & TOPIQ{$\uparrow$} & NIQE{$\downarrow$} \\  \hline
        \textcolor{gray}{FLUX.1-dev~\cite{blackforest2024flux}} & - & \multirow{3}{*}{1024} & 50 &\textcolor{gray}{54.38} & \textcolor{gray}{1x} & \textcolor{gray}{2,991} & \textcolor{gray}{0.673} & \textcolor{gray}{0.887} & \textcolor{gray}{0.674} & \textcolor{gray}{5.622} \\ 
        FLUX.1-dev-MegaFusion{\cite{wu2024megafusion}} & Pixel & & 30, 20 &\textbf{33.57} & \textbf{1.6x} & \textbf{1,825} & 0.649 & 0.747 & 0.402 & 6.543 \\ 
        LSS-FLUX.1-dev (ours) & Latent &  & 25, 25 &35.79 & 1.5x & 1,999 & \textbf{0.653} & \textbf{0.914} & \textbf{0.705} & \textbf{5.136} \\ \hline 
        \textcolor{gray}{SD3.5-m~\cite{sd3:esser2024scaling}} & - & \multirow{3}{*}{1024}& 40 &\textcolor{gray}{14.73} & \textcolor{gray}{1x} & \textcolor{gray}{576} & \textcolor{gray}{0.676} & \textcolor{gray}{0.862} & \textcolor{gray}{0.635} & \textcolor{gray}{4.825} \\ 
        SD3.5-m-MegaFusion{\cite{wu2024megafusion}} & Pixel &  & 24, 16 &\textbf{9.28} & \textbf{1.6x} & \textbf{364} & 0.667 & 0.760 & 0.444 & 6.294 \\ 
        LSS-SD3.5-m (ours) & Latent &  & 20, 21 &9.97 & 1.5x & 393 & \textbf{0.672} & \textbf{0.880} & \textbf{0.641} & \textbf{4.705} \\ \hline 
        \textcolor{gray}{Playgroundv2.5~\cite{li2024playground}} & - &  \multirow{2}{*}{1024}& 50 &\textcolor{gray}{9.98} & \textcolor{gray}{1x} & \textcolor{gray}{608} & \textcolor{gray}{0.564} & \textcolor{gray}{0.940} & \textcolor{gray}{0.661} & \textcolor{gray}{3.881} \\ 
        LSS-Playgroundv2.5 (ours) & Latent &  & 25, 37 &9.11 & 1.1x & \textbf{529} & \textbf{0.559} & \textbf{0.942} & \textbf{0.665} & \textbf{3.964} \\ \hline 
        \textcolor{gray}{SD1.5~\cite{ldm:rombach2022high}} & - &  \multirow{3}{*}{1024}& 50 &\textcolor{gray}{9.21} & \textcolor{gray}{1x} & \textcolor{gray}{285} & \textcolor{gray}{0.416} & \textcolor{gray}{0.863} & \textcolor{gray}{0.628} & \textcolor{gray}{3.707} \\ 
        SD1.5-MegaFusion++{\cite{wu2024megafusion}} & Pixel &  & 40, 5, 5 &\textbf{4.09} & \textbf{2.3x} & \textbf{139} & 0.384 & 0.438 & 0.248 & 6.871 \\ 
        LSS-SD1.5 (ours) & Latent &  & 50, 37 &8.82 & 1x & 284 & \textbf{0.451} & \textbf{0.871} & \textbf{0.609} & \textbf{4.667} \\ \hline 
        \textcolor{gray}{FLUX.1-schnell~\cite{blackforest2024flux}} & - & \multirow{2}{*}{1024}& 4 &\textcolor{gray}{4.77} & \textcolor{gray}{1x} & \textcolor{gray}{253} & \textcolor{gray}{0.668} & \textcolor{gray}{0.872} & \textcolor{gray}{0.656} & \textcolor{gray}{5.629} \\ 
        LSS-FLUX.1-schnell (ours) & Latent &  & 2, 3 &\textbf{4.39} & \textbf{1.1x} & \textbf{233} & \textbf{0.695} & \textbf{0.916} & \textbf{0.703} & \textbf{5.412} \\ \hline \hline
        %
        \textcolor{gray}{SDXL~\cite{podellsdxl}} & - & \multirow{5}{*}{2048} & 50 &\textcolor{gray}{46.73} & \textcolor{gray}{1x} & \textcolor{gray}{2,415} & \textcolor{gray}{0.468} & \textcolor{gray}{0.699} & \textcolor{gray}{0.532} & \textcolor{gray}{4.725} \\ 
        SDXL-MegaFusion{~\cite{wu2024megafusion}} & Pixel &  & 40, 10 &\textbf{24.99} & \textbf{1.9x} & \textbf{1,317} & \textbf{0.566} & 0.716 & 0.446 & 5.817 \\ 
        SDXL-Self-Cascade{~\cite{selfcascade:guo2024make}} & Latent & & 50, 50 &56.20 & 0.8x & 3,236 & 0.541 & 0.709 & 0.517 & \textbf{4.137} \\ 
        SDXL-DiffuseHigh{~\cite{kim2025diffusehigh}} & Pixel &  & 50, 15, 15 & 52.38 & 0.9x & 2,518 & 0.550 & 0.681 & 0.513 & 5.960  \\ 
        LSS-SDXL (ours) & Latent &  & 50, 37 & 44.61 & 1x & 2,395 & 0.535 & \textbf{0.732} & \textbf{0.546} & 5.581 \\ \hline 
        \textcolor{gray}{FLUX.1-schnell~\cite{blackforest2024flux}} & - &  \multirow{2}{*}{2048}& 4 &\textcolor{gray}{22.12} & \textcolor{gray}{1x} & \textcolor{gray}{917} & \textcolor{gray}{0.489} & \textcolor{gray}{0.655} & \textcolor{gray}{0.506} & \textcolor{gray}{6.481} \\ 
        LSS-FLUX.1-schnell (ours) & Latent &  & 2, 3 &\textbf{19.46} & \textbf{1.1x} & \textbf{818} & \textbf{0.657} & \textbf{0.782} & \textbf{0.586} & \textbf{5.042} \\  \hline
        \textcolor{gray}{LCM-SDXL~\cite{lcm:luo2023latent}}& - & \multirow{2}{*}{2048}  & 8 &\textcolor{gray}{5.39} & \textcolor{gray}{1x} & \textcolor{gray}{230} & \textcolor{gray}{0.437} & \textcolor{gray}{0.568} & \textcolor{gray}{0.536} & \textcolor{gray}{5.741} \\ 
        LSS-LCM-SDXL(ours) & Latent &  & 4, 6 &\textbf{5.00} & \textbf{1.1x} & \textbf{206} & \textbf{0.502} & \textbf{0.586} & \textbf{0.545} & \textbf{5.514} \\ \hline
    \end{tabular}
}
    \label{tab:results}
\end{table*}

%
\begin{figure*}[t]
\centering
\setlength{\tabcolsep}{0pt}
\begin{tabular*}{\textwidth}{@{\extracolsep{\fill}}ccccc}
SDXL & SDXL-Self-Cascade & SDXL-MegaFusion & SDXL-DiffuseHigh & LSS-SDXL (ours) \\
\includegraphics[width=3.4cm]{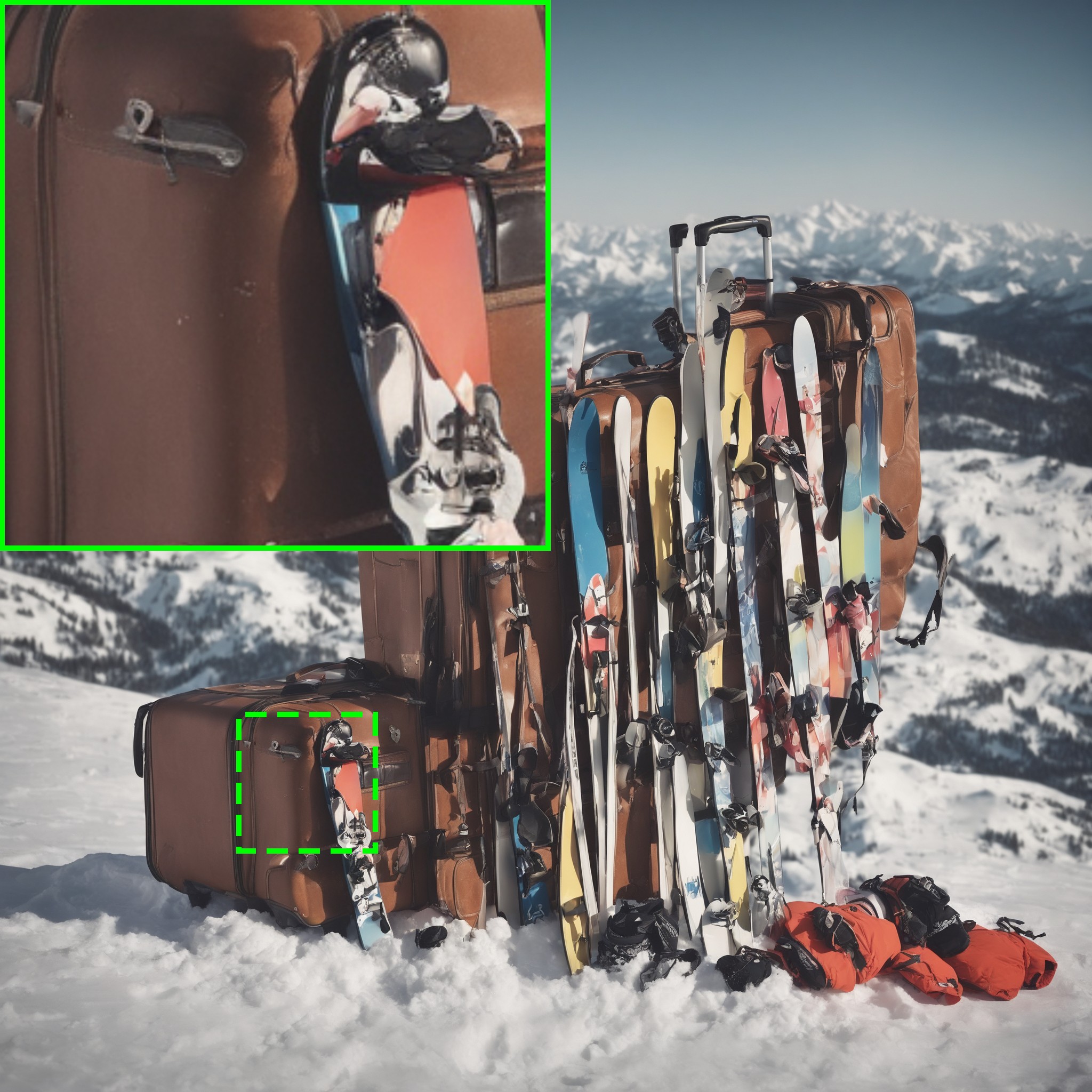}& 
\includegraphics[width=3.4cm]{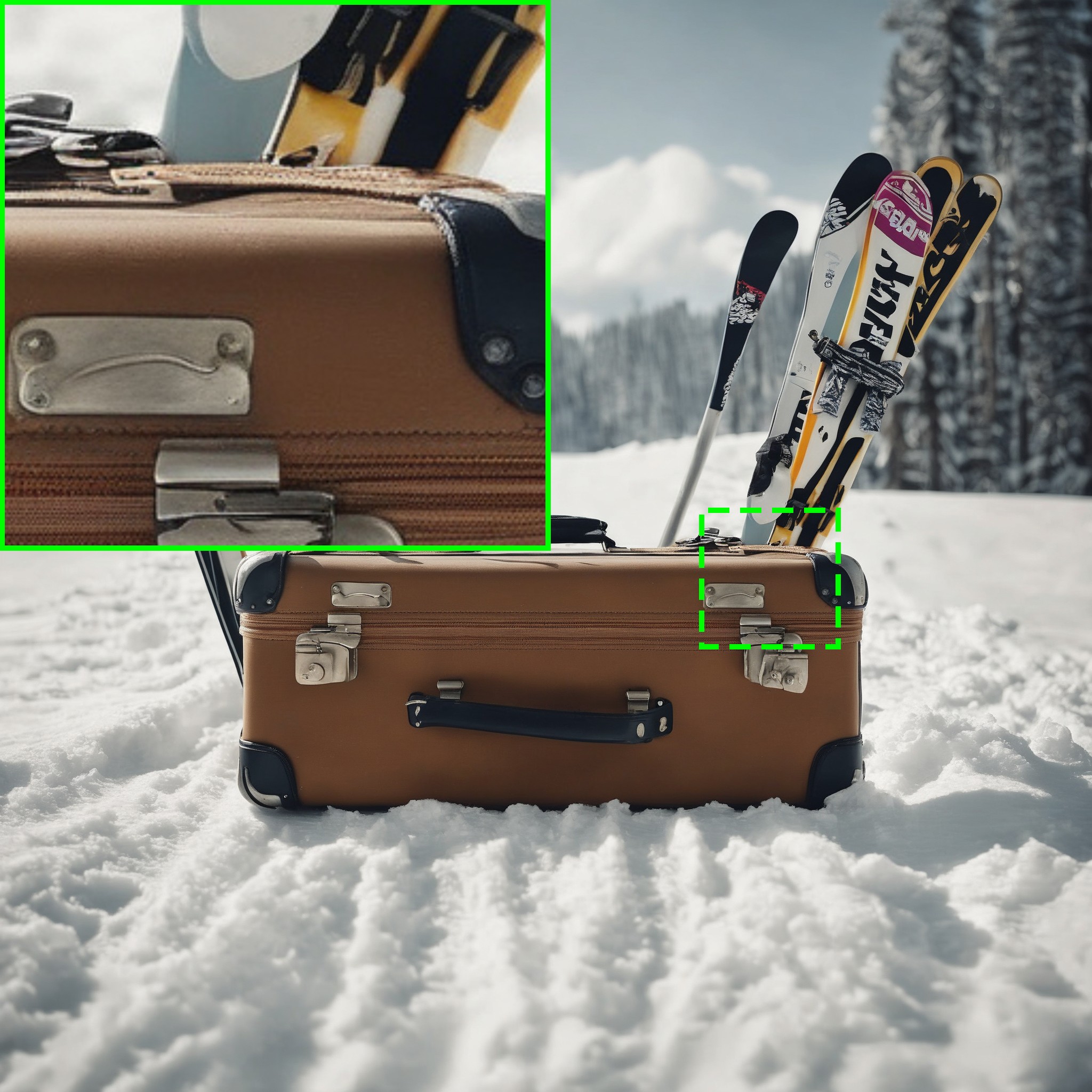}&
\includegraphics[width=3.4cm]{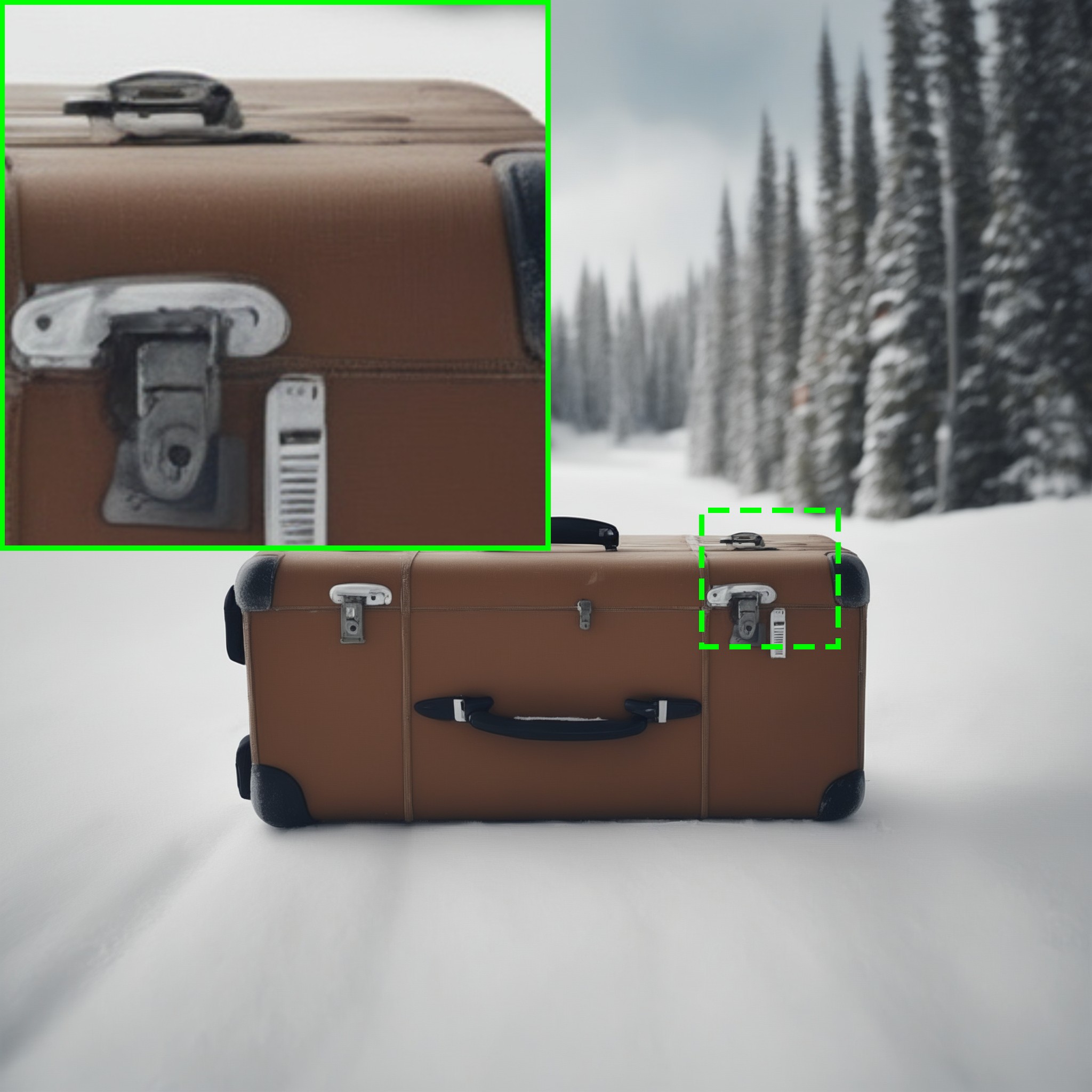}& 
\includegraphics[width=3.4cm]{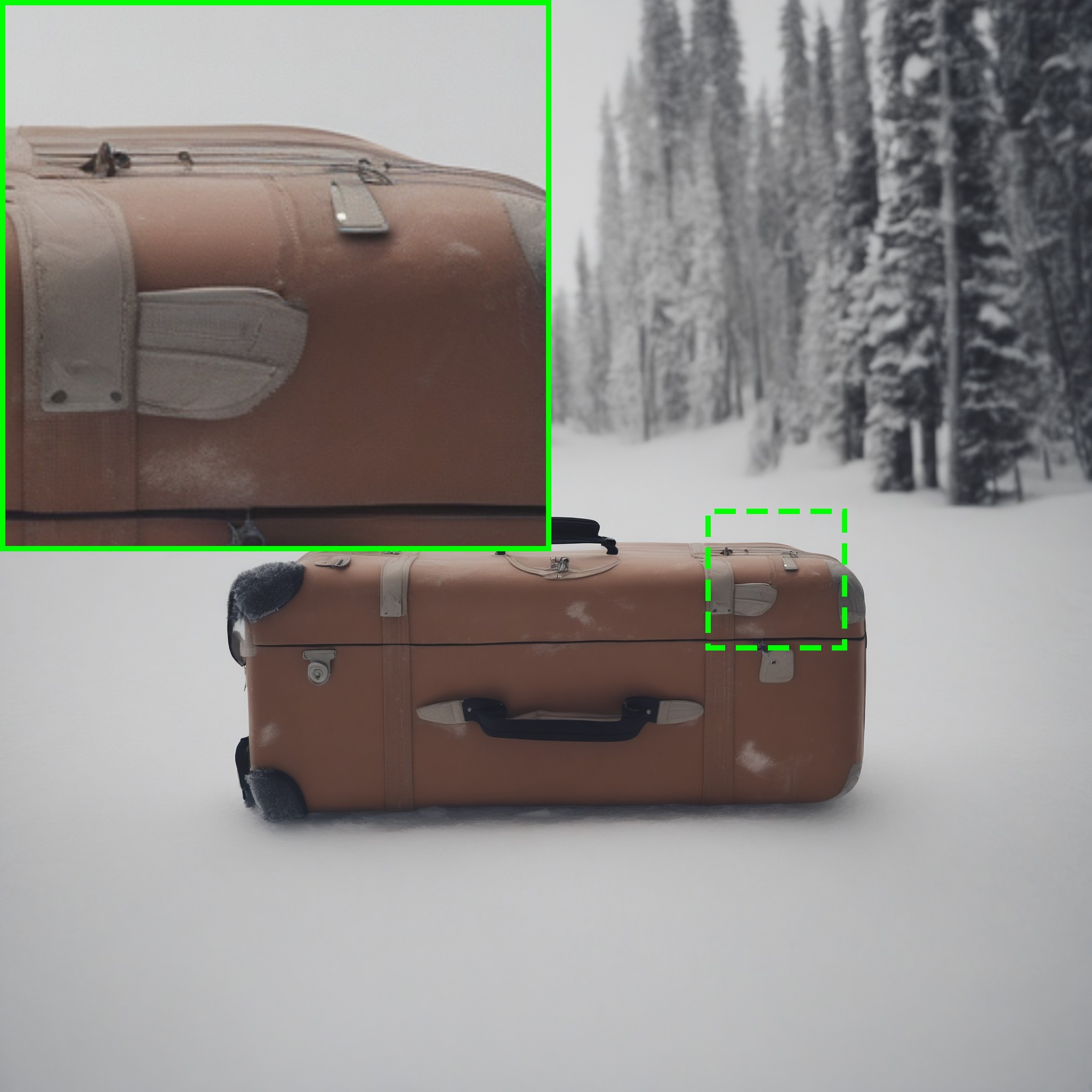}& 
\includegraphics[width=3.4cm]{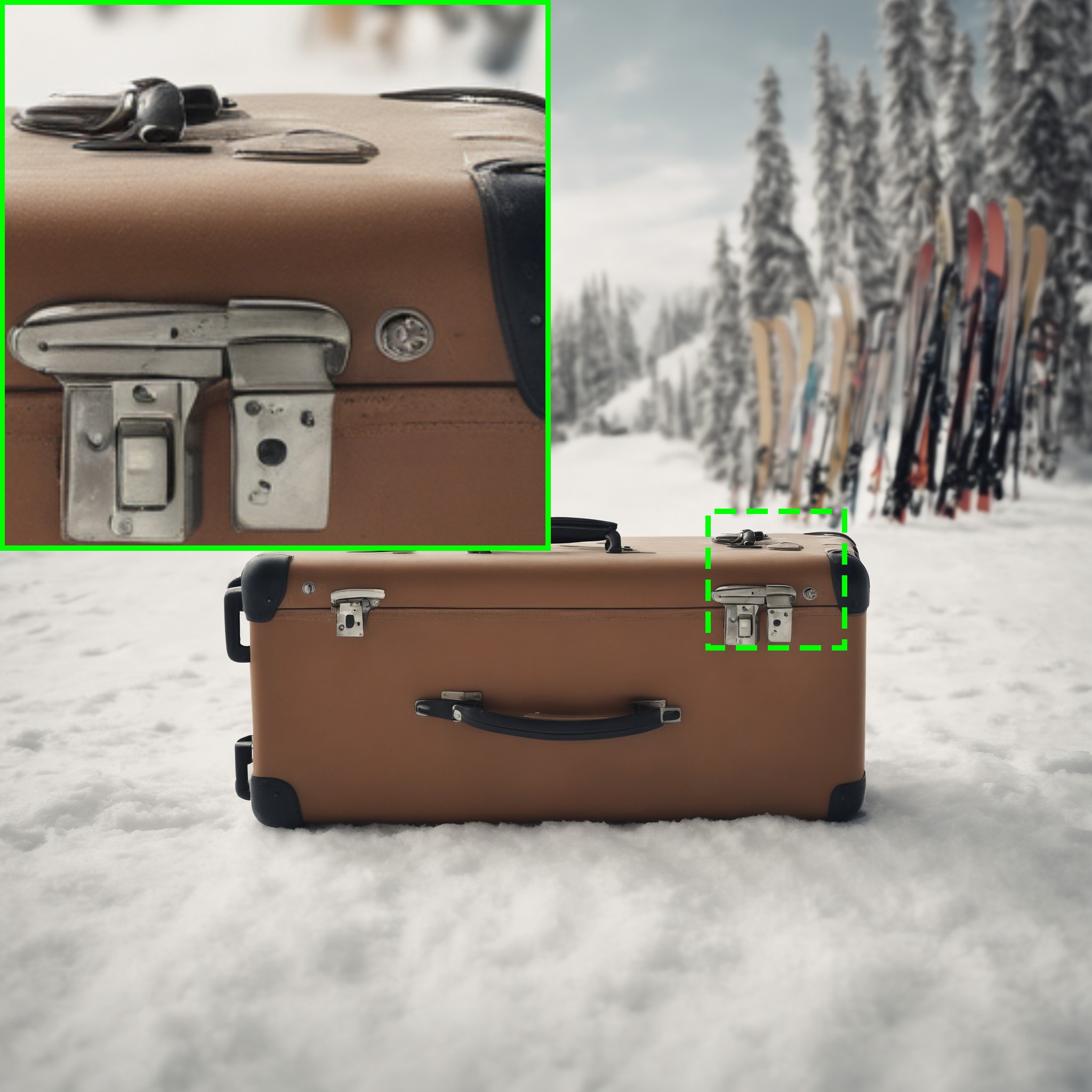}\\ 
\multicolumn{5}{c}{Text prompt: a photo of a suitcase above a skis}\\ 
\end{tabular*}
\vspace{-3mm}
\caption{Visual comparison for different generation frameworks on SDXL with random GenEval prompts at resolution $2048^2$.}
\label{fig:comparison:SDXL}
\vspace{-3mm}
\end{figure*}
%

\section{Experiments}
\label{sec:experiments}

We implement our framework using the Diffuser library {\cite{von-platen-etal-2022-diffusers}} with PyTorch as the backend. All experiments are run on a single NVIDIA A40 GPU.

\subsection{Experimental Setup}

We evaluate {\LSSGen} on two mainstream generative architectures: Rectified Flow (RF) and Diffusion Model (DM). For {\RF}, we use FLUX.1-dev and FLUX.1-schnell~\cite{blackforest2024flux}, both with 12B parameters and multi-resolution generation support. FLUX.1-dev balances image quality and efficiency, while FLUX.1-schnell is optimized for speed. We also include Stable Diffusion 3.5-medium (SD3.5-m), a 2.5B parameter model that offers a lightweight flow-matching baseline. In the {\DM} category, we evaluate: (1) SDXL~\cite{podellsdxl}, trained on resolutions from 512 to 2048; (2) Playground-v2.5~\cite{li2024playground}, an aesthetic-optimized variant of SDXL; (3) Stable Diffusion 1.5 (SD1.5)~\cite{ldm:rombach2022high}, designed for $512^2$ generation; and (4) LCM-SDXL~\cite{lcm:luo2023latent}, a distilled model supporting fast inference.

We compare {\LSSGen} with state-of-the-art (SOTA) pixel-space upsampling methods for efficient high-resolution generation. These include MegaFusion~\cite{wu2024megafusion} and DiffuseHigh~\cite{kim2025diffusehigh}, which directly upscale images in pixel space, as well as Self-Cascade~\cite{selfcascade:guo2024make}, which incorporates conditional low-resolution guidance via model finetuning. All baseline results are reproduced using the officially released code across compatible architectures.

\medskip
\noindent
{\bf Evaluation metrics:}
We evaluate {\LSSGen} using four widely adopted metrics that cover both generation quality and computational efficiency. For text-to-image alignment, we use GenEval~\cite{ghosh2023geneval}, which contains 553 prompts. We generate four images per prompt, following standard practice to ensure statistical robustness.
To assess image quality without losing high-resolution details, we avoid resize-dependent metrics like FID and instead adopt: 
(1) CLIP-IQA~\cite{clipiqa:wang2023exploring}, which modifies CLIP~\cite{clip:radford2021learning} by replacing positional embeddings and using a ResNet backbone for resolution-invariant evaluation; 
(2) TOPIQ~\cite{chen2024topiq}, designed for multi-resolution quality assessment, ideal for high-resolution outputs; (3) NIQE~\cite{niqe:mittal2012making}, a traditional statistical metric based on natural scene statistics, offering a non-learning-based perspective.
All metrics are applied to 2,212 images generated from the GenEval prompts, providing a thorough evaluation across diverse scenarios.

\subsection{Quality and Efficiency Comparison}

\noindent
\textbf{Image generation at $1024^2$ resolution:}
We evaluate {\LSSGen} against SOTA methods across multiple generative architectures. All cascade methods start from $512^2$ and upscale by $1.5 \times$ or $2 \times$ to reach the target resolution. The upper section of Table~\ref{tab:results} summarizes these results.

In {\RF} models,
{\LSSGen} delivers a $1.5 \times$ speedup while maintaining superior image quality. On FLUX.1-dev, {\LSSGen} improves image quality by 3-8\% over the baseline, with only a slight 3\% drop in GenEval alignment.
In contrast, MegaFusion suffers significant quality loss, with a 16\% drop in CLIP-IQA and a 40\% decline in TOPIQ. Similar degradations are observed in SD3.5-m.

In the {\DM} category, {\LSSGen} performs especially well when scaling beyond the model's training resolutions. On Playground-v2.5, {\LSSGen} reduces inference time by 10\% with no loss in image quality. On SD1.5, it yields a 4\% time reduction and an 8\% gain in GenEval alignment. In comparison, MegaFusion offers $2.3 \times$ speedup but causes a 61\% drop in image quality and 8\% decline in alignment.

Finally, {\LSSGen} further improves timestep-distilled models like FLUX.1-schnell, achieving a 9\% speedup while boosting both image quality and alignment performance.

\begin{table}[t]
    \centering
    \caption{Comparing different scaling methods on FLUX.1-dev.\vspace{-2mm}}
    \footnotesize
    \setlength{\tabcolsep}{4pt}  
    \begin{tabular}{ccccc}
    \hline
        Method & GenEval{$\uparrow$} & CLIP-IQA{$\uparrow$} & TOPIQ{$\uparrow$} & NIQE{$\downarrow$} \\  \hline
        Latent Interpolation & 0.605 & 0.885 & 0.651 & \textbf{4.588} \\
        Pixel Space Scaling & 0.612 & 0.752 & 0.429 & 6.495 \\
        ResNet Upsampler & \textbf{0.620} & \textbf{0.900} & \textbf{0.670} & 4.888 \\ \hline
    \end{tabular}
    \vspace{-3mm}
    \label{tab:ablation:latent_upscale}
\end{table}

\begin{figure*}[t]
\centerline{
  \begin{subfigure}[b]{=0.3\linewidth}
  \centerline{
  \includegraphics[width=\linewidth]{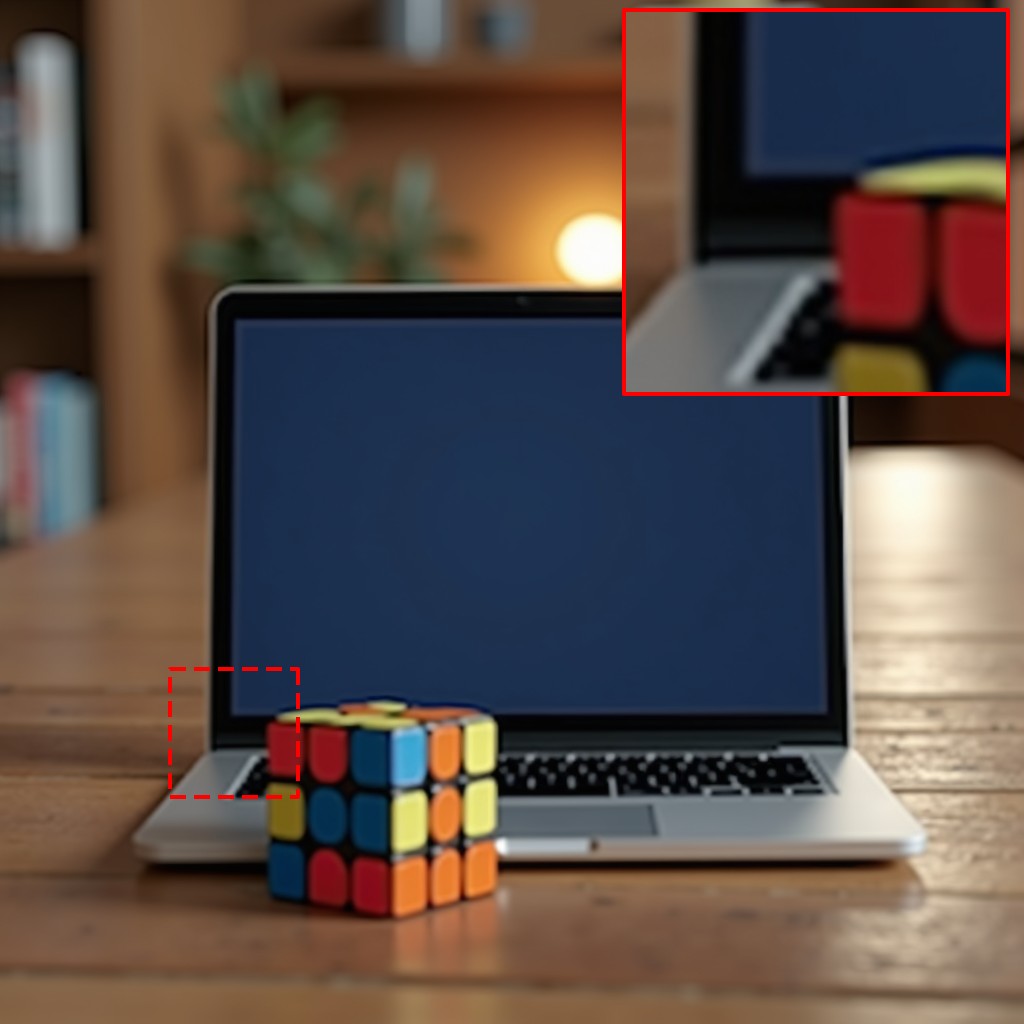}
  }
  \caption{Pixel space upsampling}\label{fig:compare_upscale:pixel}
  \end{subfigure}
  \hspace{0.3cm}
  \begin{subfigure}[b]{=0.3\linewidth}
  \centerline{
  \includegraphics[width=\linewidth]{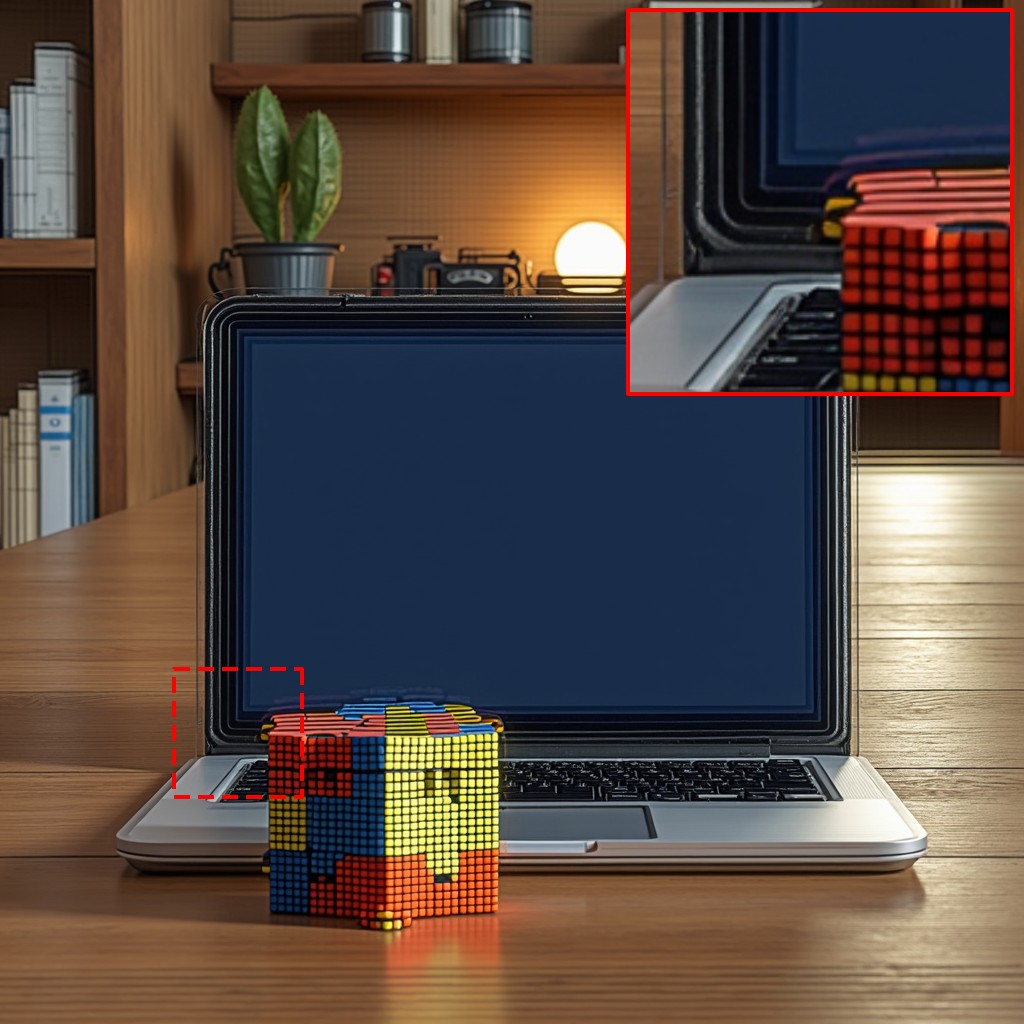}
  }
  \caption{Latent space upsampling}\label{fig:compare_upscale:latent}
  \end{subfigure}
  \hspace{0.3cm}
  \begin{subfigure}[b]{=0.3\linewidth}
  \centerline{
  \includegraphics[width=\linewidth]{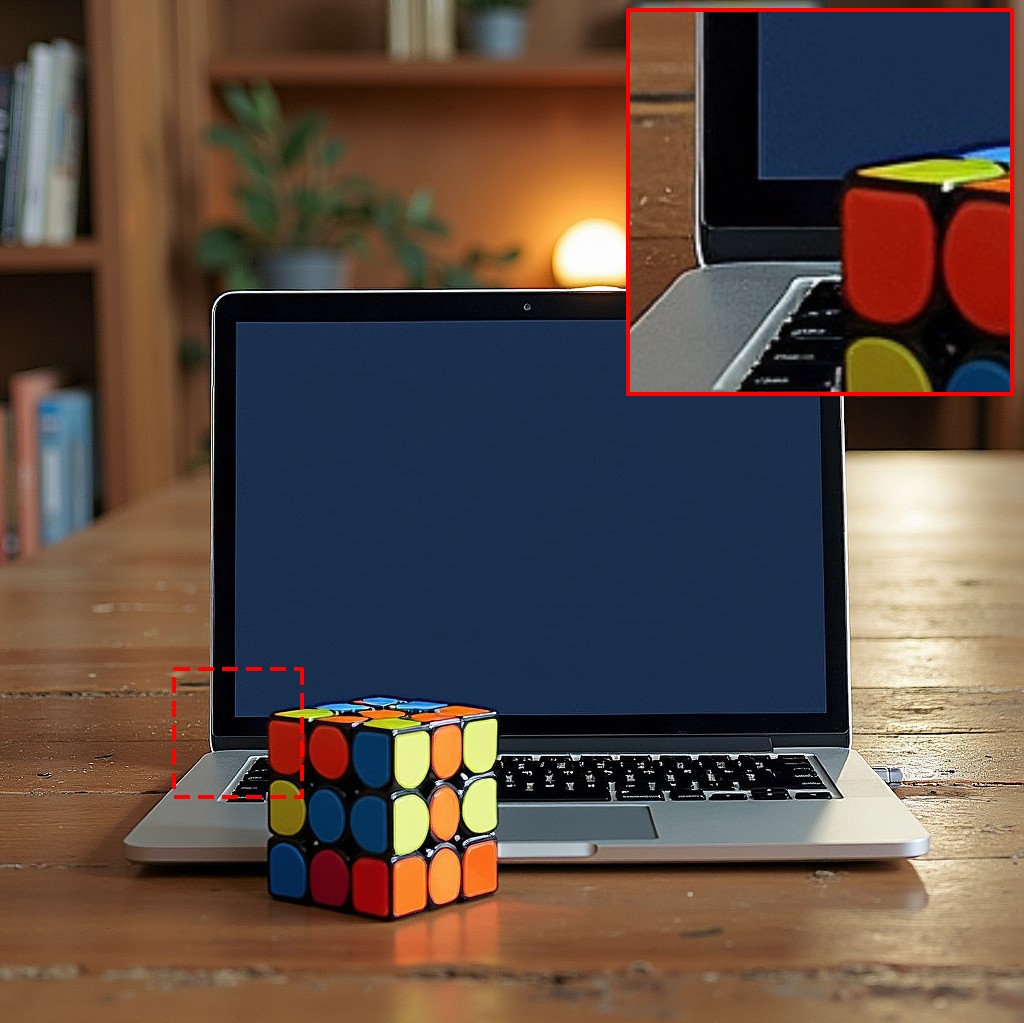}
  }
  \caption{ResNet upsampler (ours)}\label{fig:compare_upscale:resnet}
  \end{subfigure}
}
\vspace{-2mm}
\caption{A visual comparison of resolution scaling methods applied within the \textbf{{\LSSGen}} pipeline, using the FLUX.1-dev~\cite{blackforest2024flux} backbone. The prompt is {\it a laptop displays a Rubik's cube on a table}. (a) Pixel-space scaling results in significant detail degradation and blurring artifacts, even with additional denoising steps. (b) Direct latent space interpolation yields structural inconsistencies, primarily manifesting as edge duplication and boundary artifacts. (c) Our proposed ResNet upsampler preserves structural integrity with enhanced edge definition and coherent object representation, demonstrating superior perceptual quality without introducing scaling-induced artifacts.}
\vspace{-2mm}
\label{fig:compare_upscale}   
\end{figure*}

\noindent
\textbf{Image generation at $2048^2$ resolution:}
We evaluate {\LSSGen} against MegaFusion, Self-Cascade, DiffuseHigh, and standard baselines using SDXL~\cite{podellsdxl}, and extend our comparison to fast, timestep-distilled models like FLUX.1-schnell and LCM-SDXL, which can generate high-resolution images in just a few seconds.
As shown in Table~\ref{tab:results}, MegaFusion improves text-to-image alignment by 21\%, while {\LSSGen} yields a 14\% gain with SDXL.
However, MegaFusion's alignment boost comes at the cost of noticeable quality loss, with blur artifacts reflected in a lower TOPIQ score and illustrated in Fig.~\ref{fig:comparison:SDXL}. In contrast, {\LSSGen} maintains high perceptual quality with only a 5\% increase in speed over the baseline.

{\LSSGen} excels in high-resolution generation with flow-based models. On FLUX.1-schnell at $2048^2$, it improves GenEval alignment by 34\%, CLIP-IQA by 19\%, and TOPIQ by 16\%, while reducing NIQE by 22\% and speeding up inference by 1.1×. For LCM-SDXL, a diffusion-based model, it achieves a 7\% speedup, 14\% better alignment, and a 3\% improvement in CLIP-IQA.

\subsection{Ablation Study}

\noindent
{\bf Latent Upscaling Strategy:}
We evaluate three upsampling methods: naive latent space interpolation, pixel space interpolation, and our ResNet Upsampler with FLUX.1-dev of {\LSSGen}. As shown in Table~\ref{tab:ablation:latent_upscale}, our method significantly outperforms the others in both text-to-image alignment and perceptual quality.
Figure~\ref{fig:compare_upscale} visually highlights the differences: pixel-space interpolation introduces blur from re-encoding, while naive latent interpolation leads to structural discontinuities and edge artifacts. Our ResNet Upsampler mitigates these issues with a lightweight, pretrained architecture, reinforcing the importance of well-designed latent transformations for high-fidelity multi-resolution generation, as discussed in $\S$\ref{subsec:method:scaling}.
However, we identify a minor limitation where at higher resolutions (e.g., $2048^2$), the upsampled latent can become slightly over-sharpened (see supplementary). This characteristic may necessitate additional denoising steps in the final generation stage to ensure optimal quality.



\medskip
\noindent
{\bf Initial Resolution:}
We examine the impact of different initial resolutions in the {\LSSGen} framework, with results shown in Table~\ref{tab:ablation:init_resolution}. Our findings show a clear trade-off between speed and text-to-image alignment. Starting at $256^2$ yields up to $1.6 \times$ speedup for $1024^2$ generation but reduces alignment quality. 
Despite the lower starting resolution, perceptual quality remains strong across all settings. Notably, the $512^2$ to $1024^2$ setup retains nearly the same alignment score as the native $512^2$ baseline, while offering a $1.5 \times$ speedup. 
This suggests that text-to-image alignment is primarily influenced by the starting resolution, not the upscaling.
The slight drop in alignment at very low resolutions likely reflects the hierarchical training of foundation models, which are often pretrained at lower resolutions and fine-tuned on high-quality high-resolution data~\cite{dai2023emu}, limiting their alignment ability.

\begin{table*}[t]
\caption{Impact of initial resolutions in {\LSSGen} on FLUX.1-dev. All metrics are evaluated at the target resolution.
\vspace{-3mm}
}
\label{tab:ablation:init_resolution}
\centerline{
\resizebox{1.0\linewidth}{!}{
    \footnotesize
    \begin{tabular}{ccccccccc}
    \hline
        Init Resolution & Target Resolution & Stage $N$ &Time/img{$\downarrow$} & Speed{$\uparrow$} & GenEval{$\uparrow$} & CLIP-IQA{$\uparrow$} & TOPIQ{$\uparrow$} & NIQE{$\downarrow$} \\  \hline
        \textcolor{gray}{256} & \textcolor{gray}{256} & - & \textcolor{gray}{9.22} & - & \textcolor{gray}{0.584} & \textcolor{gray}{0.854} & \textcolor{gray}{0.747} & \textcolor{gray}{8.729} \\
        256 & 1024 & 3 & \textbf{34.93} & \textbf{1.6x} & 0.594 & \textbf{0.932} & 0.694 & 5.218 \\
        \textcolor{gray}{512} & \textcolor{gray}{512} & - & \textcolor{gray}{17.61} & - & \textcolor{gray}{0.652} & \textcolor{gray}{0.899} & \textcolor{gray}{0.728} & \textcolor{gray}{5.820} \\
        512 & 1024 & 2 & 36.09 & 1.5x & \textbf{0.653} & 0.914 & \textbf{0.705} & \textbf{5.136} \\ 
        \textcolor{gray}{1024} & \textcolor{gray}{1024} & - & \textcolor{gray}{54.38} & \textcolor{gray}{1x} & \textcolor{gray}{0.673} & \textcolor{gray}{0.887} & \textcolor{gray}{0.674} & \textcolor{gray}{5.622} \\ \hline
    \end{tabular}
}}
\vspace{-3mm}
\end{table*}

\begin{table}[t]
\caption{Impact of different initial sigma in {\LSSGen} on SD3.5-m.\vspace{-2mm}
}
\label{tab:ablation:init_noise_sigma}
\centerline{
\resizebox{1.025\linewidth}{!}{
    \setlength{\tabcolsep}{0.7mm}
    \vspace{-2mm}
    \begin{tabular}{c|cccccc}
    \hline
        \InitSigma & Time/img{$\downarrow$} & Speed{$\uparrow$} & GenEval{$\uparrow$} & CLIP-IQA{$\uparrow$} & TOPIQ{$\uparrow$} & NIQE{$\downarrow$} \\  \hline
        0.3 & \textbf{4.59} & \textbf{3.2x} & 0.659 & 0.874 & 0.626 & 5.931 \\
        0.4 & 5.30 & 2.8x & 0.659 & 0.884 & 0.631 & 5.541 \\
        0.5 & 6.39 & 2.3x & 0.660 & \textbf{0.885} & 0.636 & 5.217 \\
        0.6 & 7.46 & 2x & 0.663 & 0.884 & 0.639 & 5.025 \\
        0.7 & 8.90 & 1.7x & 0.675 & 0.882 & \textbf{0.641} & 4.848 \\
        0.75 & 9.97 & 1.5x & 0.672 & 0.880 & \textbf{0.641} & 4.705 \\
        0.8 & 11.02 & 1.3x & 0.673 & 0.879 & \textbf{0.641} & 4.718 \\
        0.9 & 13.53 & 1.1x & \textbf{0.677} & 0.876 & 0.640 & \textbf{4.622} \\
        \textcolor{gray}{1.0} & \textcolor{gray}{14.73} & \textcolor{gray}{1x} & \textcolor{gray}{0.676} & \textcolor{gray}{0.862} & \textcolor{gray}{0.635} & \textcolor{gray}{4.825} \\ \hline
    \end{tabular}
}}
\vspace{-2mm}
\end{table}

\medskip
\noindent
{\bf Initial Noise Coefficient ({\InitSigma}):}
As discussed in $\S$\ref{subsec:method:resolutionAR}, {\LSSGen} uses the initial noise coefficient {\InitSigma} to control the stochastic perturbation during upsampling. Table~\ref{tab:ablation:init_noise_sigma} shows how different {\InitSigma} values affect the trade-off between speed and image quality. Lower {\InitSigma} values offer greater speedups with minimal loss in semantic alignment. As {\InitSigma} approaches the theoretical optimal value of 0.75, perceptual quality improves steadily, though at the cost of reduced efficiency. We find that values in the range [0.7, 0.8] offer the best balance between image quality and a $1.3 \times$ to $1.7 \times$ speedup over the baseline. This supports our theoretical insight that $\sigma_\text{init} \approx 0.75$ is optimal for half-resolution initialization. 
However, the framework also offers flexibility for applications where throughput is the primary concern. By setting {\InitSigma} to lower values, a slight degradation in perceptual quality can be exchanged for more significant computational gains. For example, using a 
{\InitSigma} between 0.3 and 0.5 can achieve a $2.3 \times$ to $3.2 \times$ speedup.
Conversely, {\InitSigma} values exceeding 0.8 yield diminishing quality gains with significantly higher computational cost, confirming that our framework effectively captures the trade-offs in resolution-aware noise scheduling.

\medskip
\noindent
{\bf Impact of Schedule Shifting:}
We evaluate the effect of the shifting strategy from $\S$\ref{subsec:method:shifting} in the {\LSSGen} framework using SD3.5-m, with results shown in Table~\ref{tab:ablation:setting}.
Shifting reduces inference time by 26\%, achieving a $1.5 \times$ speedup over the baseline---compared to $1.1 \times$ without shifting. It also improves perceptual quality by about 3\%. Although NIQE slightly worsens by 3\%, this is a reasonable trade-off for the overall gains in efficiency and quality.
These results support our theoretical insight that concentrating more timesteps in early, low-resolution stages improves efficiency without compromising image fidelity, making shifting a natural fit for resolution-aware generation.


\medskip
\noindent
{\bf Shorten intermediate steps:}
We investigate the possibility of reducing sampling steps in lower-resolution where the denoising process is expected to be less complex.
The sampling step reduction at lower resolution stages is according to the stage $n$, with $T' = T/\left(2^{(N-n)}\right)$. Table~\ref{tab:ablation:setting} demonstrates that shortened steps at early stages of {\LSSGen} yield efficiency gains ($1.2 \times$ to $1.5 \times$) with minimal impact on generation quality. 
Computational gains emerge with negligible quality impact, illuminating the potential to compress the generation process in a resolution-dependent manner. 

\begin{table}[t]
\caption{The effects of the schedule shifting (SD3.5-m backbone) and shortening early stage timesteps (FLUX.1-dev backbone).\vspace{-2mm}
}
\label{tab:ablation:setting}
\centerline{
\setlength{\tabcolsep}{0.5mm}
\resizebox{1.0\linewidth}{!}{
    \vspace{-2mm}
    \begin{tabular}{c|cccccc}
    \hline
        Setting & Time/img{$\downarrow$} & Speed{$\uparrow$} & GenEval{$\uparrow$} & CLIP-IQA{$\uparrow$} & TOPIQ{$\uparrow$} & NIQE{$\downarrow$} \\  \hline
        Shift & \textbf{9.97 s} & \textbf{1.5x} & \textbf{0.672} & \textbf{0.880} & \textbf{0.641} & 4.705 \\ 
        No Shift & 13.53 s & 1.1x & 0.653 & 0.852 & 0.621 & \textbf{4.546} \\ \hline \hline
        Shorten Steps & \textbf{35.79 s} & \textbf{1.5x} & 0.653 & \textbf{0.914} & \textbf{0.705} & \textbf{5.136} \\ 
        Regular Steps & 44.81 s & 1.2x & \textbf{0.658} & 0.910 & 0.700 & 5.246 \\ \hline
    \end{tabular}
}    
}
\vspace{-4mm}
\end{table}

%% file: sec/6_conclusion.tex
\section{Conclusion}
\label{sec:conclusion}

This paper presents {\LSSGen}, a general and efficient framework for accelerating inference in both diffusion and flow-based generative models. The core idea is to shift the early generation steps to a low-resolution latent space and progressively upscale the result to the target resolution. This design enables high-resolution synthesis with significantly reduced computation while preserving visual fidelity.

Unlike previous pixel-space upsampling methods, {\LSSGen} operates entirely in latent space, requiring no architectural changes or model retraining. It offers a lightweight and generalizable ResNet-based latent upsampler and a principled schedule-shifting mechanism to maximize efficiency during denoising. Extensive experiments across multiple SOTA models (e.g., FLUX, SDXL, SD1.5, Playground-v2.5, and LCM) and resolutions ($1024^2$ and $2048^2$) show that {\LSSGen} achieves up to 246\% improvement in perceptual quality, 1.5× speedup, and stronger semantic alignment, even on models not trained for high resolutions.

In addition to strong empirical results, the paper provides theoretical formulations for resolution-aware SNR adjustment, initialization strategies, and probabilistic path planning for dynamic latent upscaling. These insights help bridge the gap between efficiency and fidelity.




{\bf Future work} includes extending {\LSSGen} to more complex tasks such as video generation, image editing, and inpainting, where latent-space resolution dynamics are even more critical. We also plan to explore adaptive scaling schedules for further gains.


%% file: sec/_supplementary.tex
\clearpage
\setcounter{page}{1}
\renewcommand{\theequation}{S\arabic{equation}}
\renewcommand{\thefigure}{S\arabic{figure}}
\renewcommand{\thetable}{S\arabic{table}}
\setcounter{equation}{0}    
\setcounter{figure}{0}    
\setcounter{table}{0}  
\maketitlesupplementary


\section{Complexity Analysis}
We analyze the computational complexity implications of resolution scaling in diffusion models. When image resolution increases by a factor of $k$, the computational requirements grow quadratically as $O(k^2)$ due to the direct relationship between resolution and total pixel count. This scaling effect becomes particularly significant in transformer-based architectures like {\DiT}~\cite{dit:peebles2023scalable}, where self-attention~{\cite{vaswani2017attention}} operations exhibit quadratic complexity with respect to sequence length. 
For generation tasks beyond $1024^2$ resolution, this computational burden becomes prohibitive. Our progressive dynamic resolution approach addresses this limitation by performing initial probabilistic paths at lower resolutions, thereby substantially reducing overall computational demands. The computational complexity manifests in both transformer-based and CNN architectures, though with different scaling characteristics. For transformer models utilizing self-attention mechanisms with complexity ${O}(n^2)$ \cite{vaswani2017attention}, where $n$ represents the number of image patches, the total computational complexity scales as:
%
$$
C(H,W) = O((H \times W) \cdot (H \times W)) = O((H \times W)^2)
$$
%
where $C(H,W)$ denotes the computational cost for an image of height $H$ and width $W$. This formulation yields two crucial insights: 
\begin{enumerate}
    \item Reducing patch count through our dynamic resolution strategy provides quadratic computational savings in self-attention operations. 
    \item The quartic relationship between resolution and computational requirements (in both FLOPs and memory) demonstrates the theoretical efficiency of our progressive resolution approach compared to fixed-resolution methods.
\end{enumerate}

\section{Case Study}

\subsection{Effect of Initial Noise Intensity Parameter}
\label{subsec:appendix:case:init_sigma}
\begin{figure*}[t]
\centering
\begin{tabular*}{\textwidth}{@{\extracolsep{\fill}}ccc}
\toprule
\includegraphics[width=5.4cm]{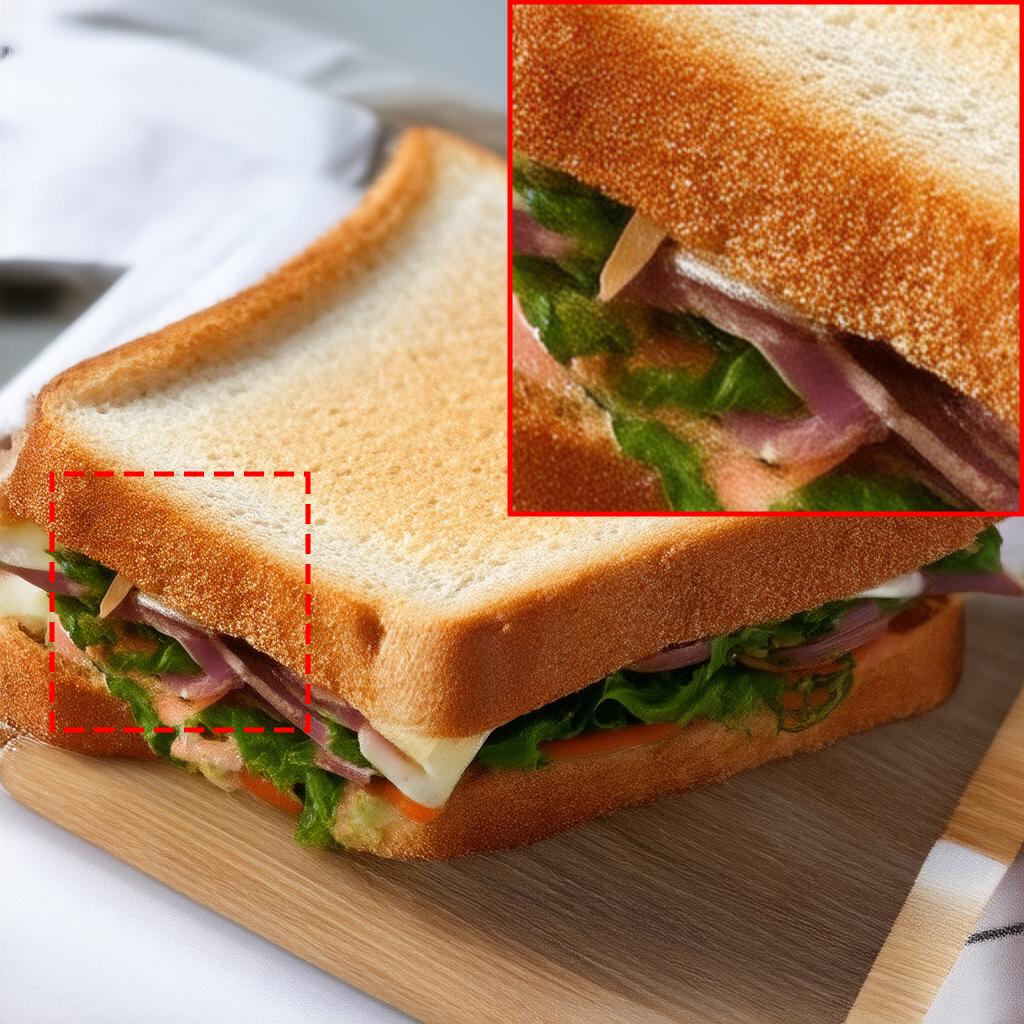}& 
\includegraphics[width=5.4cm]{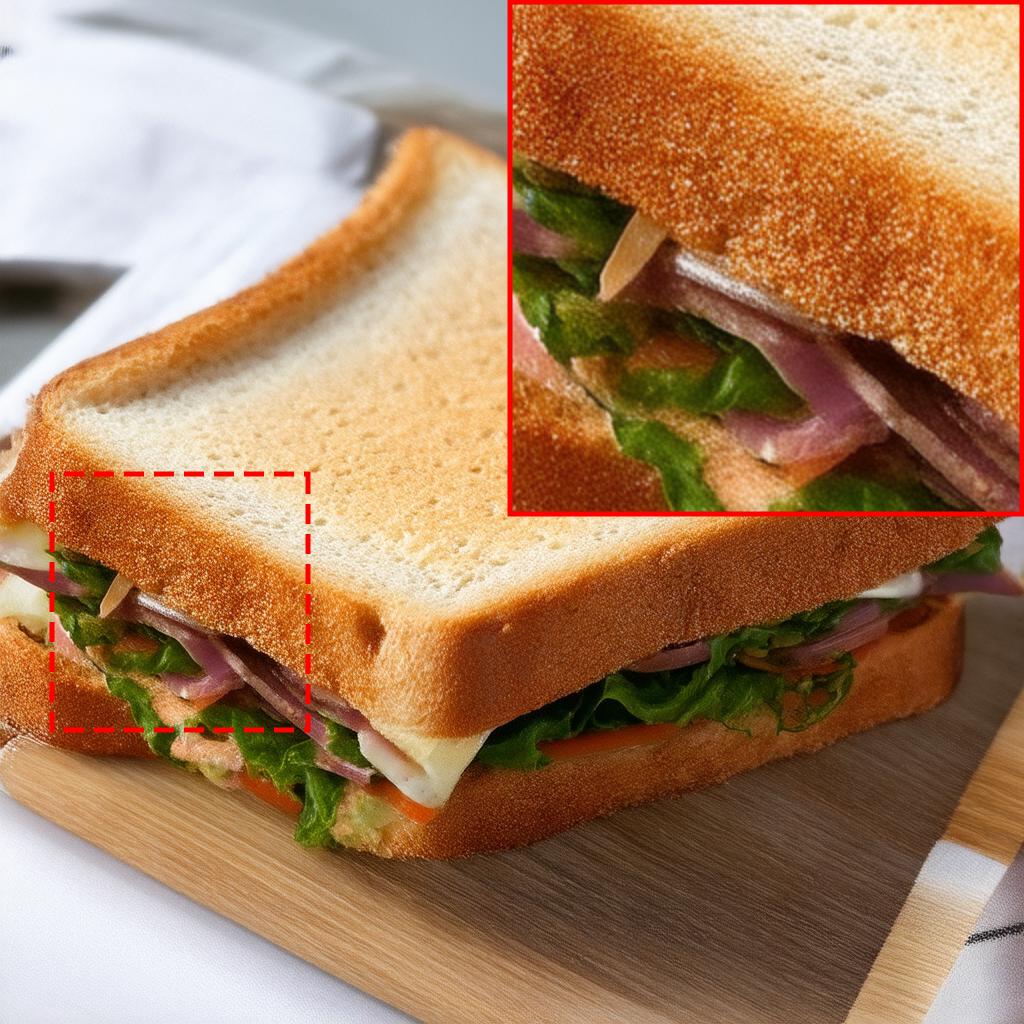}&
\includegraphics[width=5.4cm]{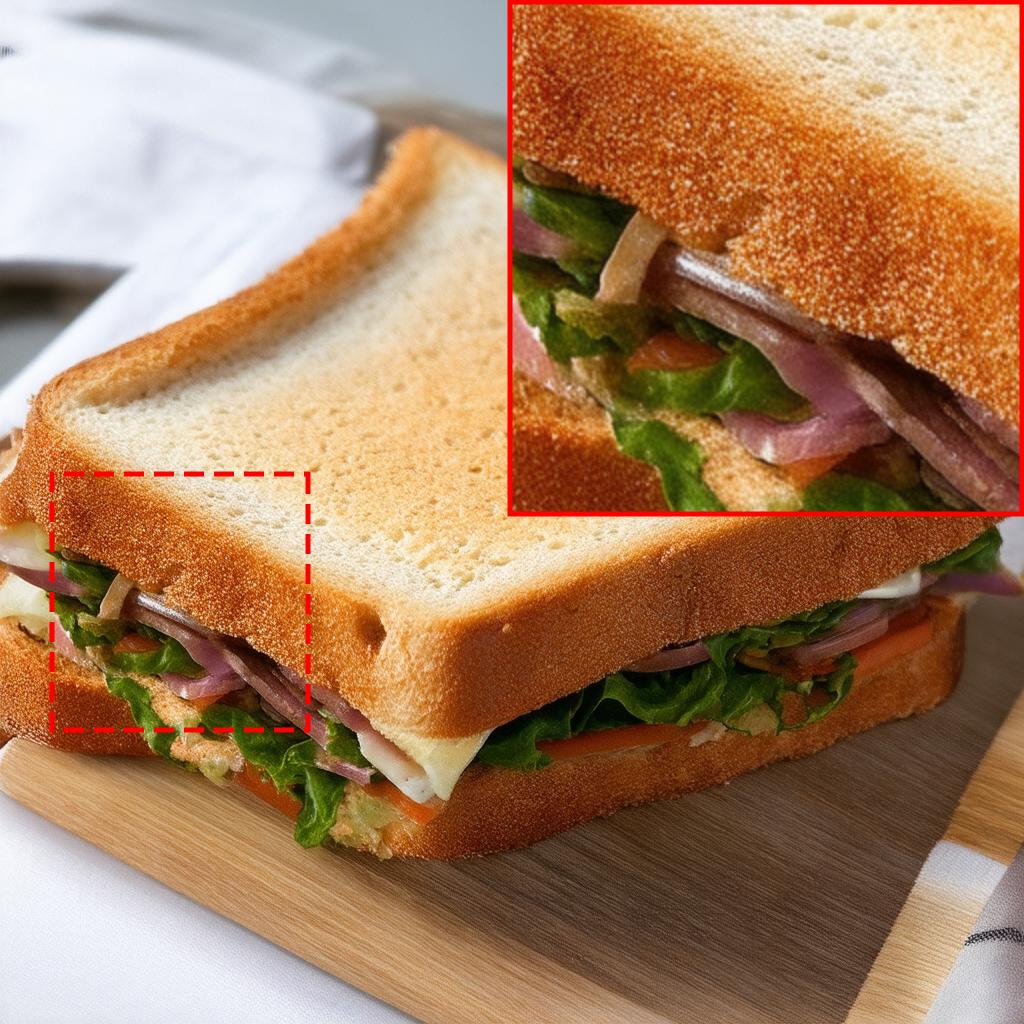}\\
$\sigma_\text{init} = 0.3$&
$\sigma_\text{init} = 0.4$&
$\sigma_\text{init} = 0.5$\\ 
\midrule
\includegraphics[width=5.4cm]{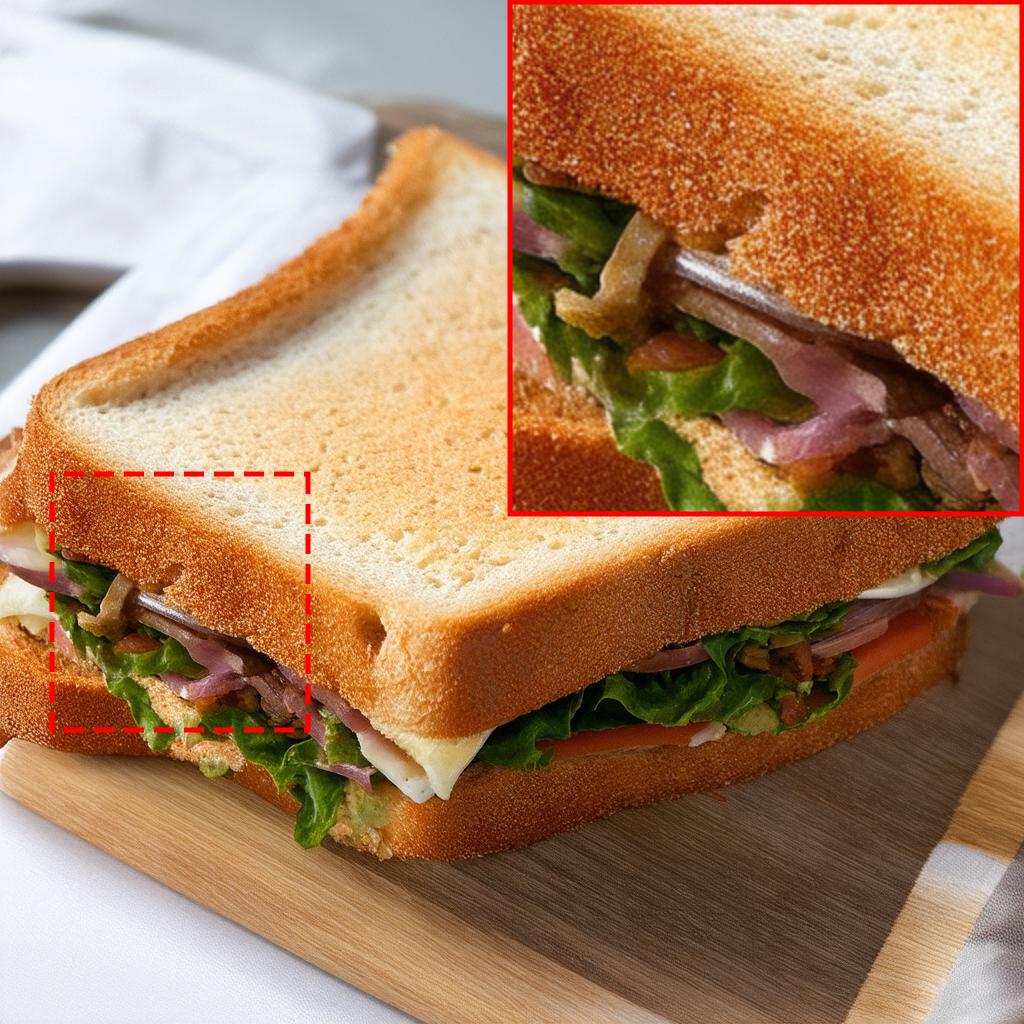}&
\includegraphics[width=5.4cm]{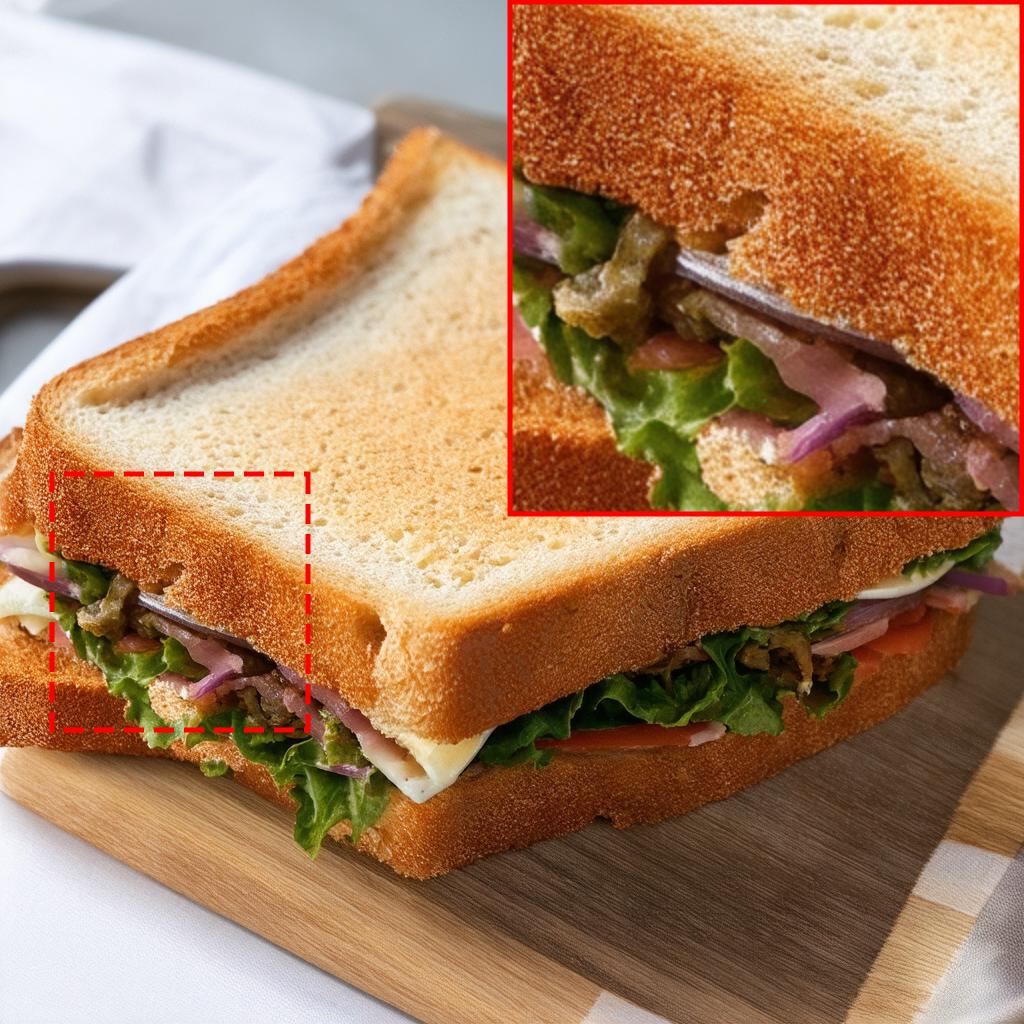}&
\includegraphics[width=5.4cm]{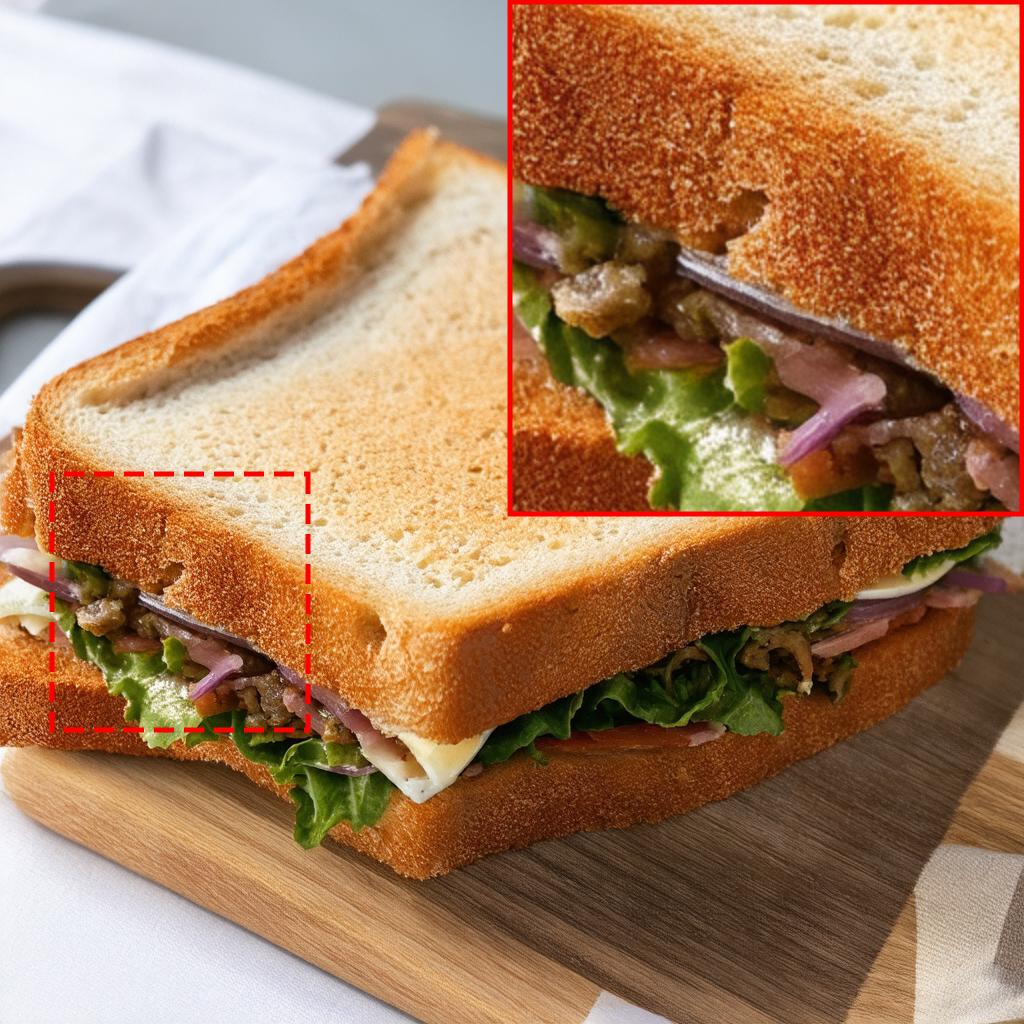}\\
$\sigma_\text{init} = 0.6$&
$\sigma_\text{init} = 0.7$&
$\sigma_\text{init} = 0.75$\\ 
\midrule
\includegraphics[width=5.4cm]{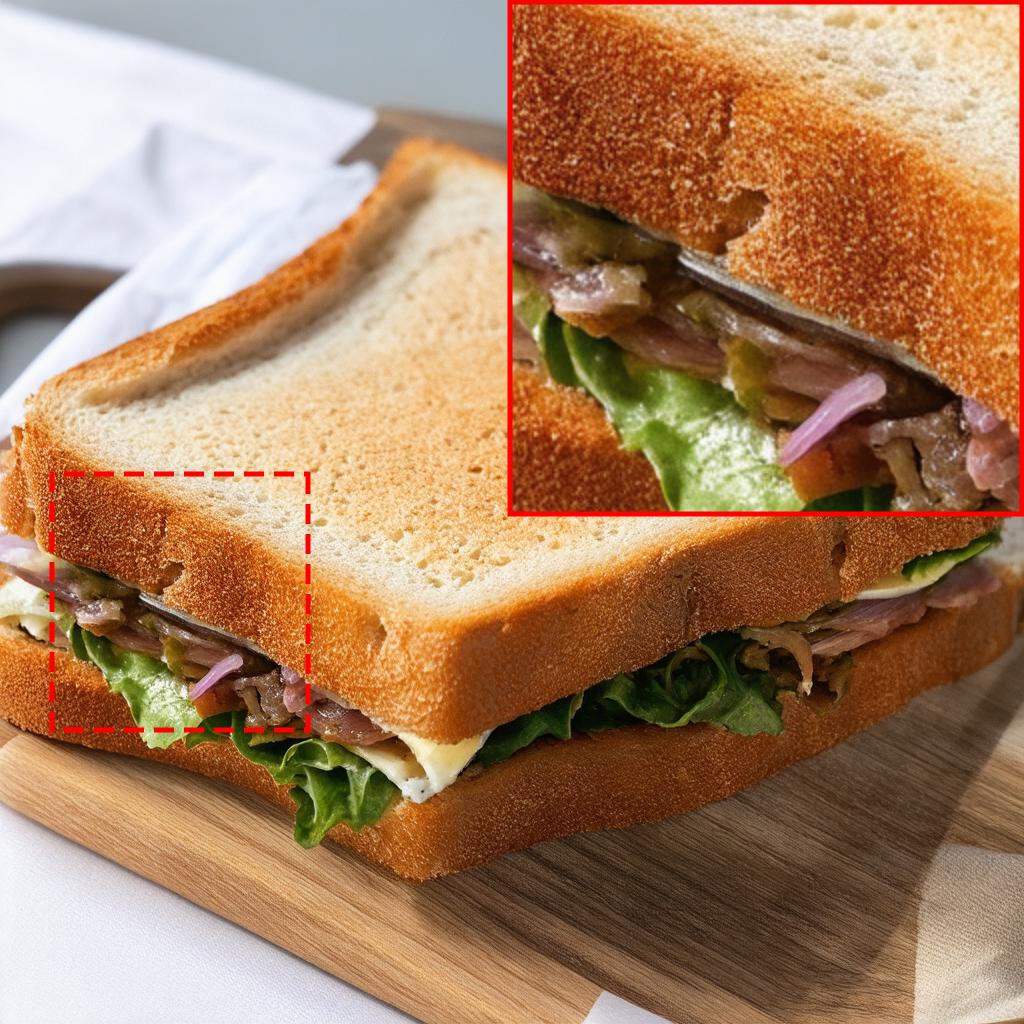}& 
\includegraphics[width=5.4cm]{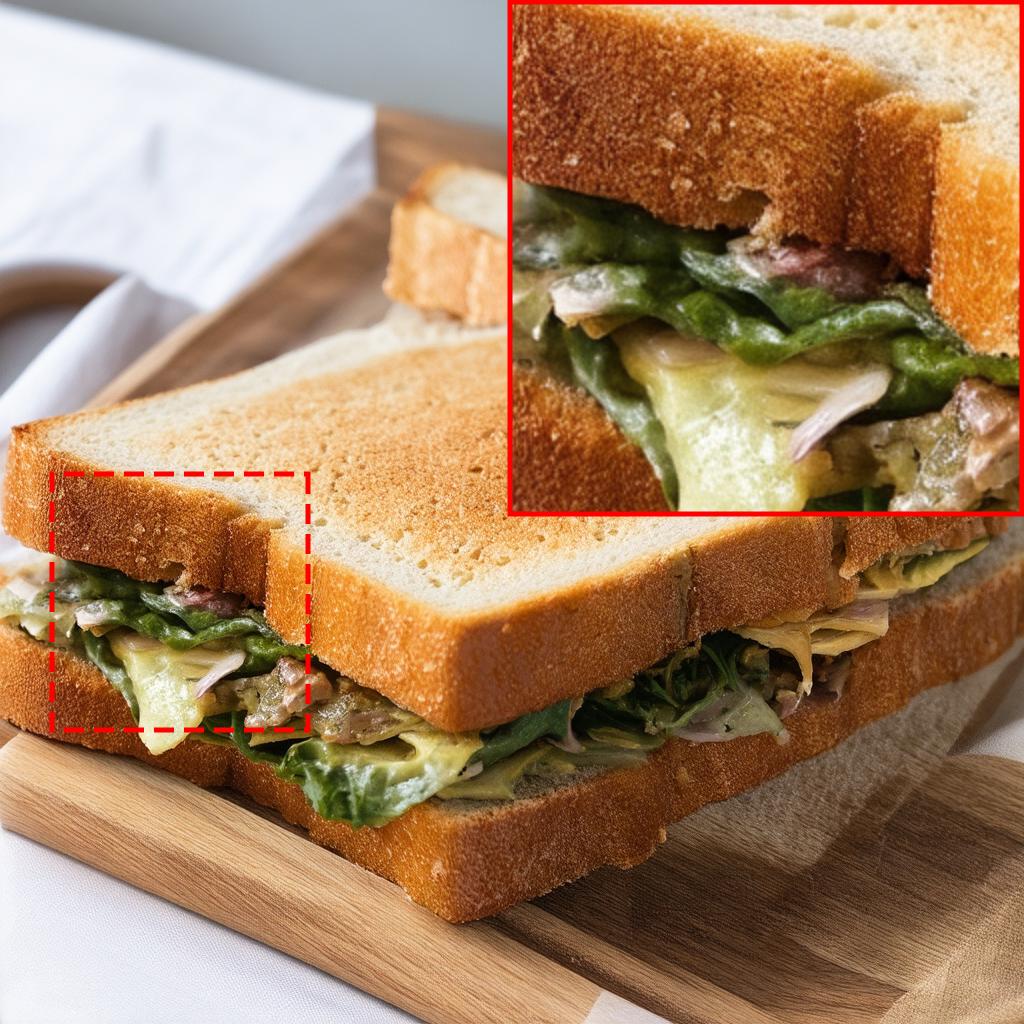}\\
$\sigma_\text{init} = 0.8$&
$\sigma_\text{init} = 0.9$\\
\bottomrule
\end{tabular*}
\caption{Comparison between different $\sigma_\text{init}$ setting on SD3.5-medium~{\cite{sd3:esser2024scaling}}.}
\label{fig:ablation:init_sigma:SD3}
\end{figure*}

The parameter {\InitSigma} controls the initial noise intensity in the {\DynaFlow} generation stages. Fig.~\ref{fig:ablation:init_sigma:SD3} illustrates the visual impact of different {\InitSigma} values. The results demonstrate that higher {\InitSigma} values produce images with enhanced detail sharpness. Among all configurations examined, $\sigma_\text{init}=0.75$ yields optimal quality across the tested parameter range. 
Our method performs upsampling operations in latent space rather than pixel space, which preserves the structural integrity of the generated images across different parameter settings. This observation suggests a configurable quality-speed tradeoff: a slight reduction in image detail fidelity can yield significant inference acceleration. Thus, {\InitSigma} serves as an effective control parameter that enables practitioners to balance generation quality against computational efficiency according to application requirements.
\subsection{Comparative Analysis of Progressive Scaling Approaches in Flow Model}
\label{subsec:appendix:case:flux}
\begin{figure*}[t]
\centering
\begin{tabular*}{\textwidth}{@{\extracolsep{\fill}}ccc}
\toprule
FLUX.1-dev & FLUX.1-dev-MegaFusion & LSS-FLUX.1-dev (ours) \\
\midrule
\includegraphics[width=5.3cm]{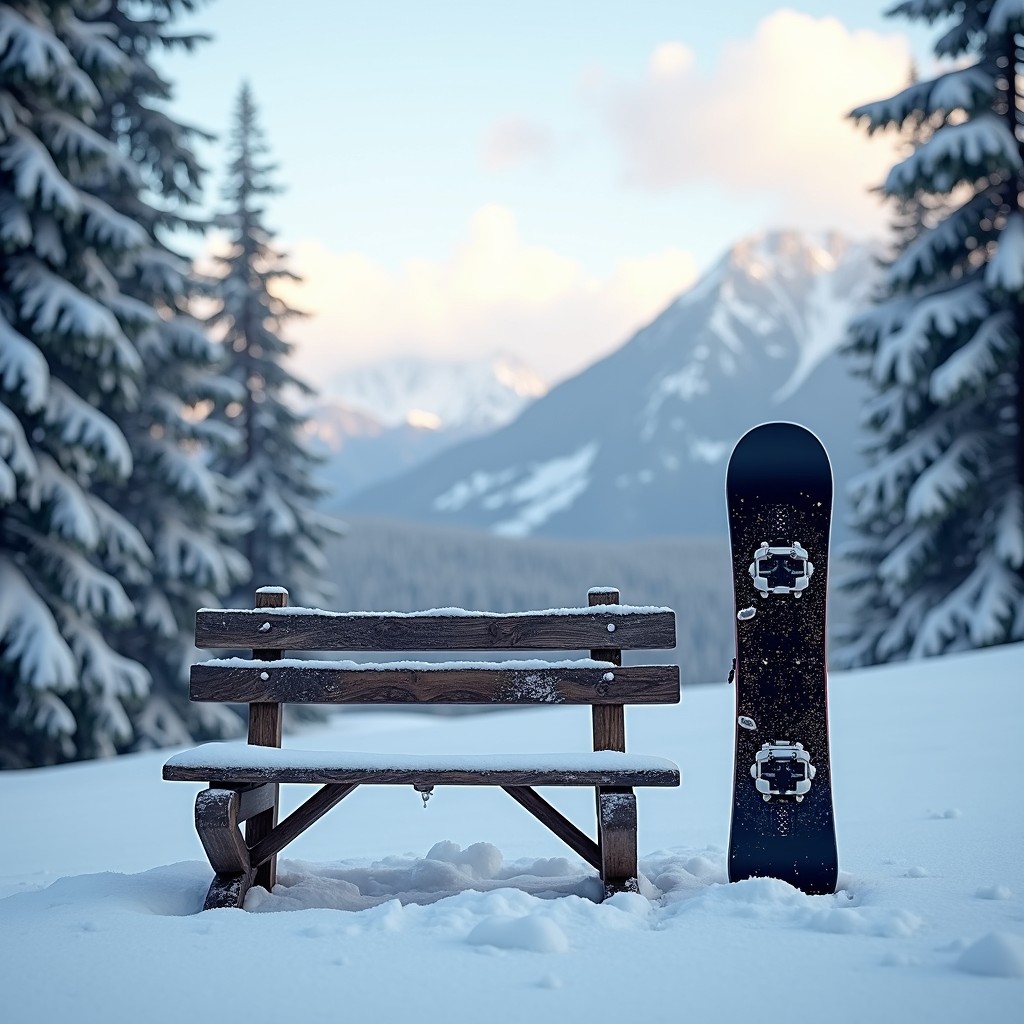}& 
\includegraphics[width=5.3cm]{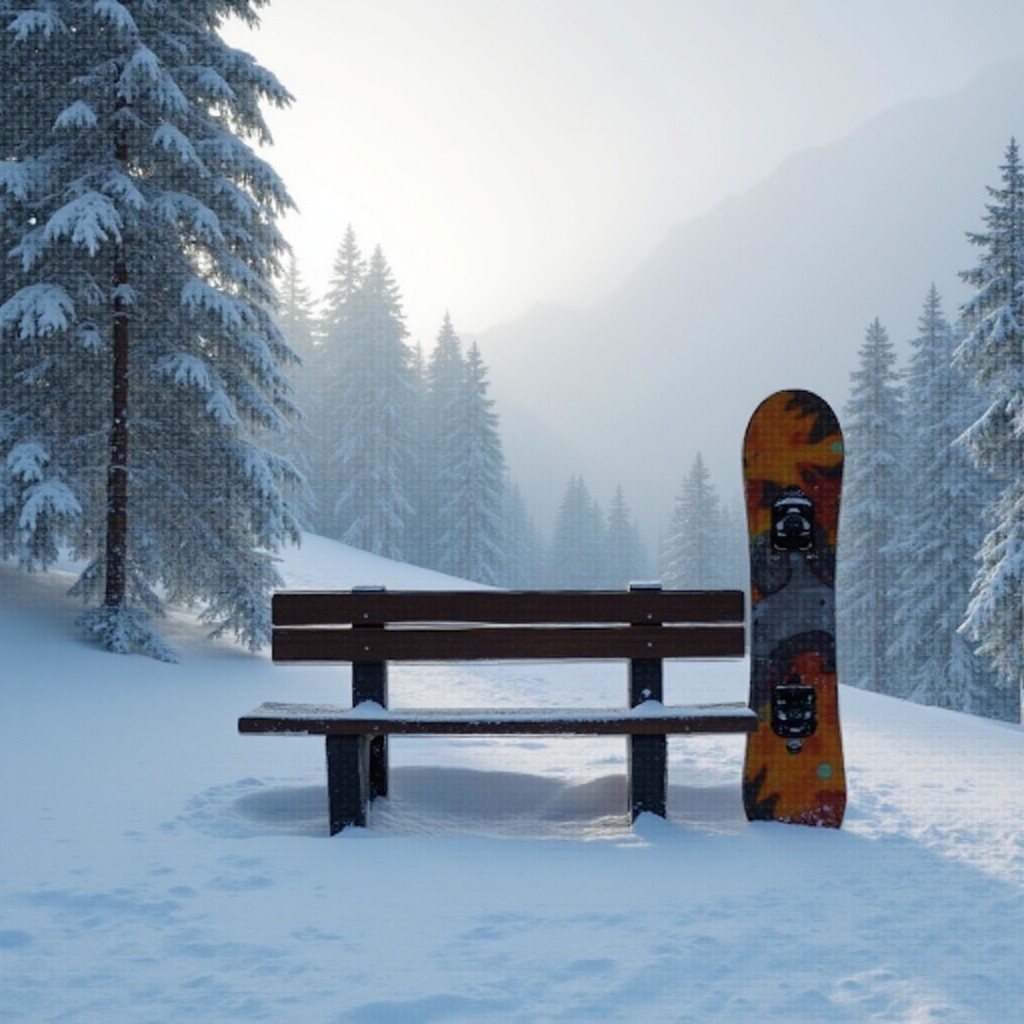}& 
\includegraphics[width=5.3cm]{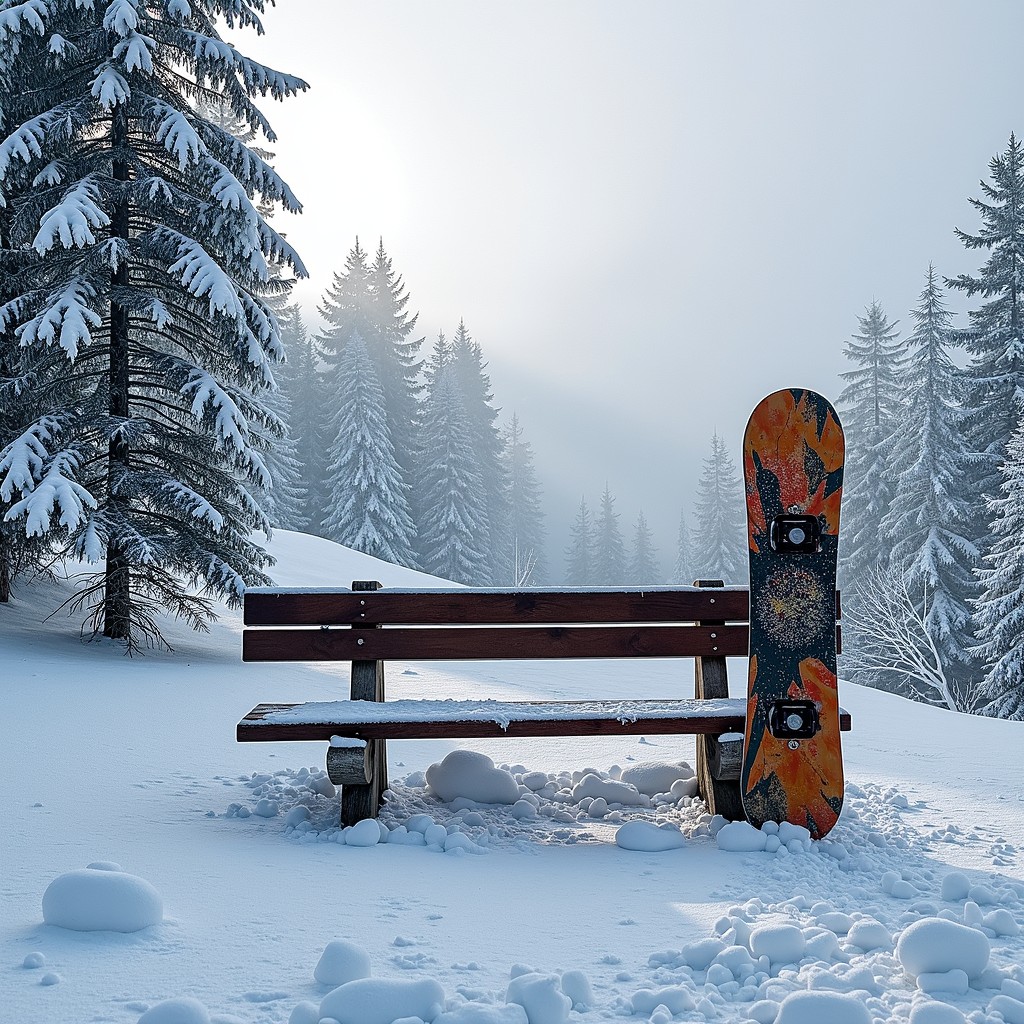}\\ 
\multicolumn{3}{c}{a photo of a bench and a snowboard.}\\
\midrule
\includegraphics[width=5.3cm]{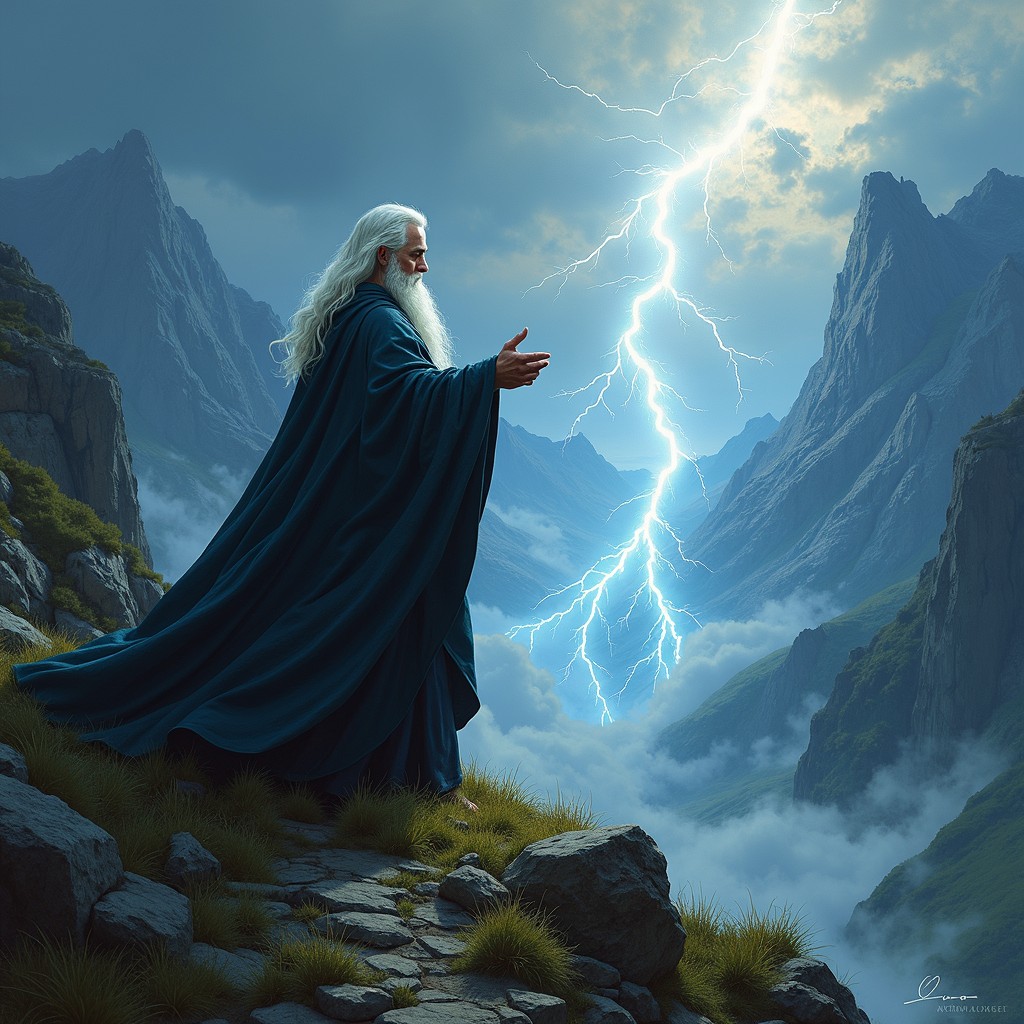}& 
\includegraphics[width=5.3cm]{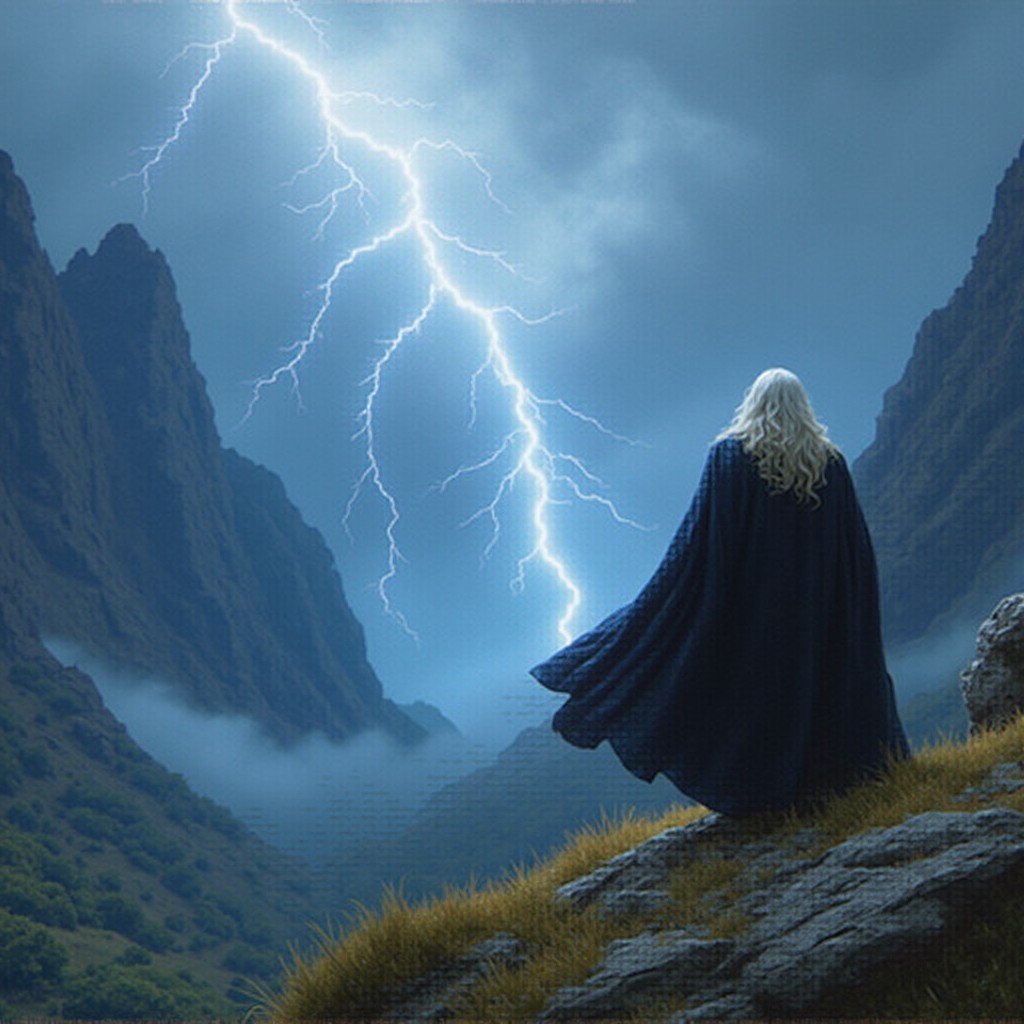}& 
\includegraphics[width=5.3cm]{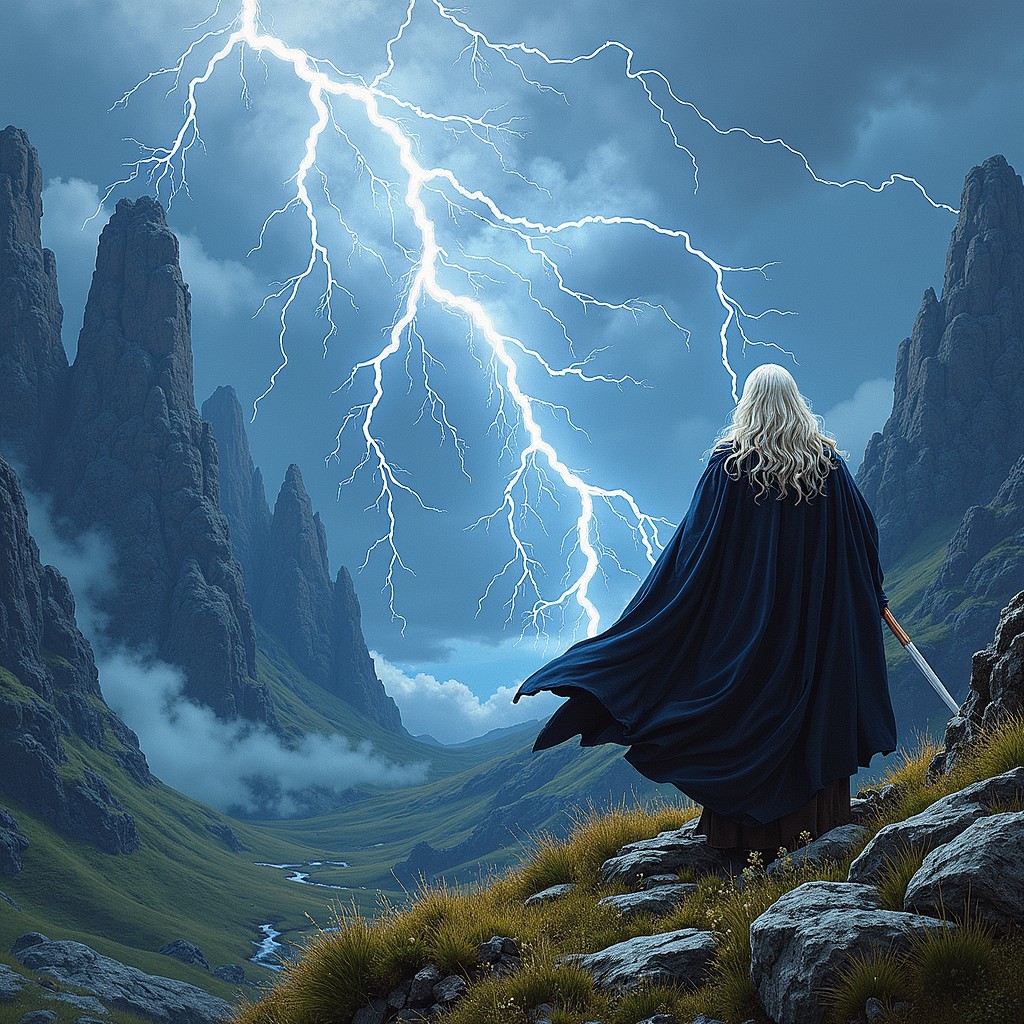}\\ 
\multicolumn{3}{c}{An epic painting of Gandalf the Black summoning thunder and lightning in the mountains.}\\
\midrule
\includegraphics[width=5.3cm]{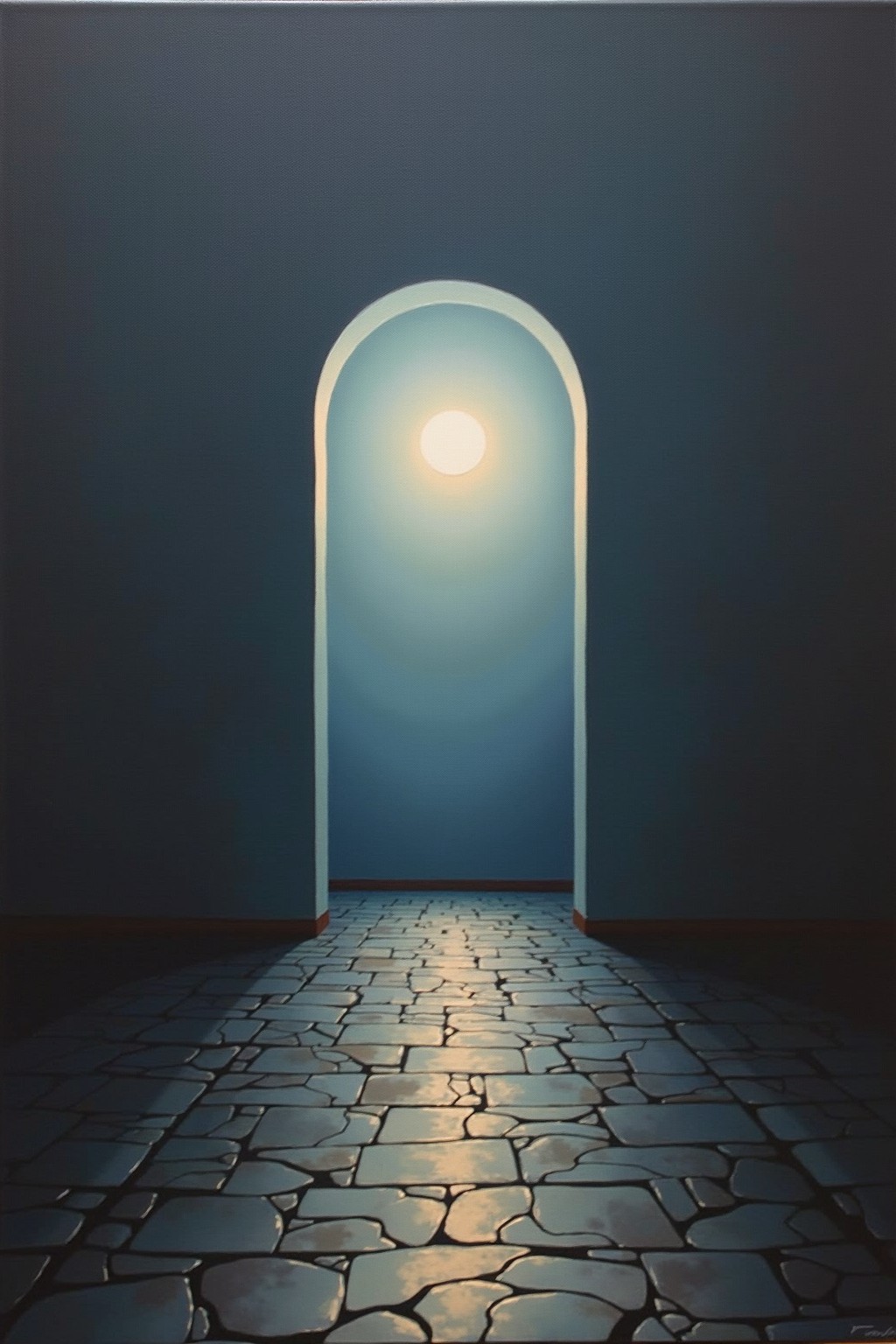}& 
\includegraphics[width=5.3cm]{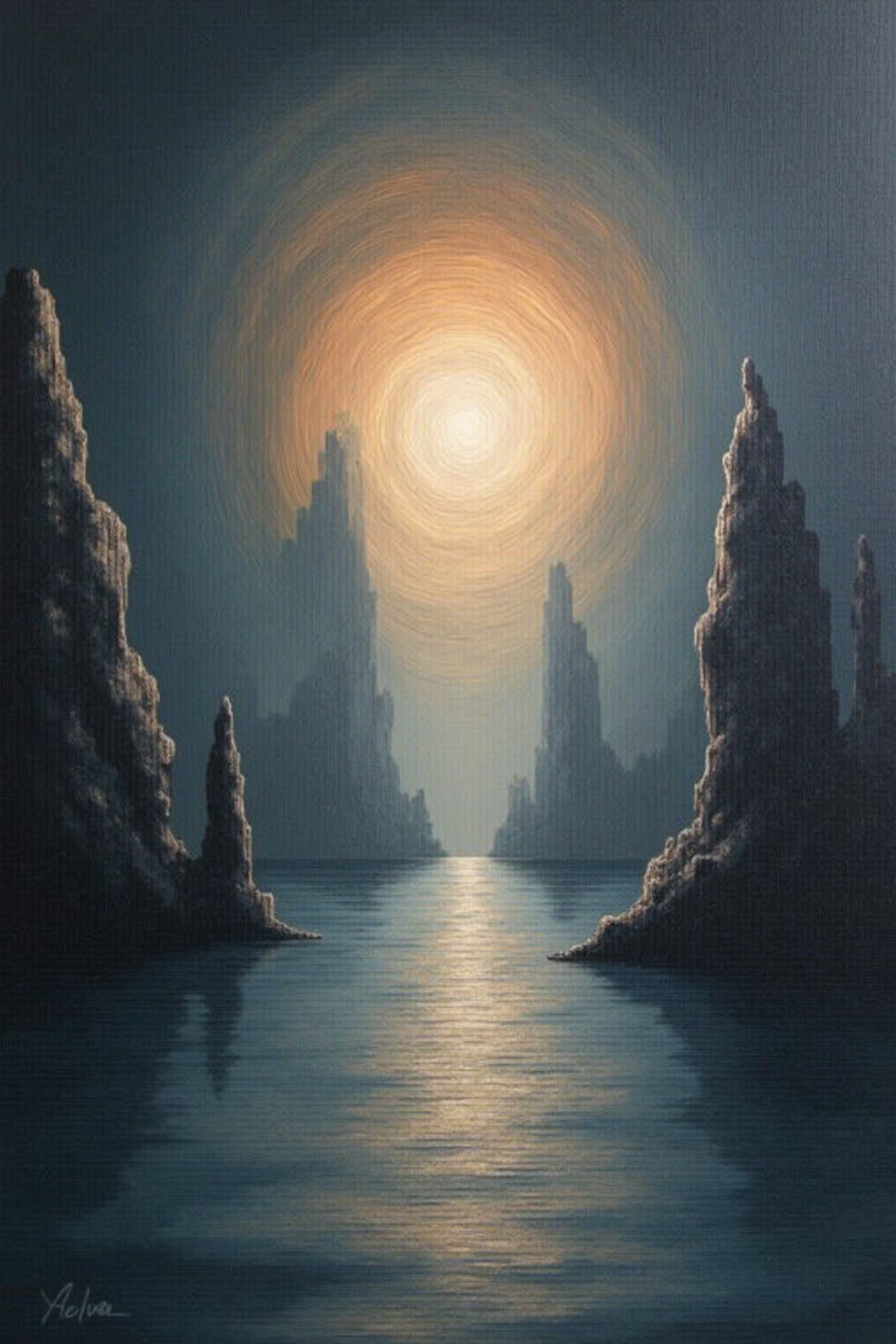}& 
\includegraphics[width=5.3cm]{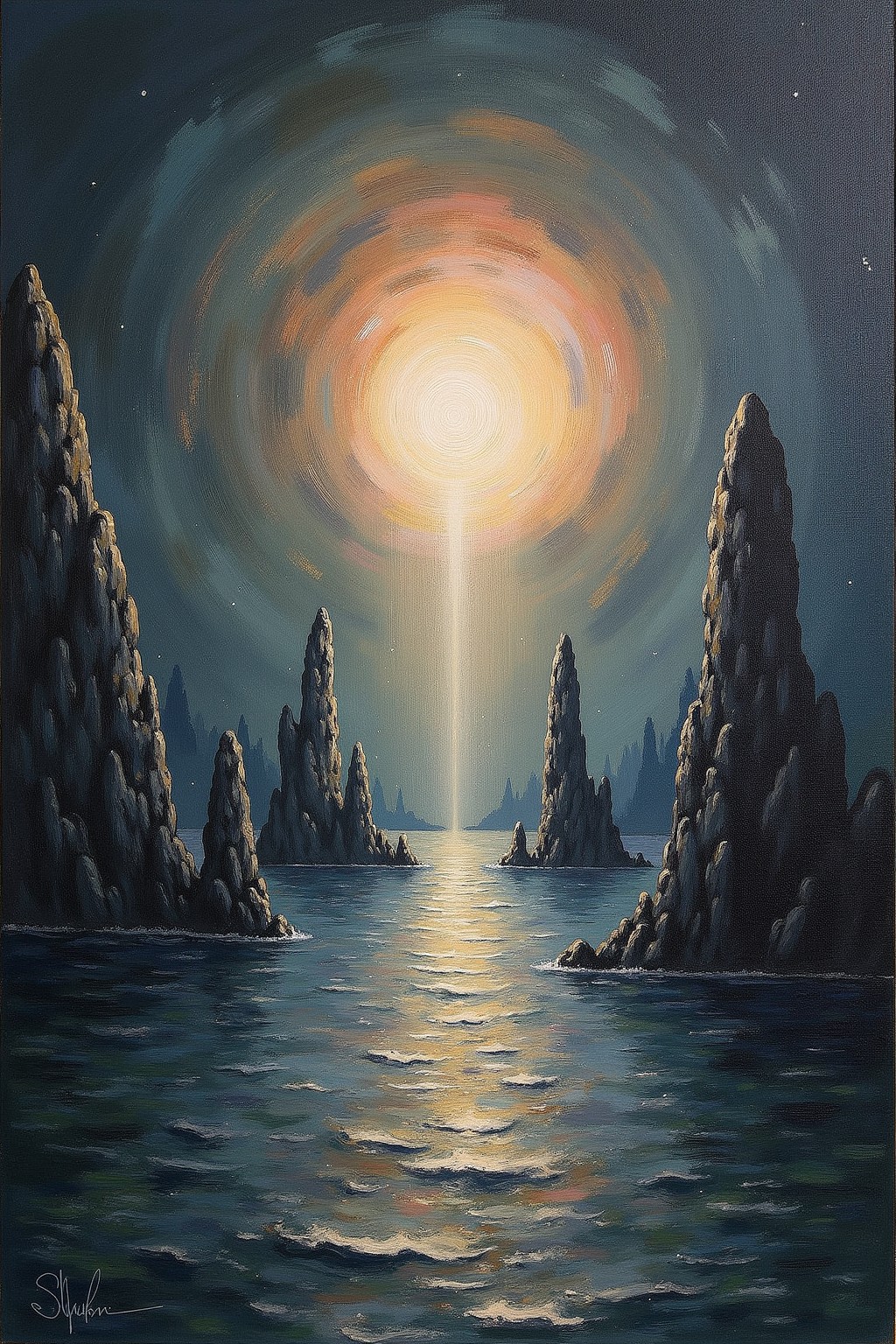}\\
\multicolumn{3}{c}{An oil painting of a latent space.}\\
\bottomrule
\end{tabular*}
\caption{More comparison between different progressive approaches on FLUX.1-dev~{\cite{blackforest2024flux}} with $1024^2$ resolution.}
\label{fig:comparison:FLUX-dev:appe}
\end{figure*}

Fig.~\ref{fig:comparison:FLUX-dev:appe} presents a qualitative comparison of various progressive scaling approaches. We evaluate the baseline FLUX.1-dev~\cite{blackforest2024flux}, the pixel-space approach MegaFusion~\cite{wu2024megafusion}, and our proposed method {\DynaFlow}. The visual results demonstrate that {\DynaFlow} preserves fine-grained details with superior fidelity compared to MegaFusion, which exhibits characteristic blur artifacts inherent to pixel-space scaling transformations. This comparison substantiates the efficacy of latent-space manipulation in preserving high-frequency components during multi-resolution synthesis. The enhanced perceptual quality is particularly evident in complex textures and sharp boundaries, where our approach maintains structural coherence across different resolution scales. These observations align with our quantitative metrics that indicate significant improvements in both computational efficiency and generation quality.
\subsection{Comparative Analysis of LSSGen in Distillated Flow Model}
\label{subsec:appendix:case:flux-schnell}
\begin{figure*}[t]
\centering
\begin{tabular*}{\textwidth}{@{\extracolsep{\fill}}ccc}
\toprule
\centering Prompt & FLUX.1-schnell & LSS-FLUX.1-schnell (ours) \\
\midrule
\rotatebox{90}{\parbox{2.65cm}{\centering A photo of a black hole.}}& 
\includegraphics[width=5.3cm]{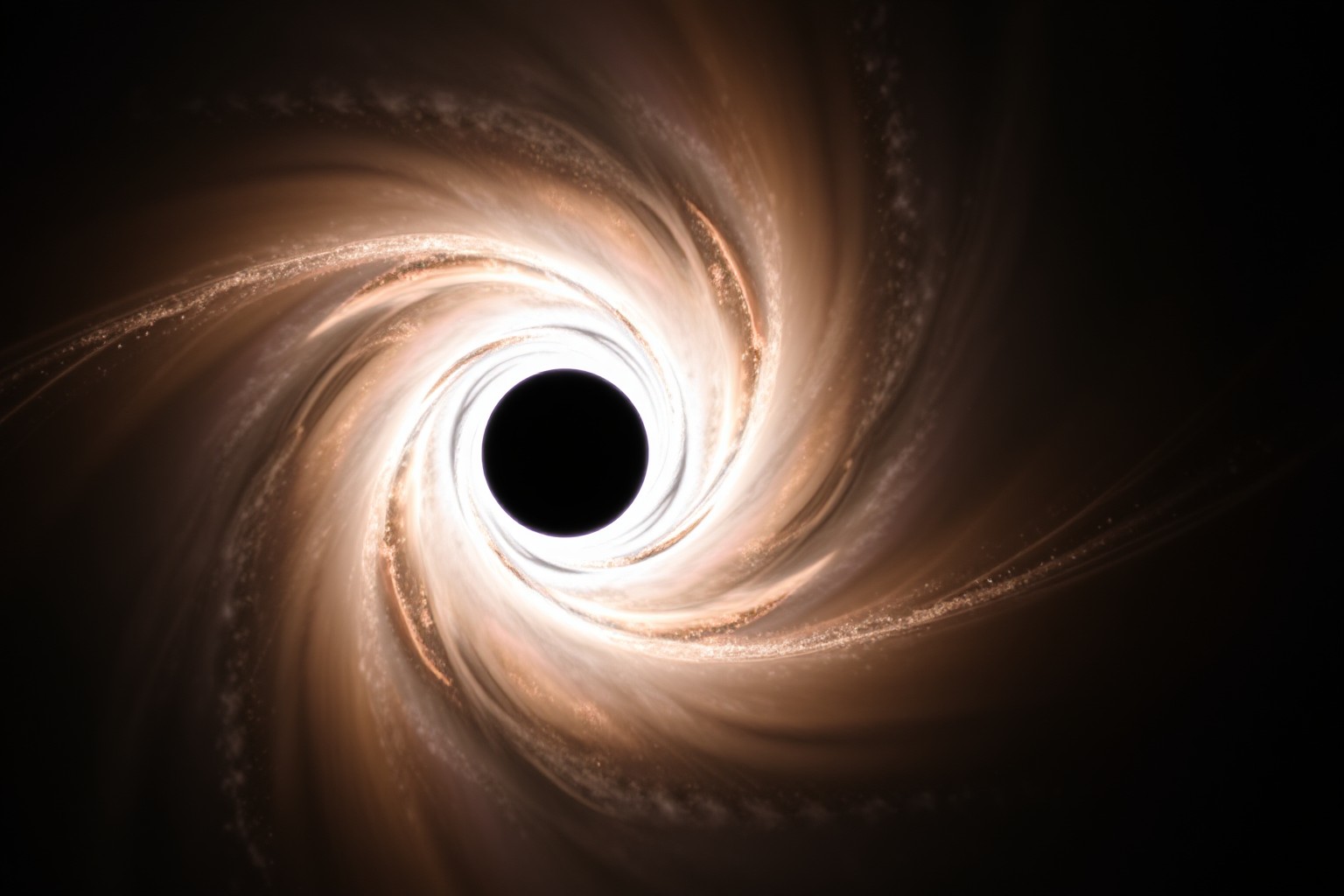}& 
\includegraphics[width=5.3cm]{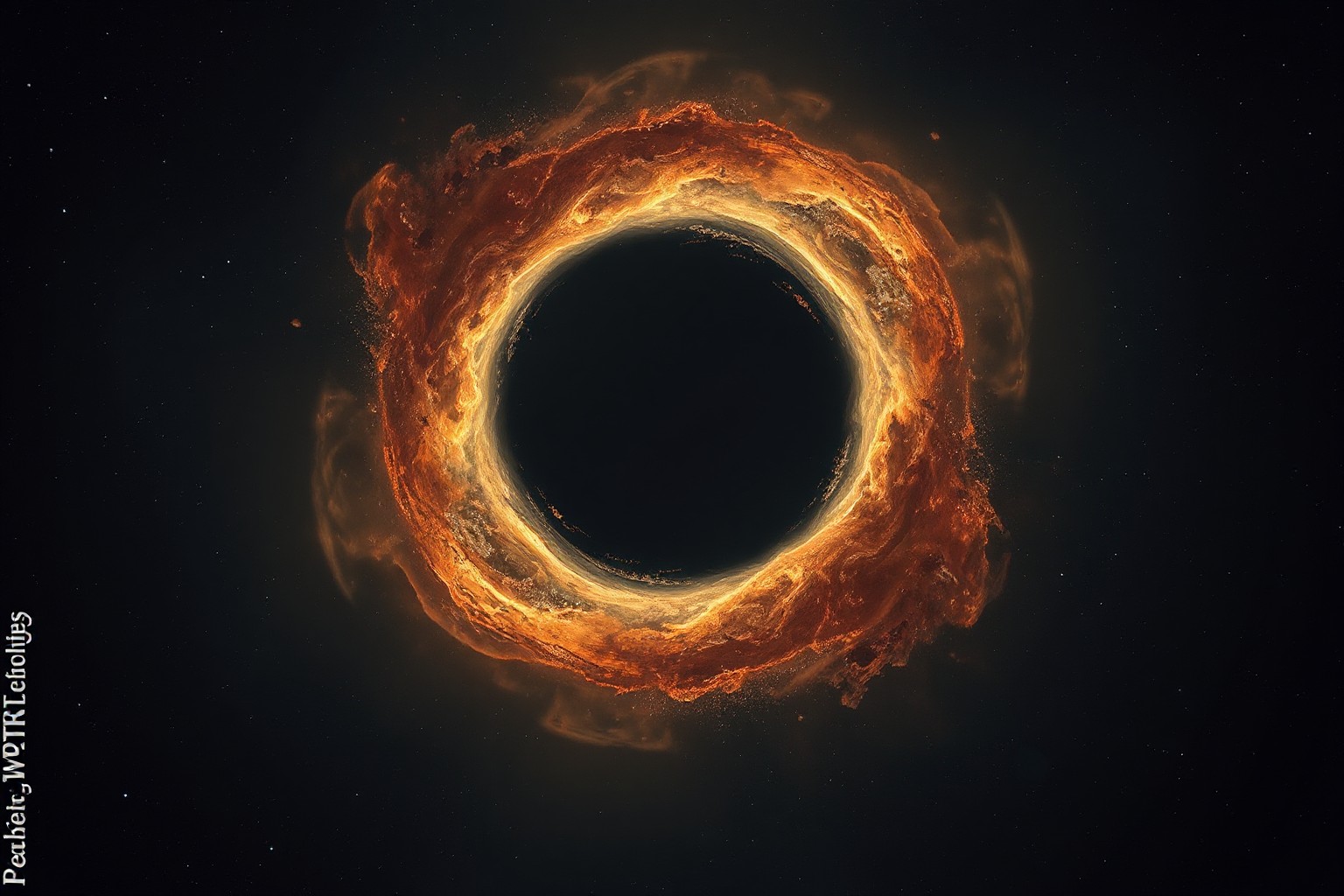}\\ 
\midrule
\rotatebox{90}{\parbox{5.3cm}{\centering On a beautiful snowy mountain top, stands a sign with 'LSSGen' on it.}}&
\includegraphics[width=5.3cm]{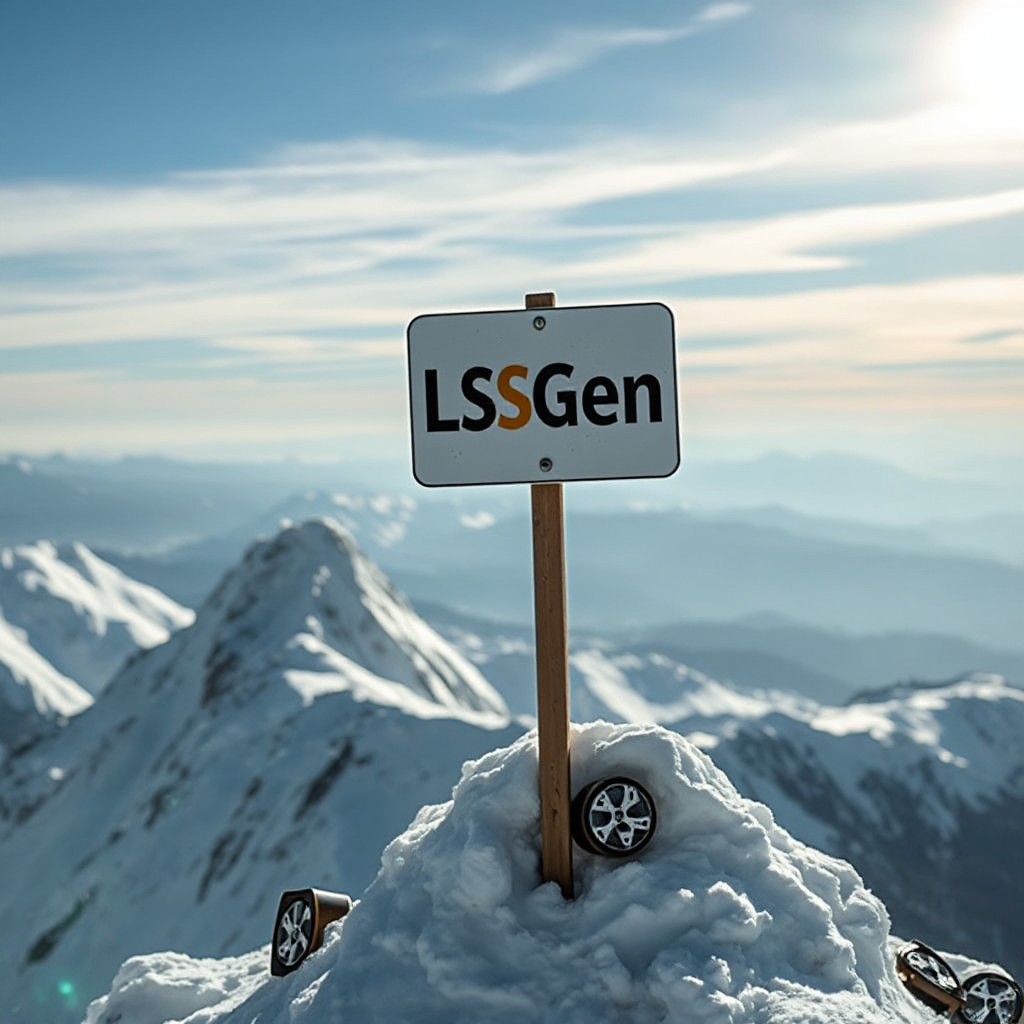}& 
\includegraphics[width=5.3cm]{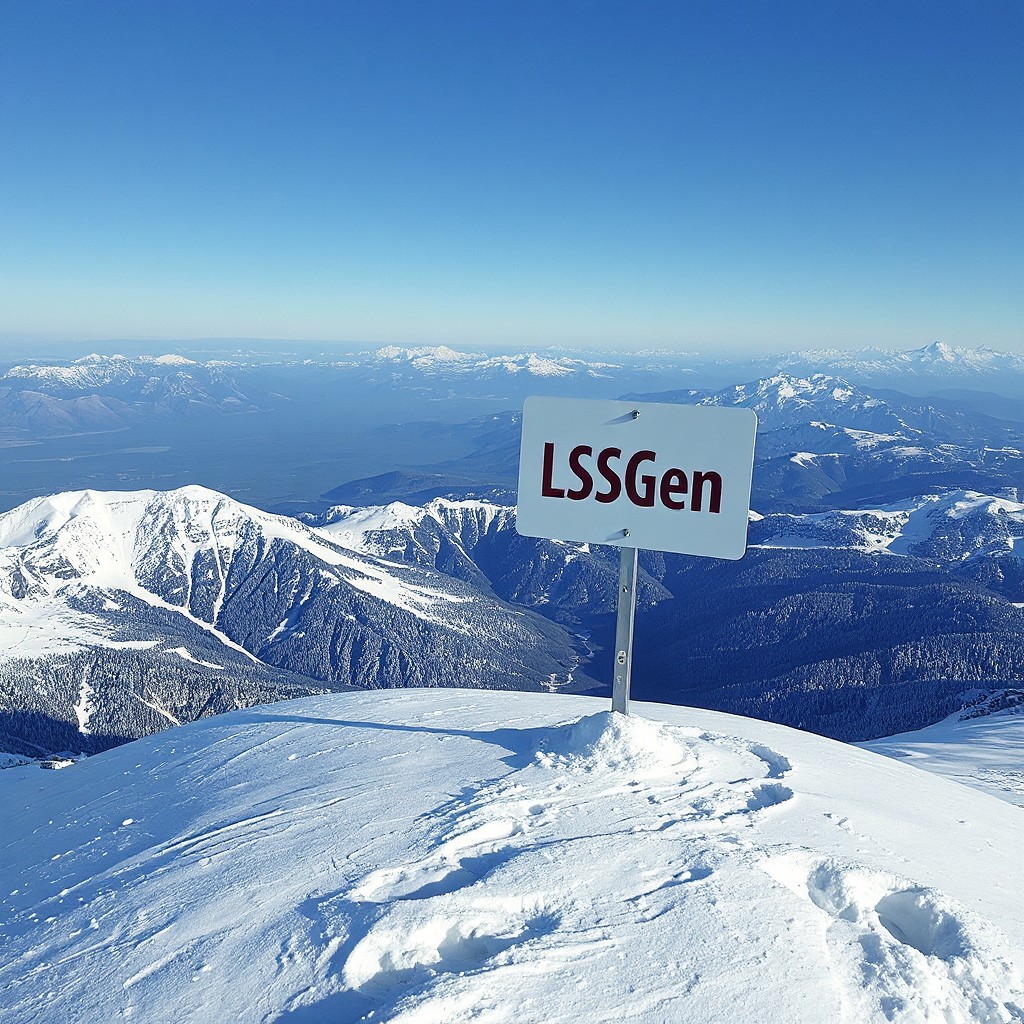}\\ 
\midrule
\rotatebox{90}{\parbox{5.3cm}{\centering A beautiful lady holding a sign saying 'LSSGen'.}}&
{\includegraphics[width=5.3cm]{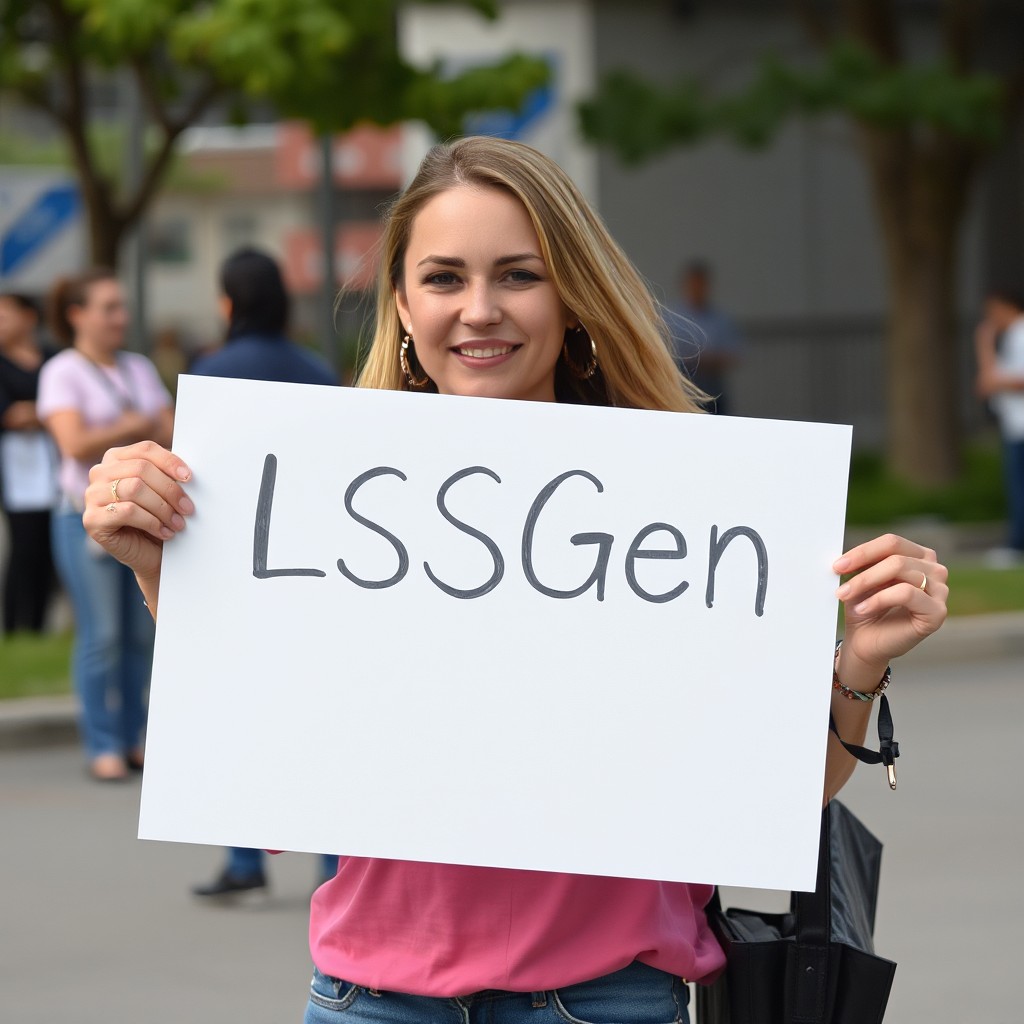}}& 
{\includegraphics[width=5.3cm]{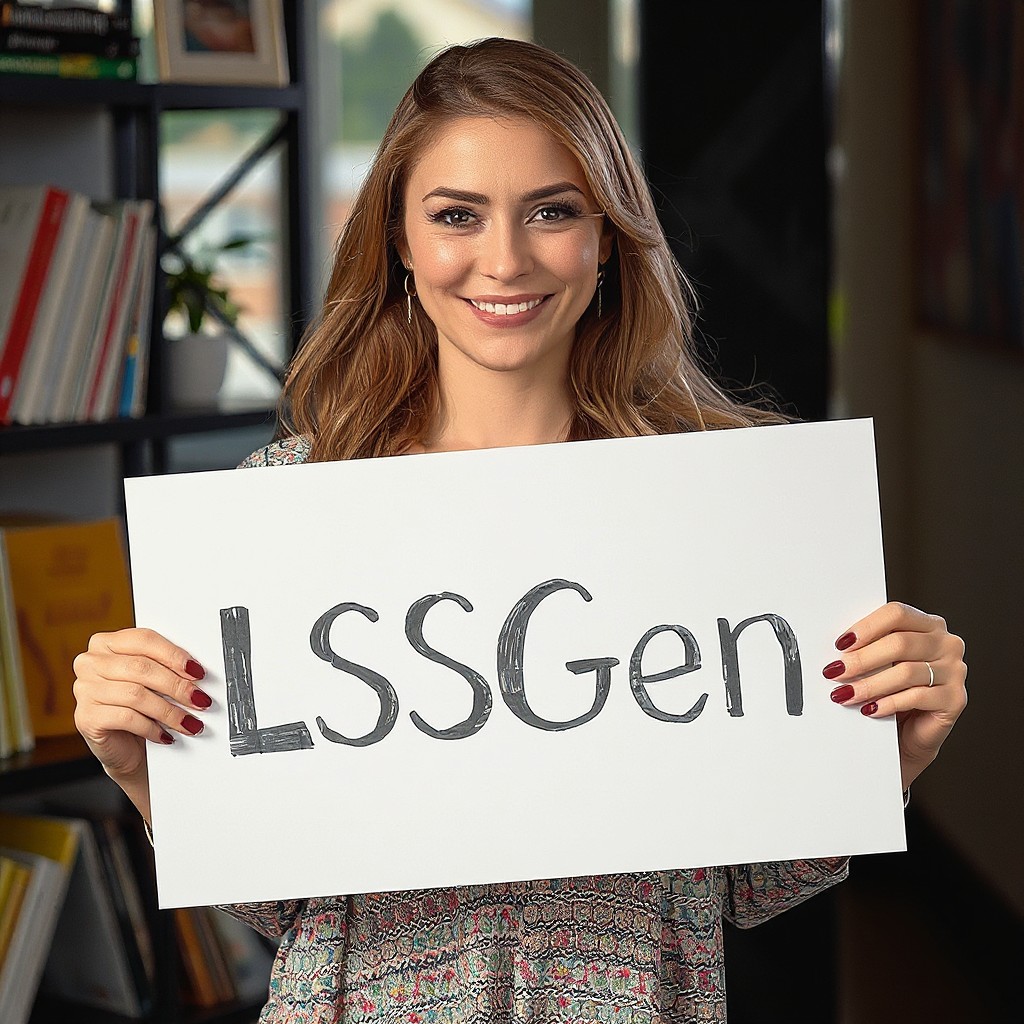}}\\
\midrule
\rotatebox{90}{\parbox{5.3cm}{\centering A giant rocket launching from a huge chocolate cake.}}&
\includegraphics[width=5.3cm]{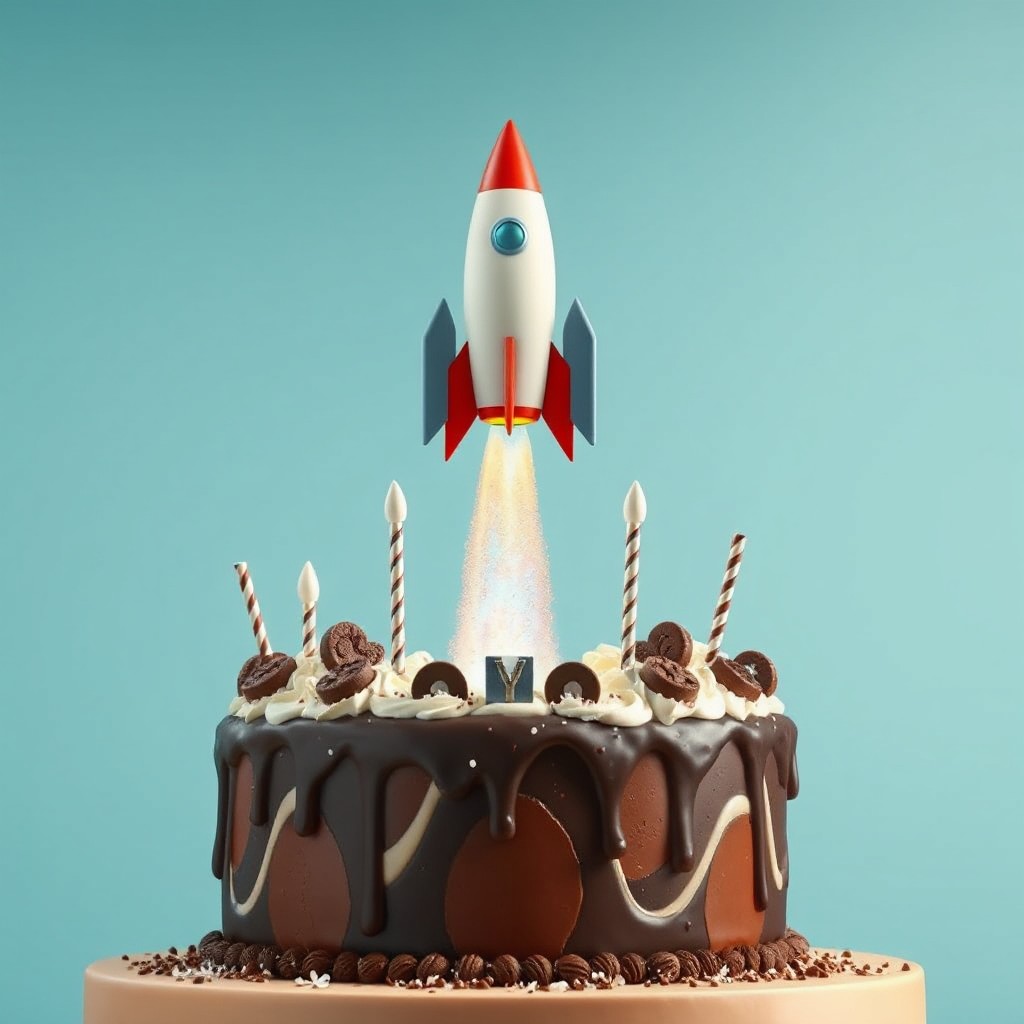}& 
\includegraphics[width=5.3cm]{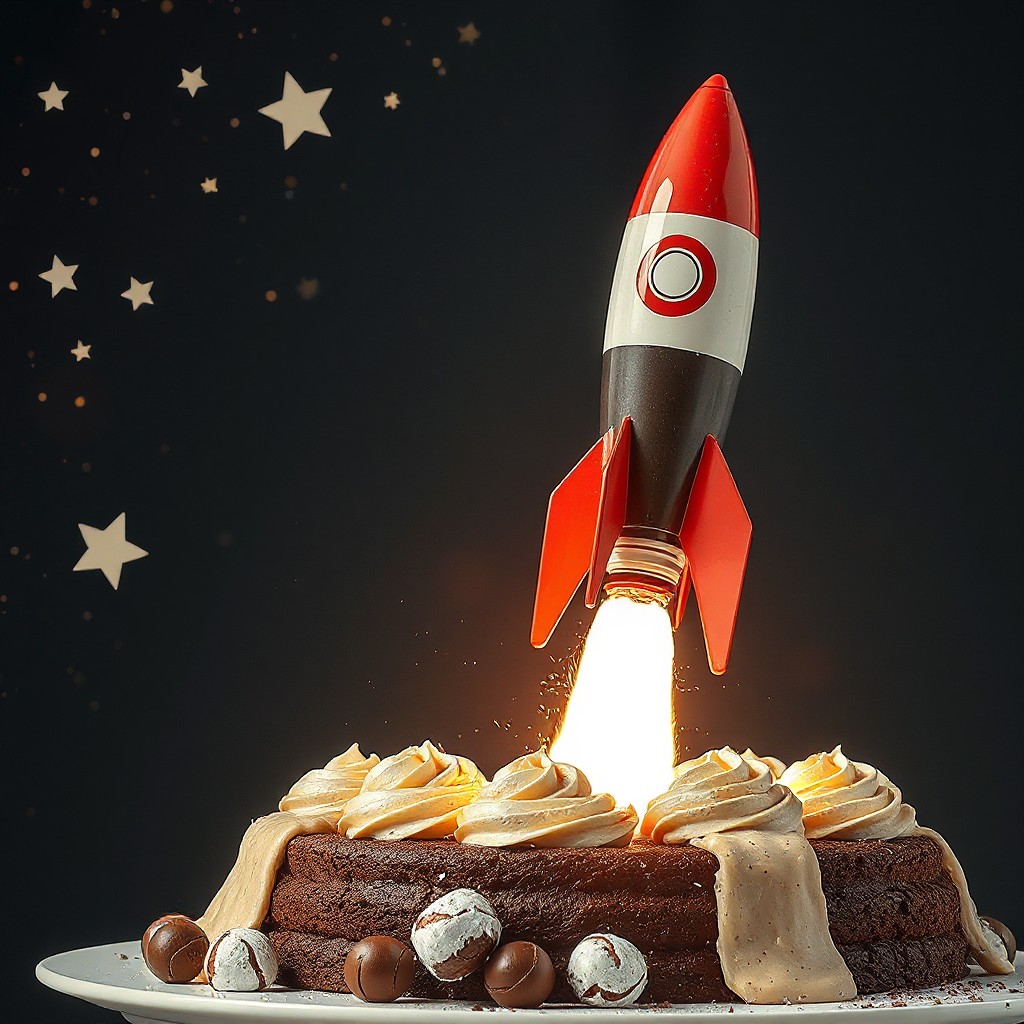}\\ 
\bottomrule
\end{tabular*}
\caption{More comparison between different progressive approaches on FLUX.1-schnell~{\cite{blackforest2024flux}} with $1024^2$ resolution.}
\label{fig:comparison:FLUX-schnell:appe}
\end{figure*}

Fig.~\ref{fig:comparison:FLUX-schnell:appe} presents a quantitative evaluation contrasting the non-scaling baseline approach with our proposed {\DynaFlow} framework. We systematically compare FLUX.1-schnell~\cite{blackforest2024flux} against {\DynaFlow} across multiple generation tasks. The empirical results demonstrate that {\DynaFlow} consistently produces outputs with enhanced perceptual quality characterized by superior detail preservation and edge definition. This quality enhancement stems from our latent space scaling operations that maintain high-frequency information through the generation process.
The latent-space manipulations introduce precisely controlled detail amplification that manifests as enhanced definition in textural elements. This characteristic proves advantageous for most generative tasks, though practitioners should note potential over-sharpening effects when synthesizing human facial features, where the enhanced detail reproduction may accentuate fine wrinkles beyond natural appearance. Most significantly, our progressive upsampling methodology achieves a substantial inference time reduction from 4.77s to 4.36s (an 1.1x acceleration) while maintaining generation quality integrity. This performance optimization exemplifies the efficiency even further with timestep-distilled models, indicating the broadness of our methods. The demonstrated compatibility with model distillation techniques underscores the generalizability of our latent space transformations across varied architectural configurations and computational constraints.
\subsection{Comparative Analysis of LSSGen in Diffusion Models}
\label{subsec:appendix:case:sdxl}
\begin{figure*}[t]
\centering
\setlength{\tabcolsep}{3pt}
\begin{tabular*}{\textwidth}{@{\extracolsep{\fill}}cccccc}
\toprule
Prompt & SDXL & SDXL-Self-Cascade & SDXL-MegaFusion & SDXL-DiffuseHigh & LSS-SDXL (ours) \\
\midrule
\parbox[c]{1.5cm}{\centering{a photo of a cake below a baseball bat}} &
\begin{tabular}{@{}c@{}}\includegraphics[width=2.7cm]{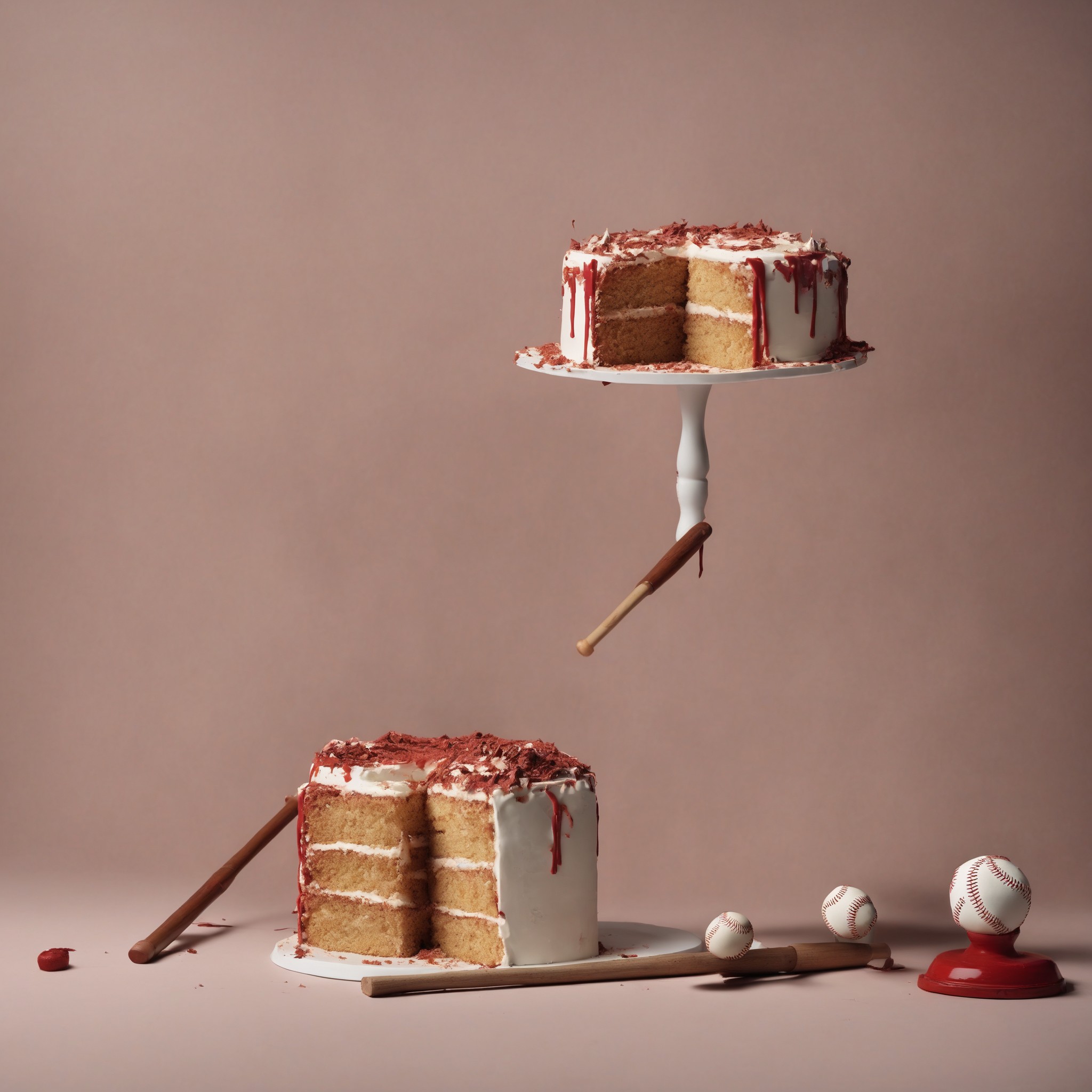} \\ \includegraphics[width=2.7cm]{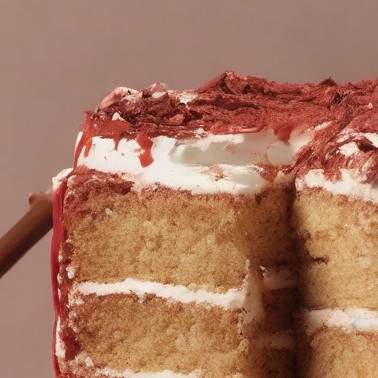} \end{tabular}& 
\begin{tabular}{@{}c@{}}\includegraphics[width=2.7cm]{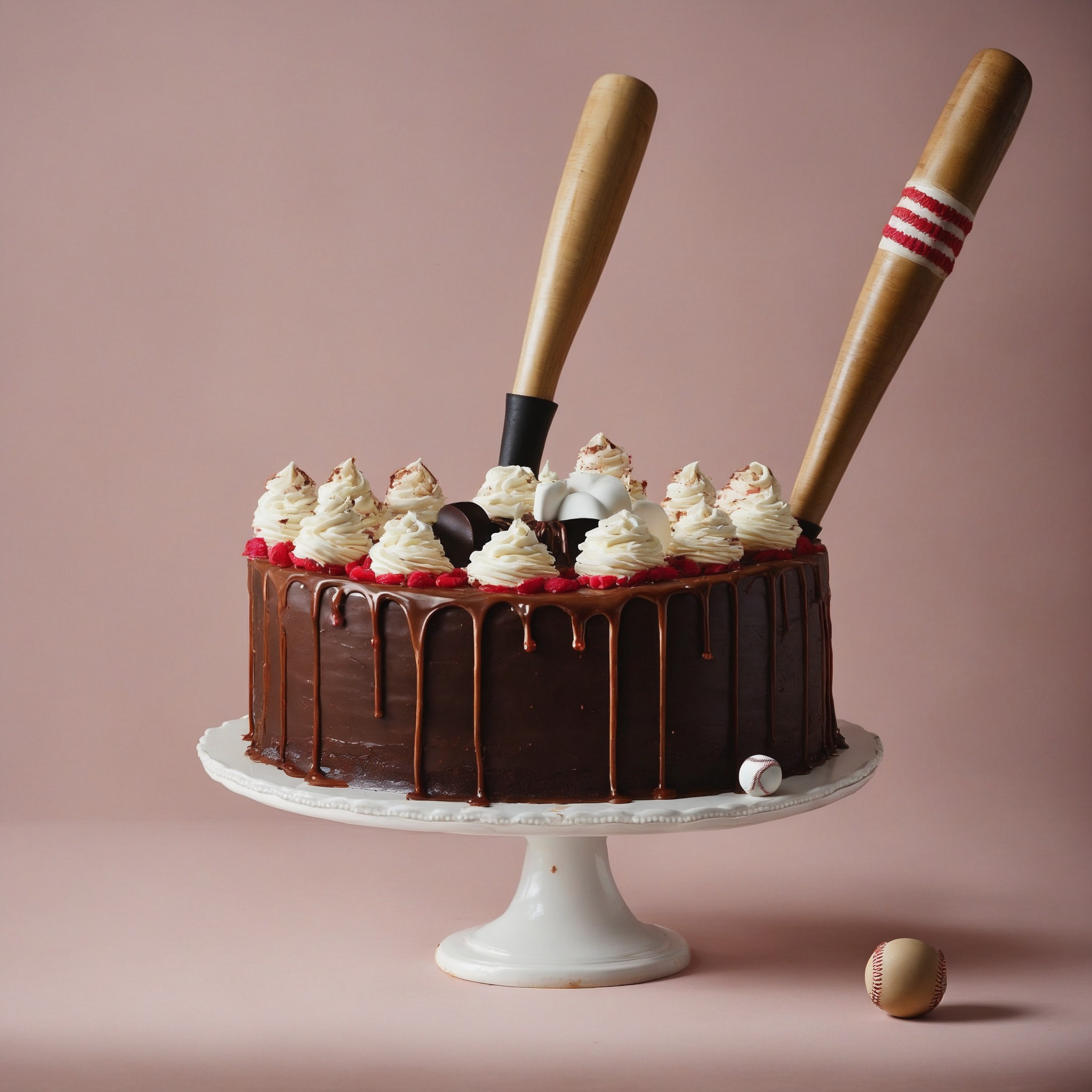}\\ \includegraphics[width=2.7cm]{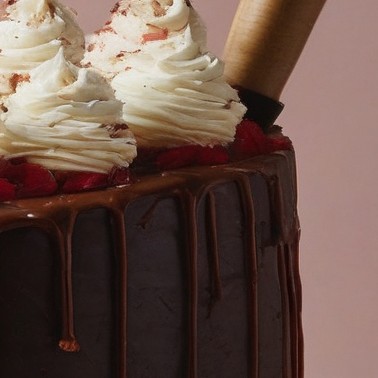}  \end{tabular}&
\begin{tabular}{@{}c@{}}\includegraphics[width=2.7cm]{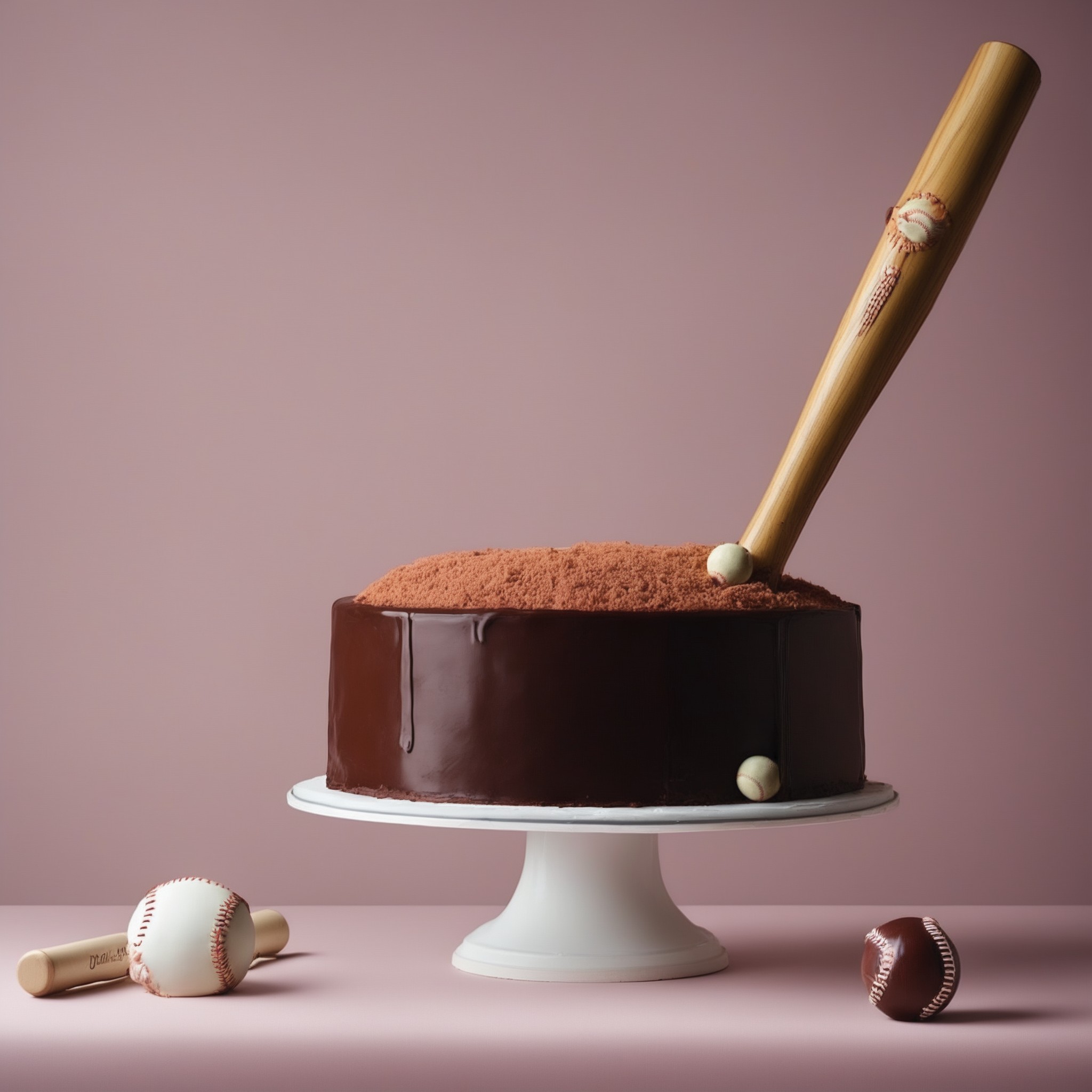}\\ \includegraphics[width=2.7cm]{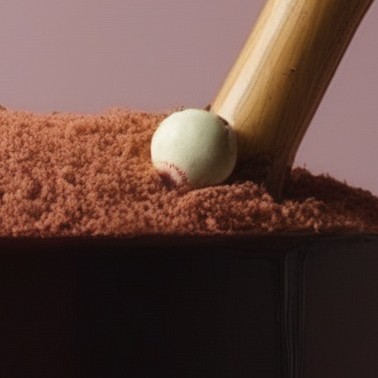}  \end{tabular}& 
\begin{tabular}{@{}c@{}}\includegraphics[width=2.7cm]{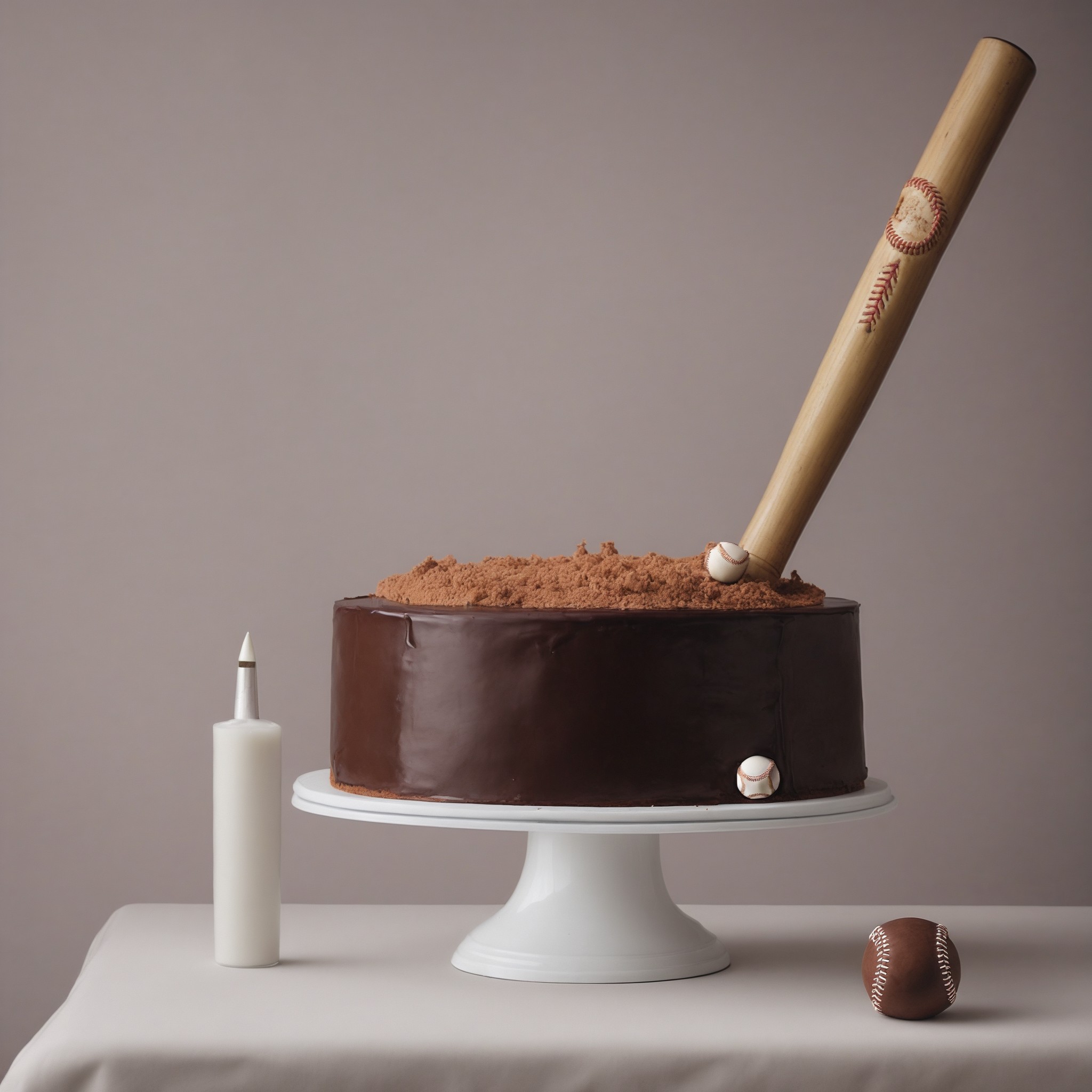}\\ \includegraphics[width=2.7cm]{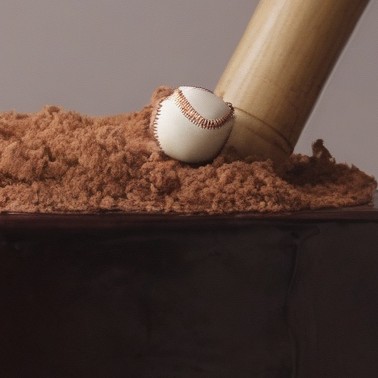}  \end{tabular}& 
\begin{tabular}{@{}c@{}}\includegraphics[width=2.7cm]{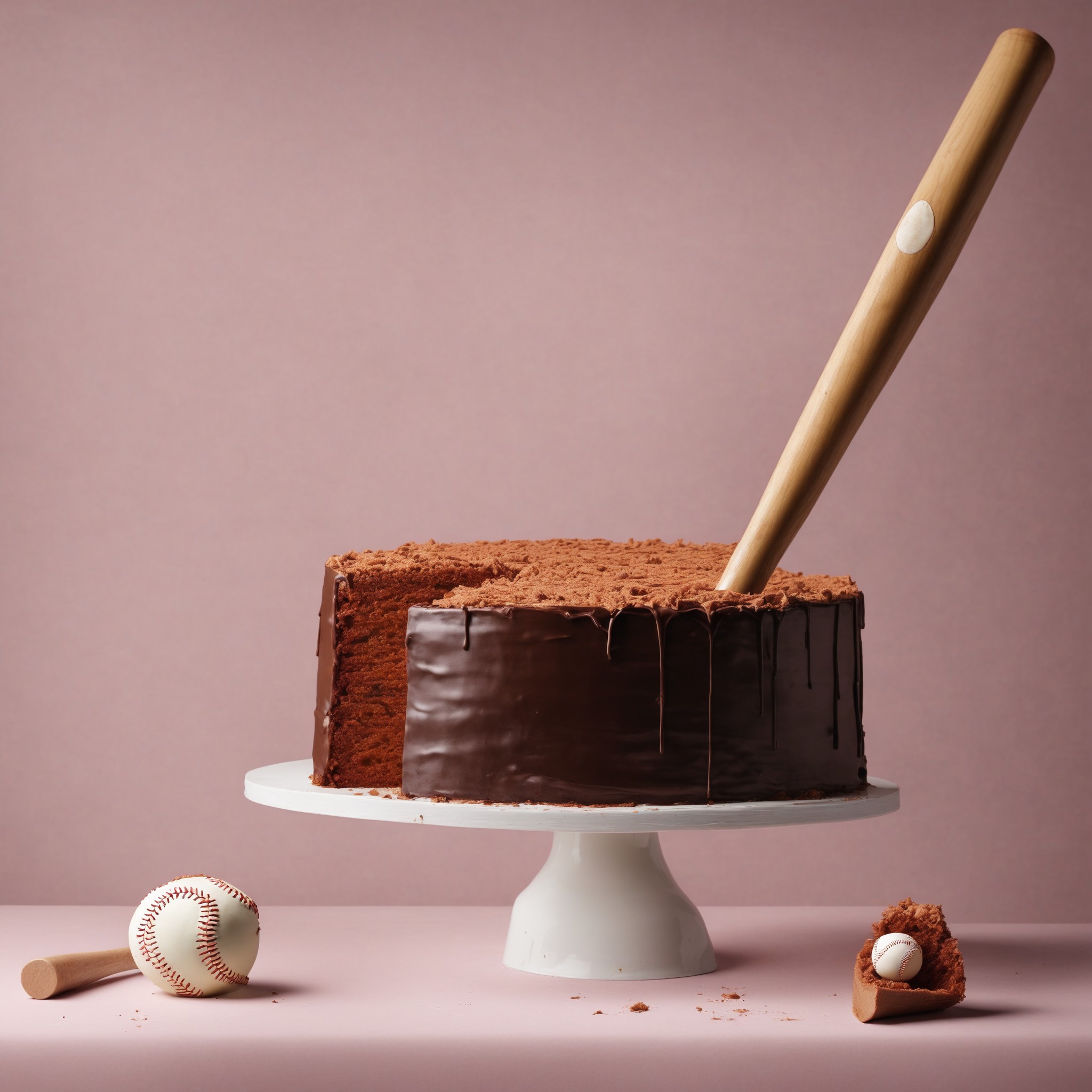}\\ \includegraphics[width=2.7cm]{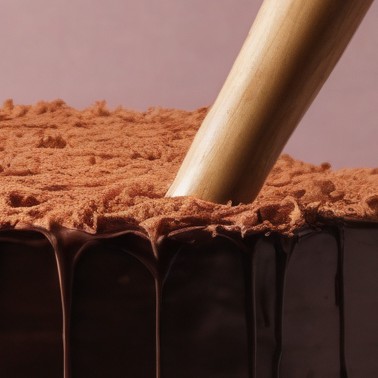}  \end{tabular}\\ 
\midrule
\parbox[c]{1.5cm}{\centering{a photo of an apple and a toothbrush}} &
\begin{tabular}{@{}c@{}}\includegraphics[width=2.7cm]{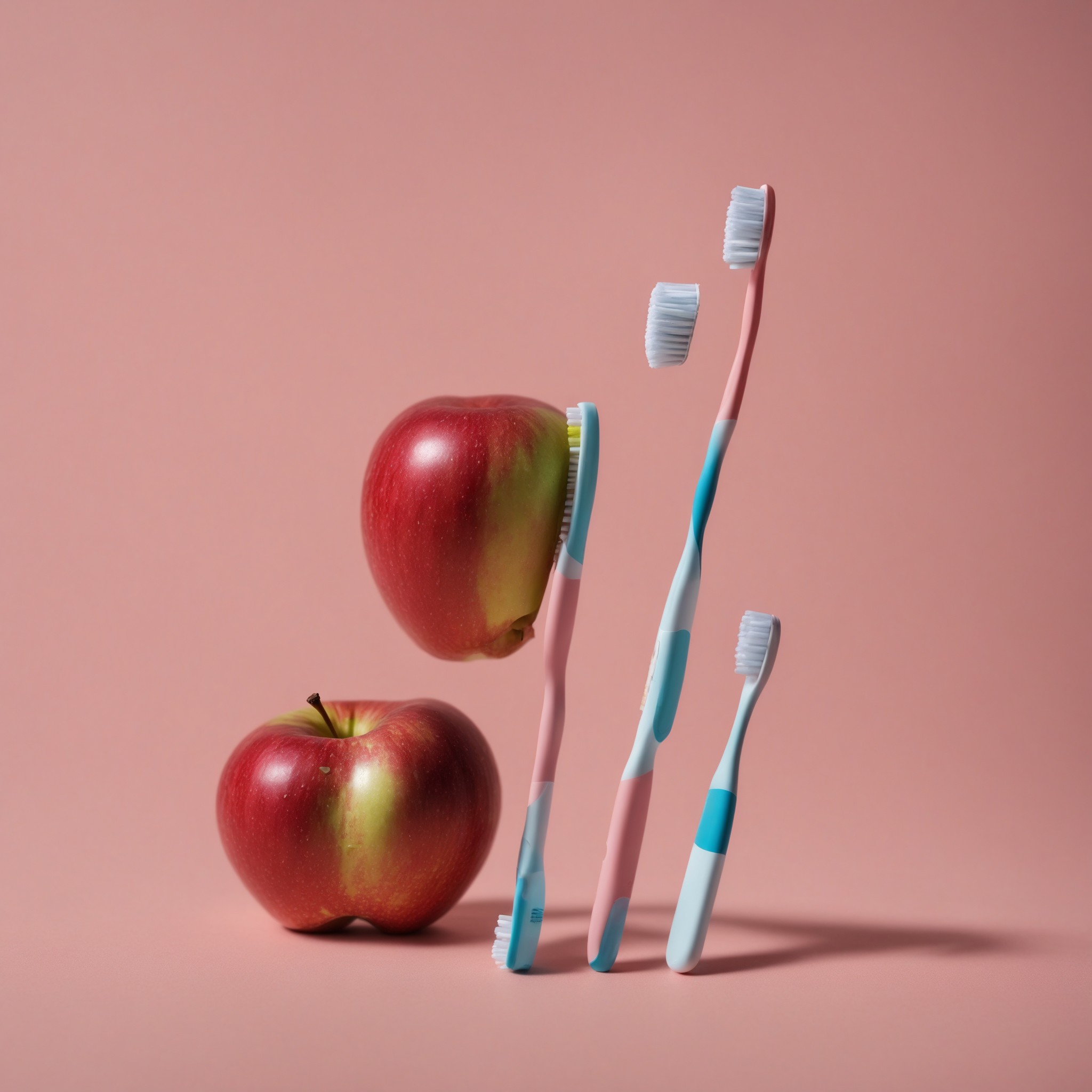} \\ \includegraphics[width=2.7cm]{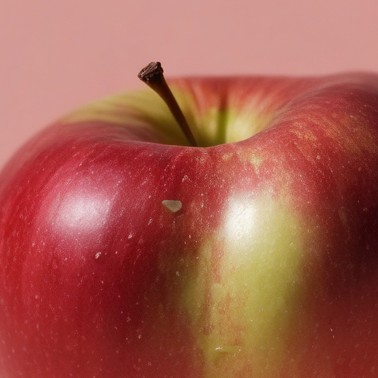} \end{tabular}& 
\begin{tabular}{@{}c@{}}\includegraphics[width=2.7cm]{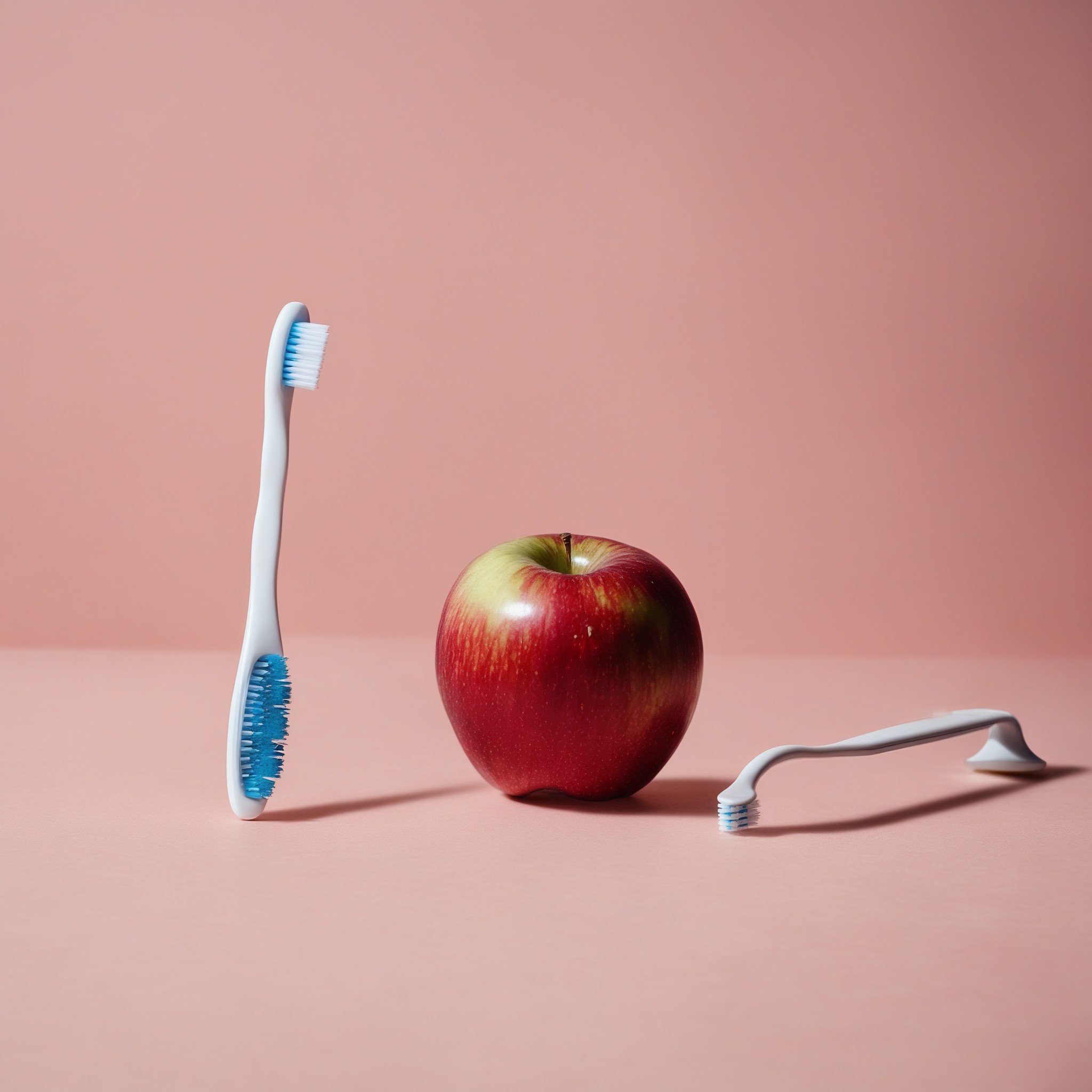}\\ \includegraphics[width=2.7cm]{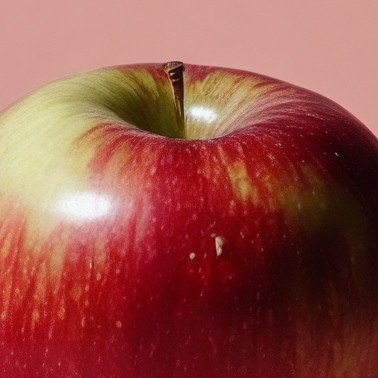}  \end{tabular}&
\begin{tabular}{@{}c@{}}\includegraphics[width=2.7cm]{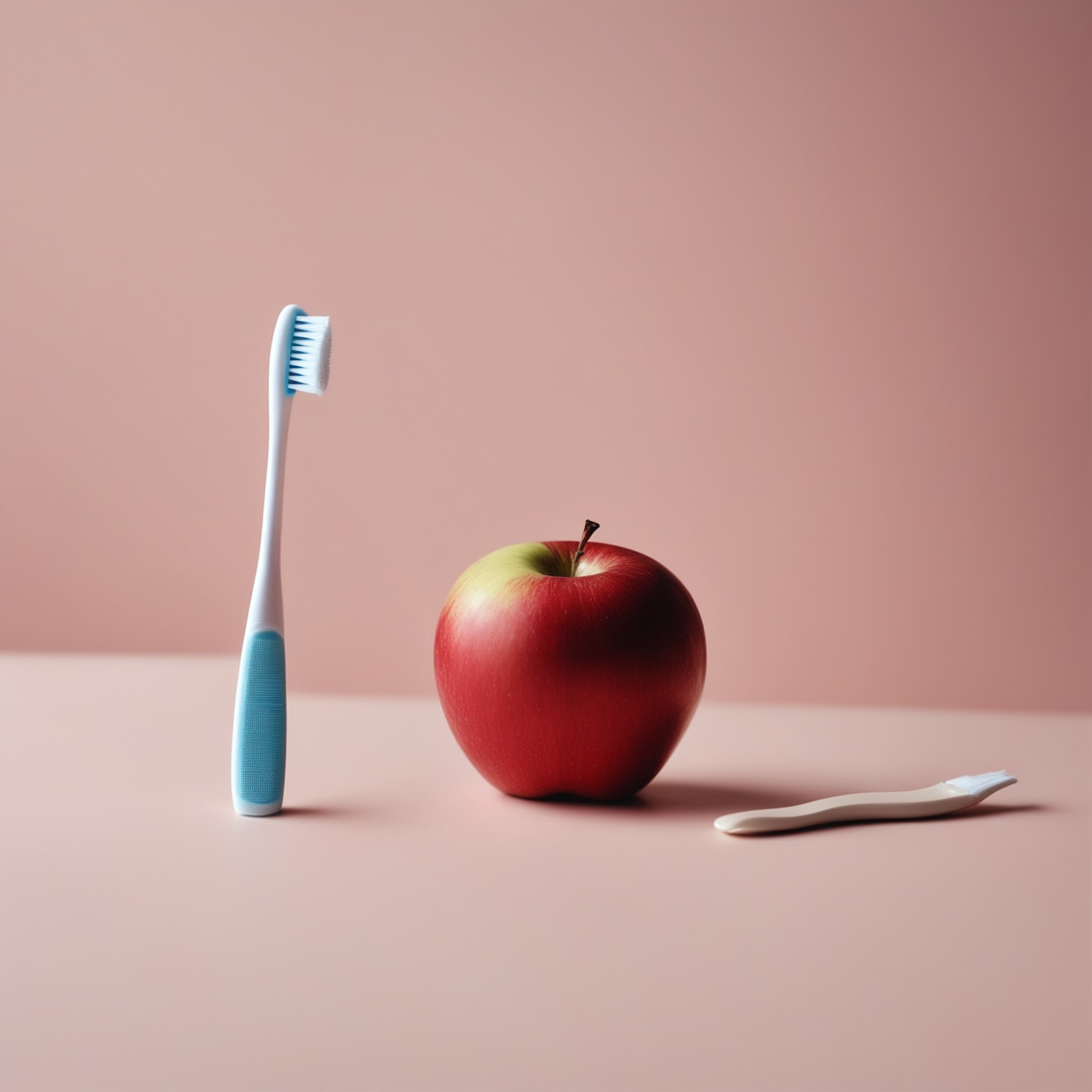}\\ \includegraphics[width=2.7cm]{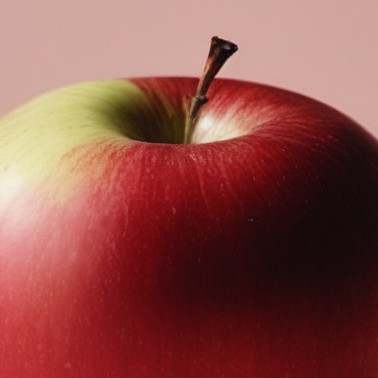}  \end{tabular}& 
\begin{tabular}{@{}c@{}}\includegraphics[width=2.7cm]{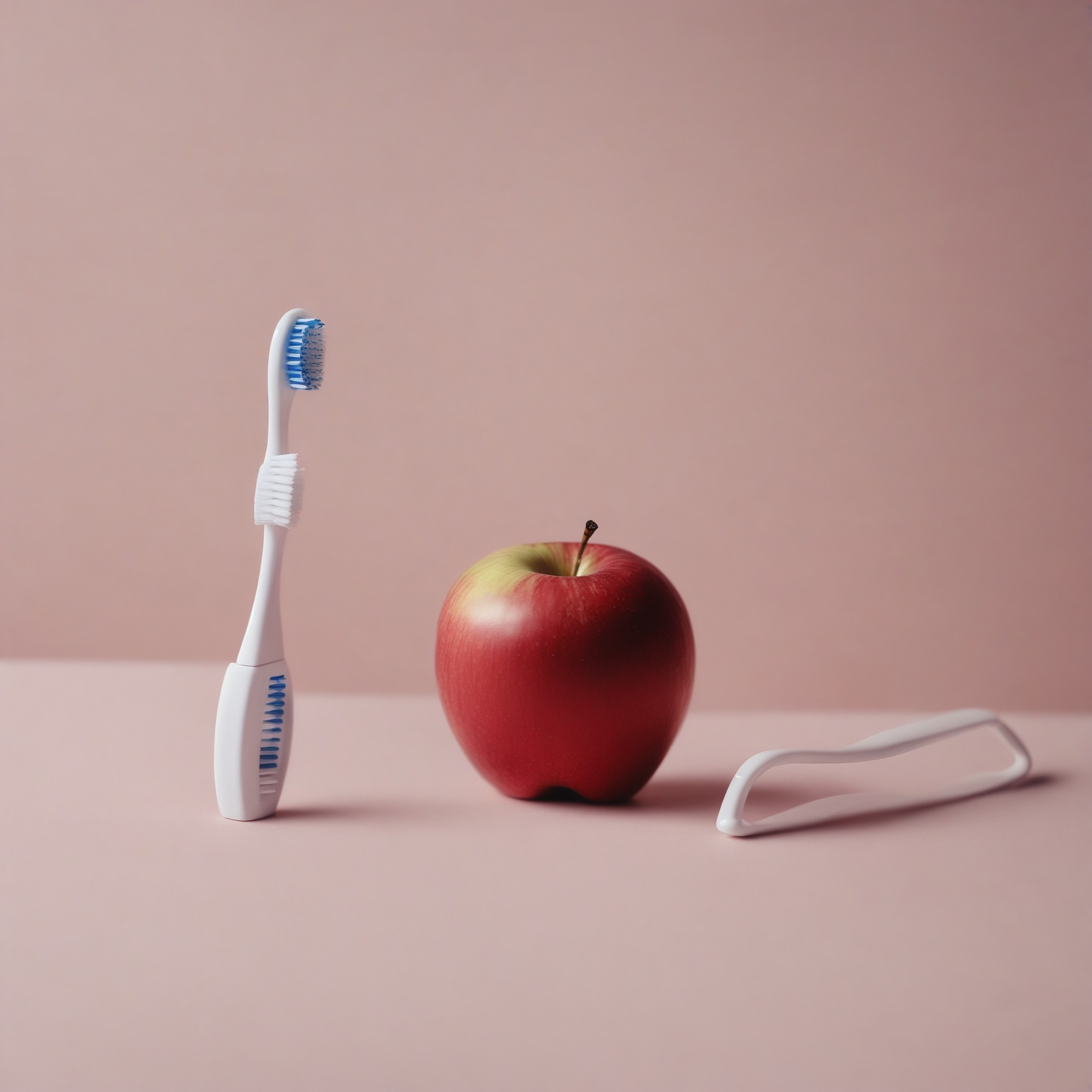}\\ \includegraphics[width=2.7cm]{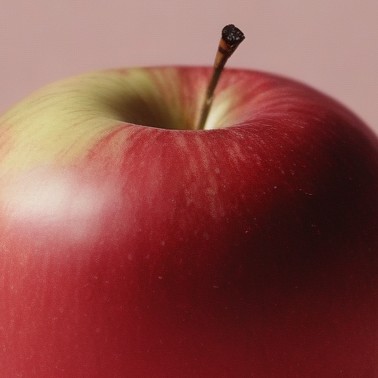}  \end{tabular}& 
\begin{tabular}{@{}c@{}}\includegraphics[width=2.7cm]{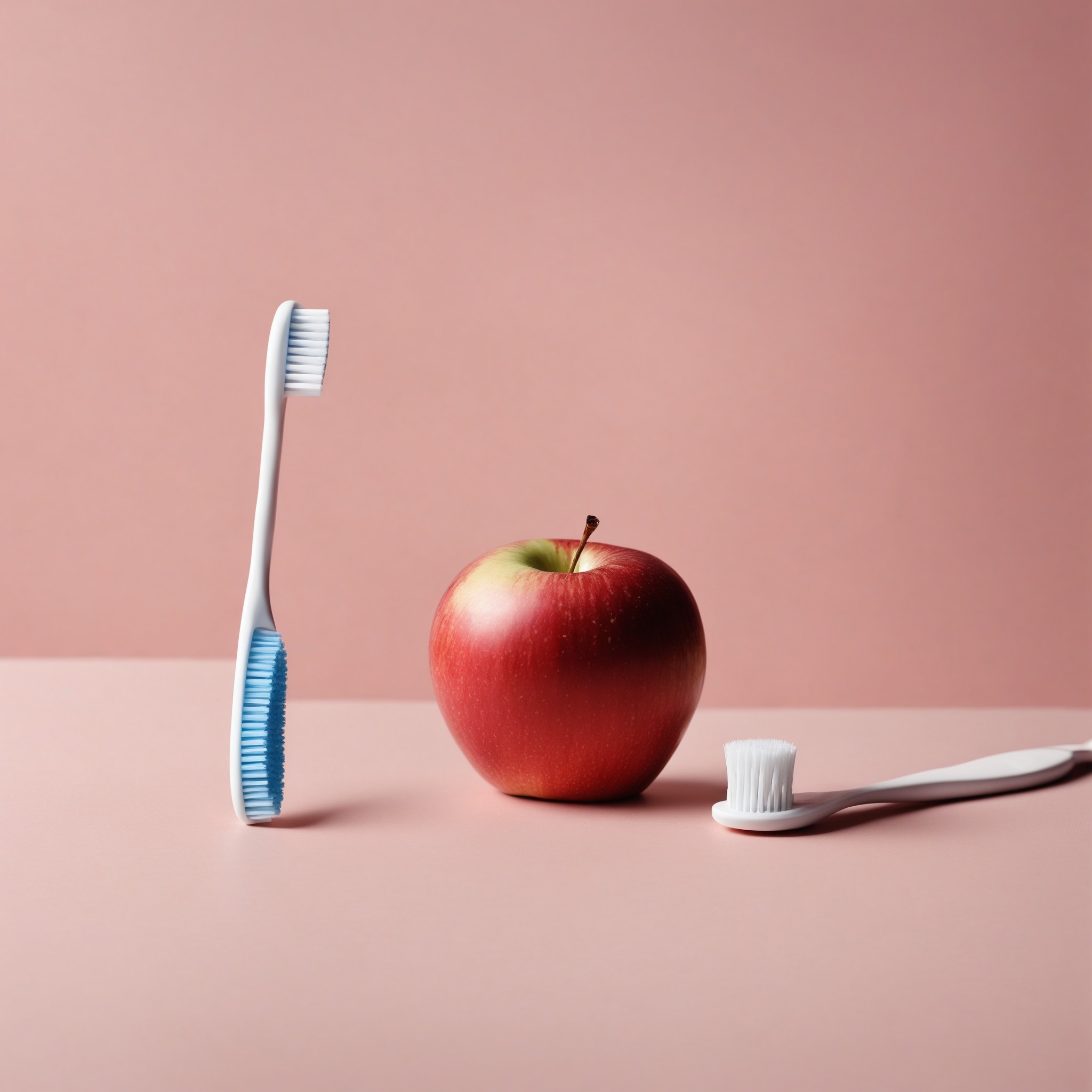}\\ \includegraphics[width=2.7cm]{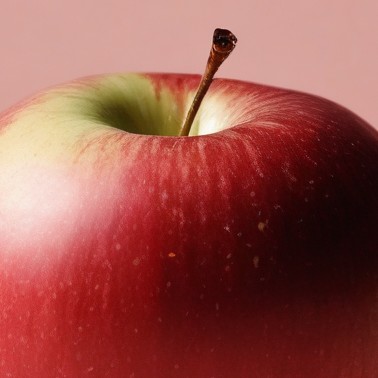}  \end{tabular}\\ 
\midrule
\parbox[c]{1.5cm}{\centering{a photo of a red umbrella and a green cow}} &
\begin{tabular}{@{}c@{}}\includegraphics[width=2.7cm]{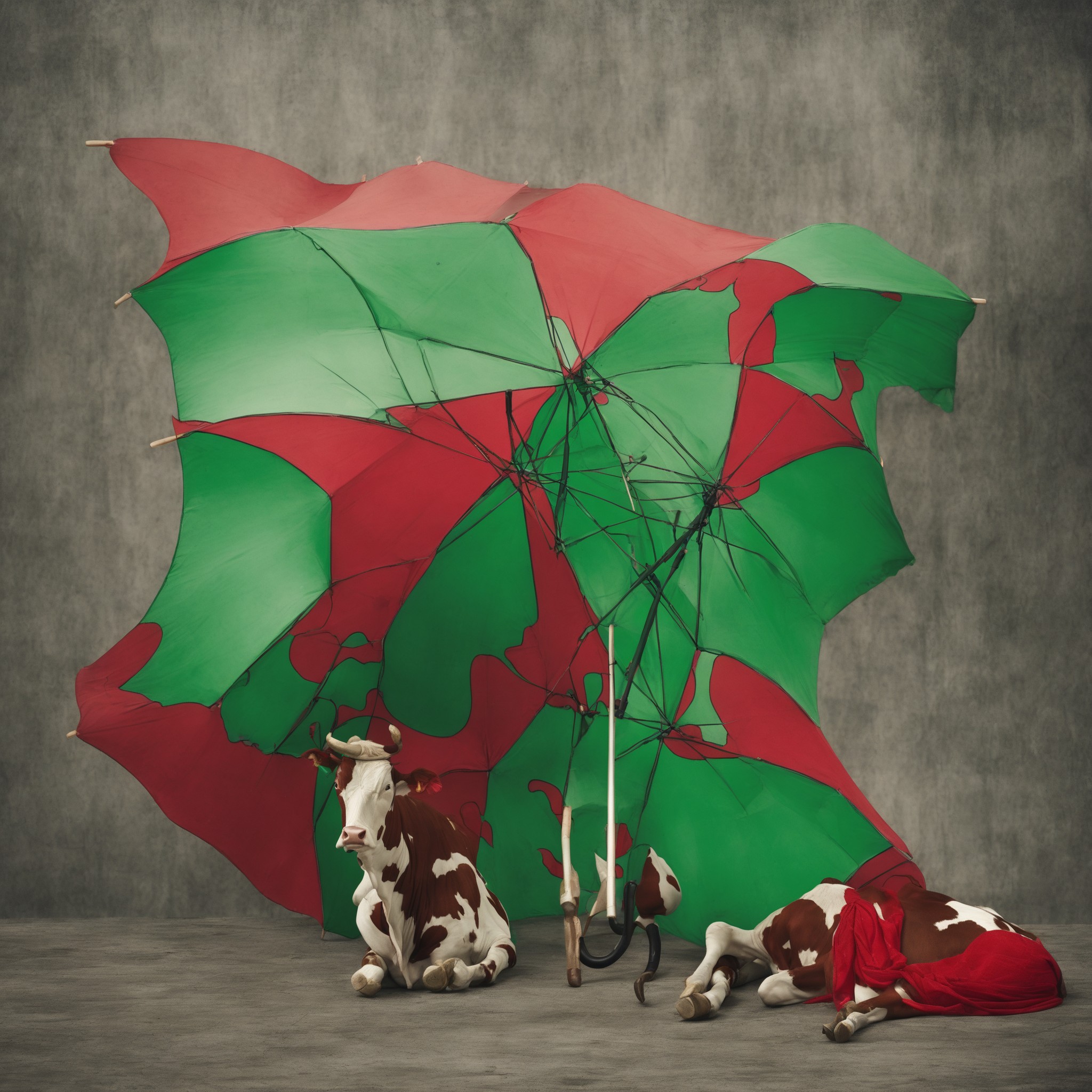} \\ \includegraphics[width=2.7cm]{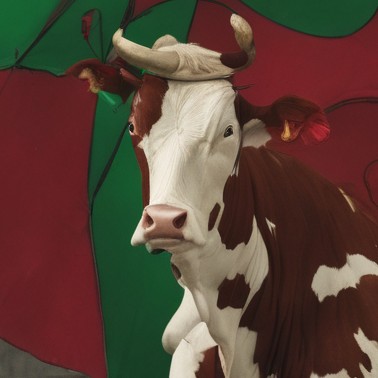} \end{tabular}& 
\begin{tabular}{@{}c@{}}\includegraphics[width=2.7cm]{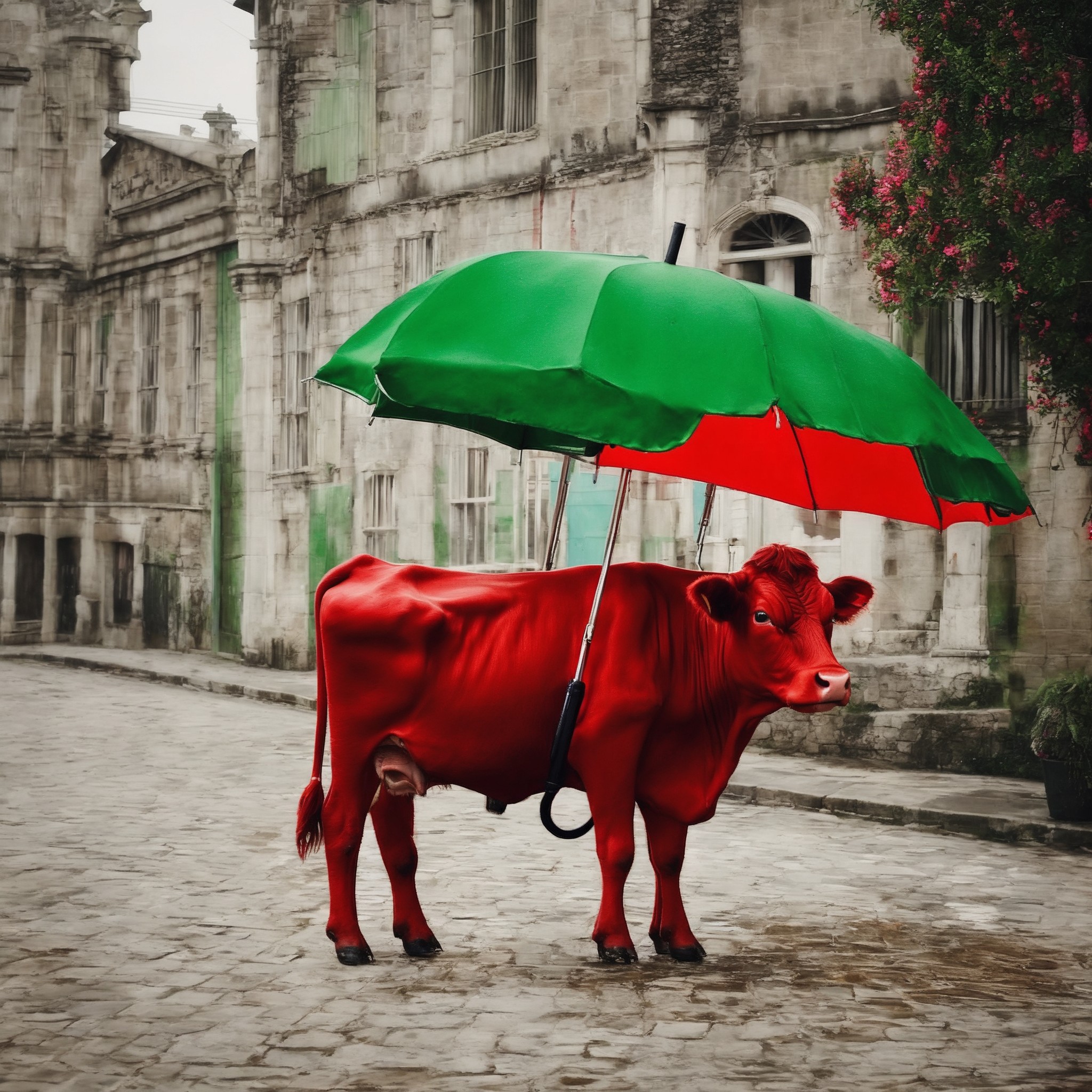}\\ \includegraphics[width=2.7cm]{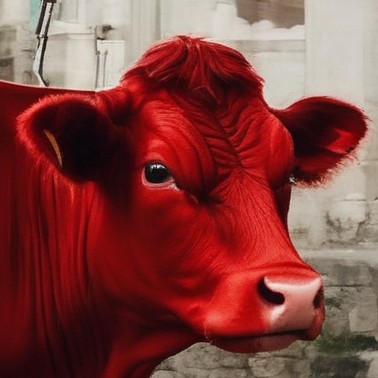}  \end{tabular}&
\begin{tabular}{@{}c@{}}\includegraphics[width=2.7cm]{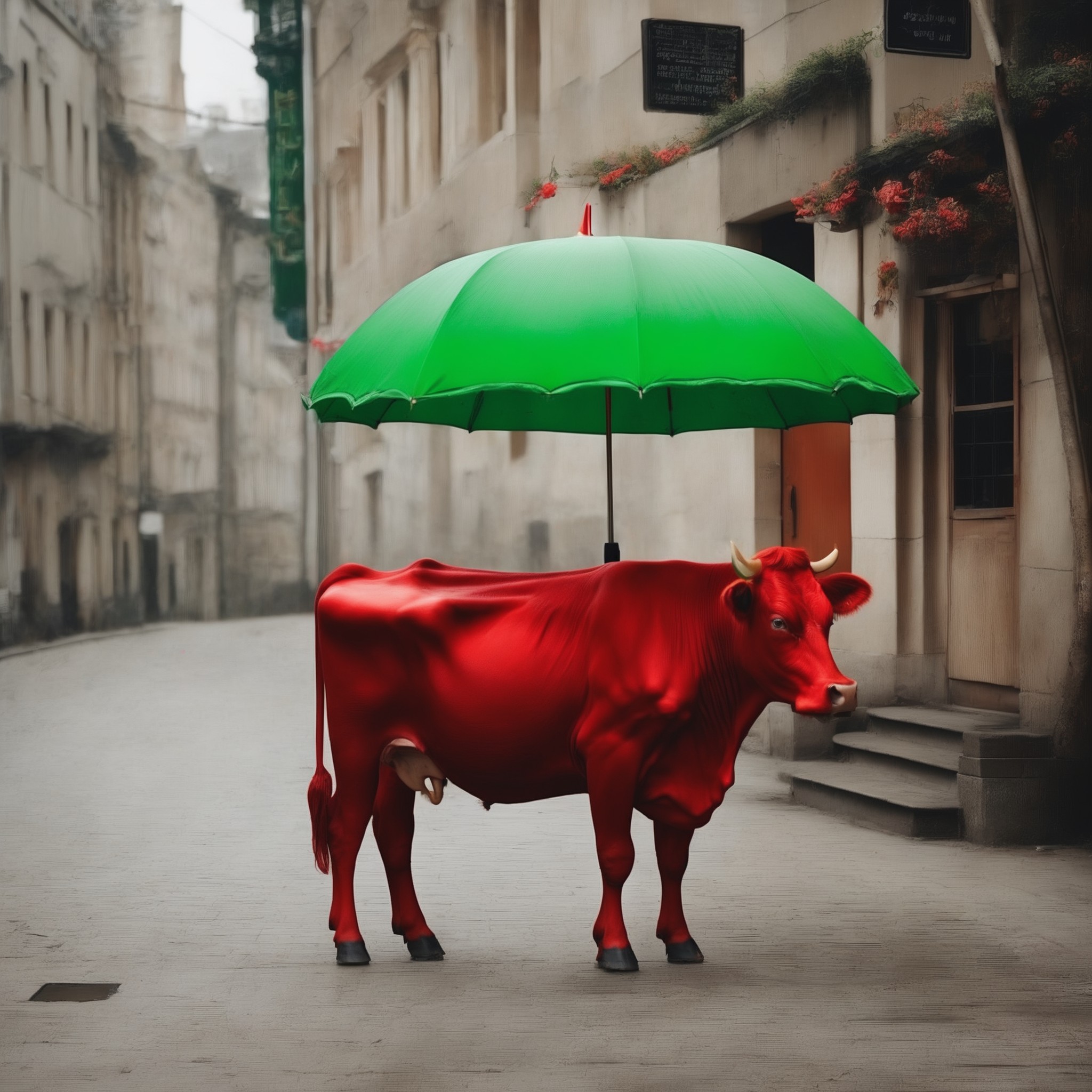}\\ \includegraphics[width=2.7cm]{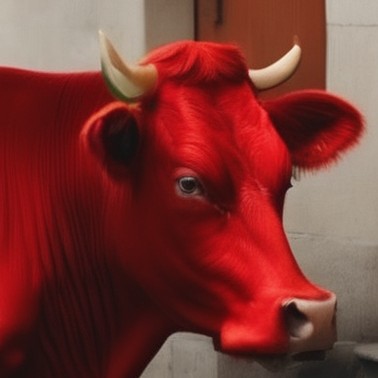}  \end{tabular}& 
\begin{tabular}{@{}c@{}}\includegraphics[width=2.7cm]{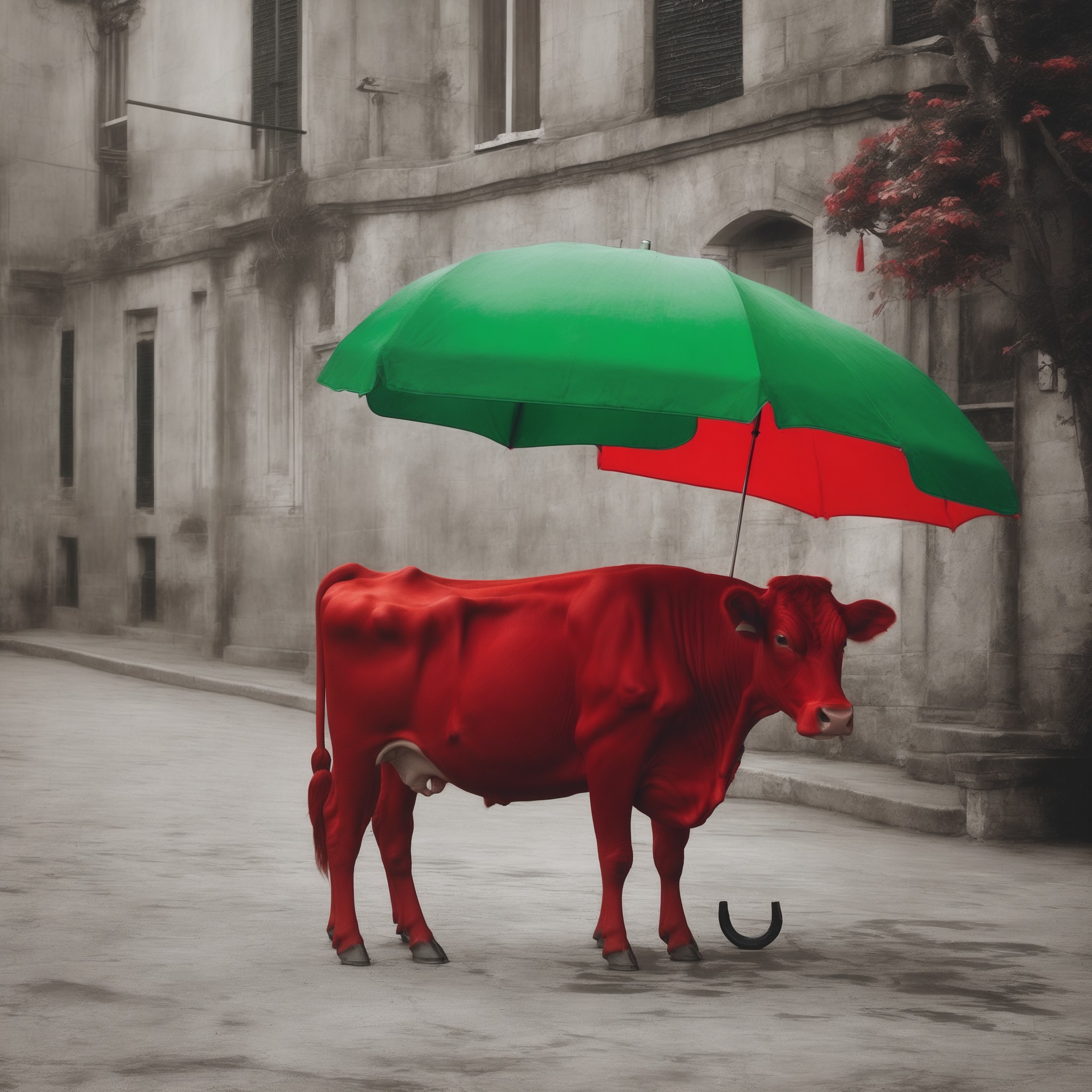}\\ \includegraphics[width=2.7cm]{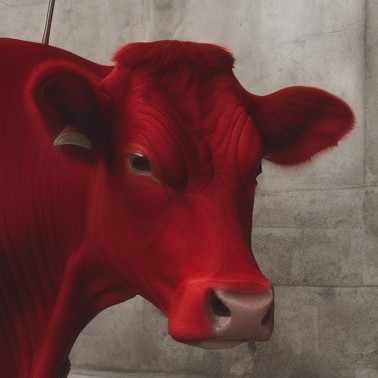}\end{tabular}& 
\begin{tabular}{@{}c@{}}\includegraphics[width=2.7cm]{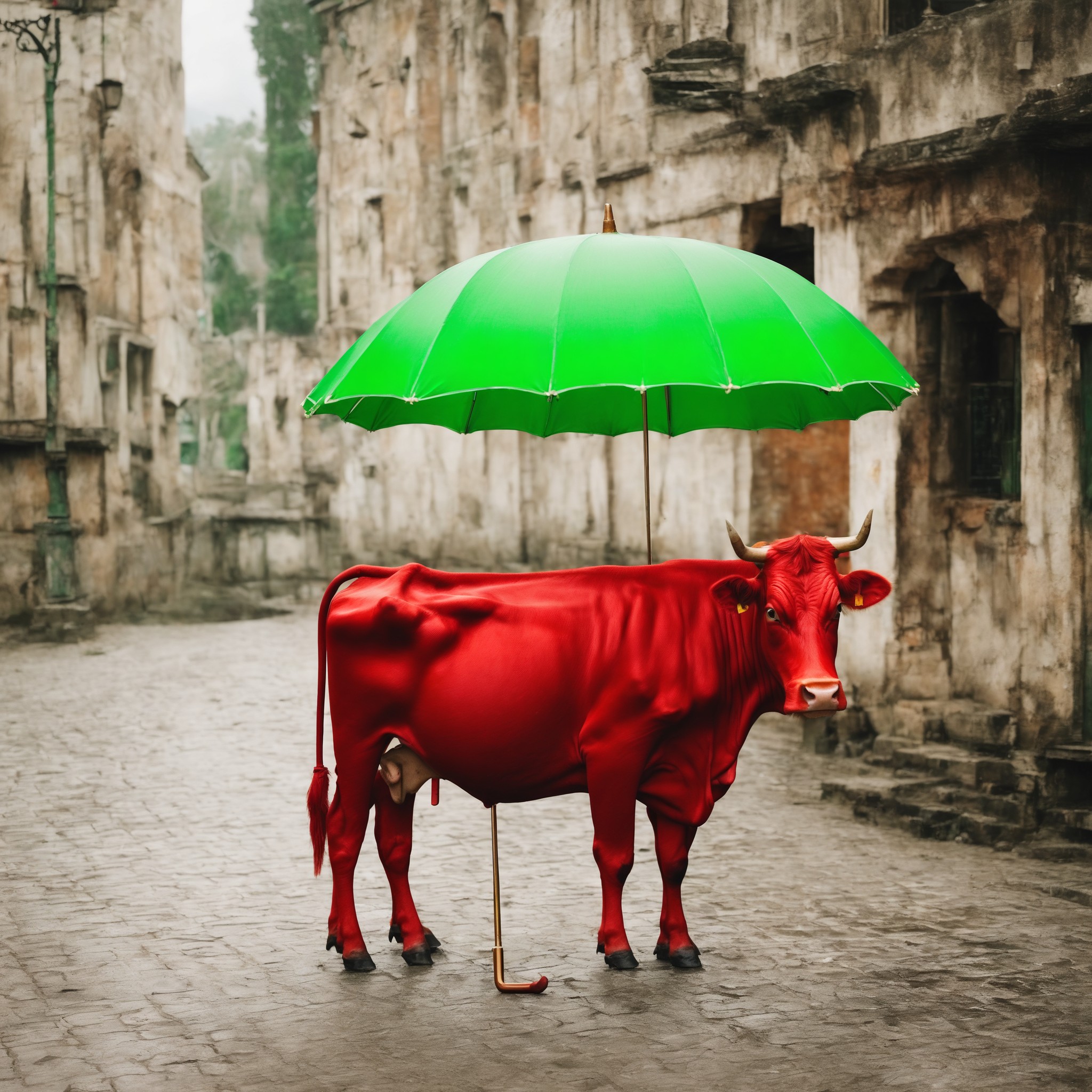}\\ \includegraphics[width=2.7cm]{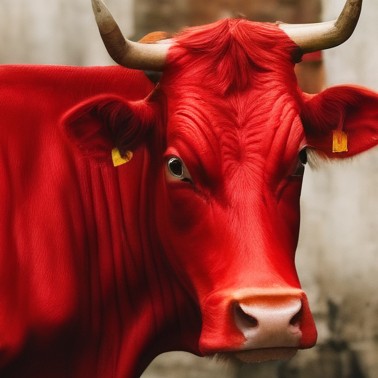}  \end{tabular}\\ 
\bottomrule
\end{tabular*}
\caption{More comparison between different progressive approaches on SDXL~{\cite{podellsdxl}}, where the top row represents the original $2048^2$ output images and bottom row represents cropped results.}
\label{fig:comparison:SDXL:appe}
\end{figure*}

Fig.~\ref{fig:comparison:SDXL:appe} presents a quantitative evaluation contrasting various approaches with our proposed {\DynaFlow} framework. We systematically compare SDXL~\cite{podellsdxl}, Self-Cascade~\cite{selfcascade:guo2024make}, and MegaFusion~\cite{wu2024megafusion} against {\DynaFlow} across multiple prompts. The results demonstrate that {\DynaFlow} produces outputs with quality comparable to Self-Cascade while offering significant advantages in inference speed and universal applicability. In contrast, MegaFusion exhibits consistent blurriness across all generated samples, highlighting the limitations of pixel-space transformation methods.
The superior performance of our approach stems from effective latent space manipulations that preserve semantic structure and fine details during the scaling process. This enables {\DynaFlow} to maintain perceptual quality while achieving computational efficiency that makes it practical for real-world applications requiring both high-quality outputs and responsive generation times.
\begin{algorithm*}
\caption{LSSGen: Latent Space Scaling Generation}
\label{alg:lssgen_manual_comments}
\begin{algorithmic}[1]
    \State \textbf{Input:}
    \State \quad \texttt{min\_resolution} \quad // \textit{Initial generation resolution}
    \State \quad \texttt{target\_resolution} \quad // \textit{Final desired resolution}
    \State \quad \texttt{base\_resolution} \quad // \textit{Reference resolution for step calculation}
    \State \quad \texttt{base\_steps} \quad // \textit{Number of diffusion steps for the base resolution}
    \State \quad \texttt{init\_noise\_level} \quad // \textit{Initial noise factor $\sigma_{init}$}
    \State \quad \texttt{shorten\_steps} \quad // \textit{Boolean to enable step reduction}

    \vspace{5pt}

    \State \textbf{Initialize:}
    \State \texttt{stages} $\leftarrow$ Define progressive scaling stages from \texttt{min\_resolution} to \texttt{target\_resolution}
    \State \texttt{latents} $\leftarrow$ \texttt{RandomNoiseGenerator}(\texttt{min\_resolution})

    \vspace{5pt}

    \For{\texttt{stage\_res} in \texttt{stages}}
        \If{\texttt{stage\_res} $>$ \texttt{min\_resolution}}
            \State \texttt{upsampled\_latents} $\leftarrow$ \texttt{Upsampler}(\texttt{latents})
            \State \texttt{noise} $\leftarrow$ \texttt{RandomNoise}(\texttt{stage\_res})
            \State \texttt{latents} $\leftarrow$ \texttt{upsampled\_latents} * (1 - \texttt{init\_noise\_level}) + \texttt{noise} * \texttt{init\_noise\_level}
        \EndIf
        
        \vspace{5pt}
        
        \State  // \textit{Calculate steps for the current stage}
        \If{\texttt{shorten\_steps} and \texttt{stage\_res} $<$ \texttt{base\_resolution}}
            \State \texttt{steps} $\leftarrow$ \texttt{base\_steps} / int(\texttt{base\_resolution} / \texttt{stage\_res})
        \Else
            \State \texttt{steps} $\leftarrow$ \texttt{base\_steps}
        \EndIf
        
        \vspace{5pt}
        
        \State \texttt{latents} $\leftarrow$ \texttt{DiffusionPipeline}(\texttt{latents}, \texttt{steps}, \texttt{stage\_res})
    \EndFor

    \vspace{5pt}

    \State \texttt{images} $\leftarrow$ \texttt{VAE\_Decode}(\texttt{latents})
    \State \textbf{return} \texttt{images}

\end{algorithmic}
\end{algorithm*}
\subsection{Details of Timesteps in LSSGen}
\label{subsec:appendix:timestep}
This section provides a detailed description of the input parameters for our proposed {\LSSGen}, as presented in Algorithm~\ref{alg:lssgen_manual_comments}. These parameters allow for precise control over the progressive generation process, enabling users to balance computational efficiency and final image quality.
Each parameter is defined as follows:

\noindent\texttt{min\_resolution}: This integer value specifies the initial, lowest resolution at which the generative process begins. The first stage of the pipeline synthesizes a latent tensor at this resolution from pure noise. We use 512 in FLUX.1-dev.

\noindent\texttt{target\_resolution}: Defines the final, desired resolution of the output image. The \LSSGen framework progressively upscales the latent representation through multiple stages until this target resolution is reached.

\noindent\texttt{base\_resolution}: Serves as a reference resolution for the dynamic step calculation. When the shorten\_steps flag is enabled, any generation stage operating at a resolution lower than base\_resolution will use a proportionally reduced number of denoising steps.

\noindent\texttt{base\_steps}: The baseline number of denoising steps performed by the diffusion pipeline (e.g., FLUX.1-dev and SDXL are 50 steps) for any stage operating at or above the base\_resolution.

\noindent\texttt{init\_noise\_level}: This floating-point value corresponds to the initial noise coefficient ({\InitSigma}) discussed in our methodology (Section 4.1). It governs the ratio between the signal from the upsampled latent and the stochastic noise injected at the beginning of a new stage. Based on our analysis, this is typically set to 0.75 for optimal quality. The parameter can be lower if efficiency is desired.

\noindent\texttt{shorten\_steps}: A boolean flag that enables (True) or disables (False) the intermediate step reduction strategy. When enabled, this optimization accelerates the initial, lower-resolution stages by reducing their denoising step count, significantly improving overall inference speed with minimal impact on quality.

Together, these parameters provide fine-grained control over the speed-quality trade-off within the LSSGen framework, making it adaptable to various hardware constraints and use cases.

\begin{figure*}[t]
\centering
\setlength{\tabcolsep}{0pt}
\begin{tabular*}{\textwidth}{@{\extracolsep{\fill}}p{5.6cm} p{5.6cm} p{5.6cm}}
\toprule
\includegraphics[width=5.6cm]{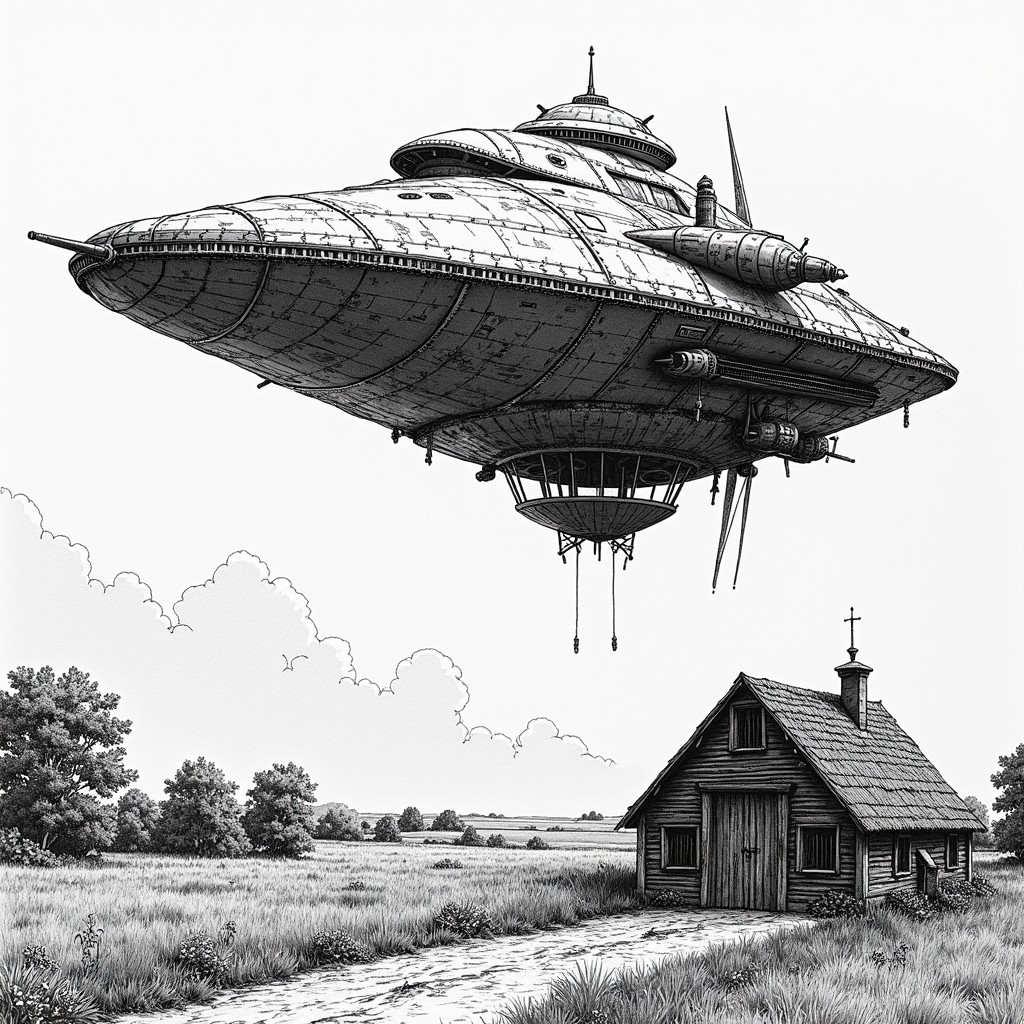}& 
\includegraphics[width=5.6cm]{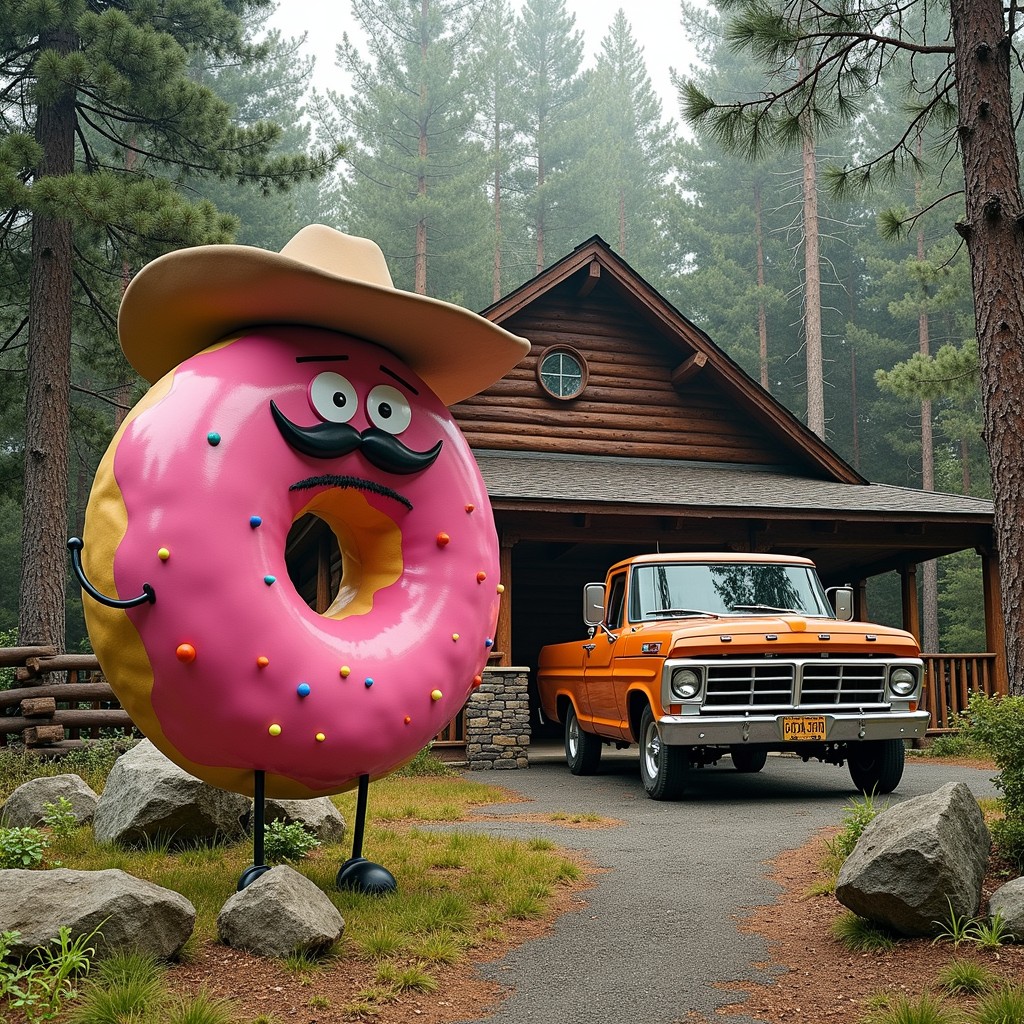}& 
\includegraphics[width=5.6cm]{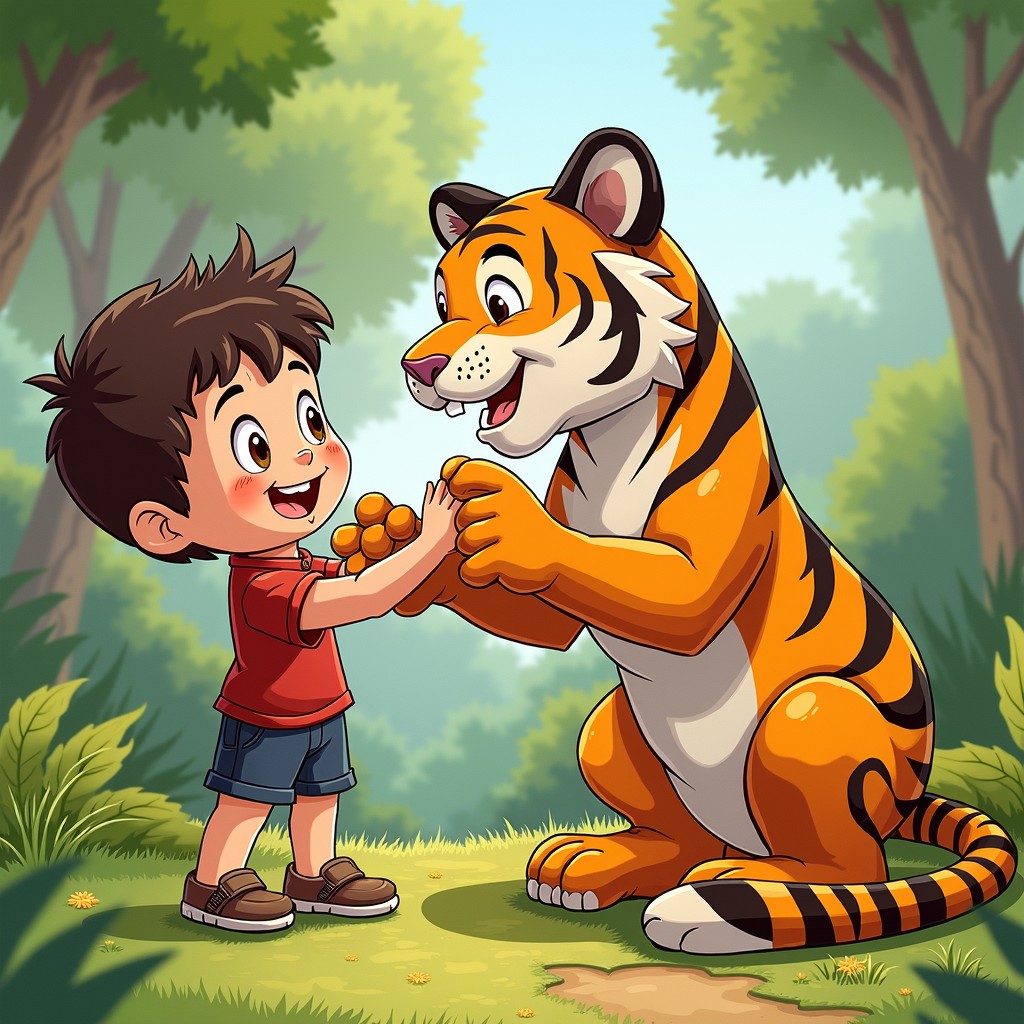}\\ 
detailed pen and ink drawing of a massive complex alien space ship above a farm in the middle of nowhere & an anthopomorphic pink donut with a mustache and cowboy hat standing by a log cabin in a forest with an old 1970s orange truck in the driveway & a cartoon of aboy playing with a tiger \\
\midrule
\includegraphics[width=5.6cm]{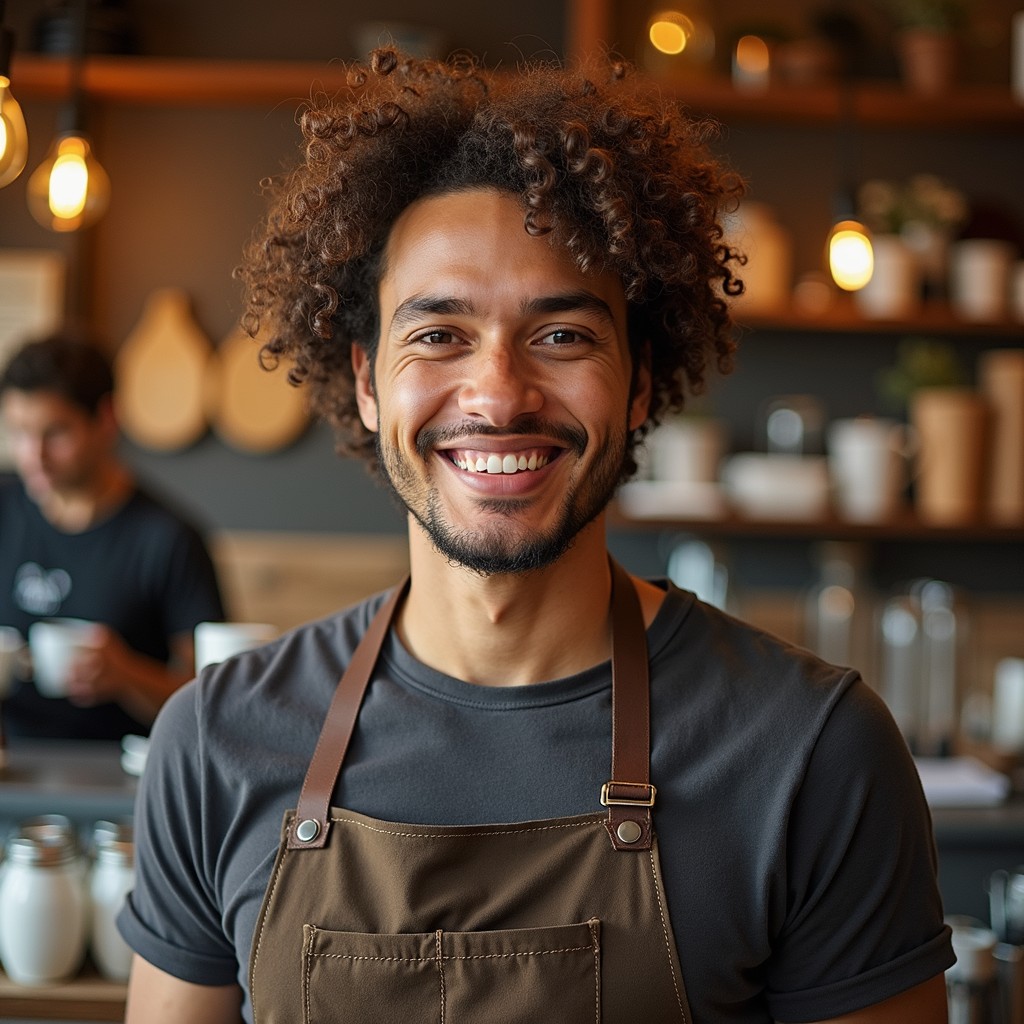}& 
\includegraphics[width=5.6cm]{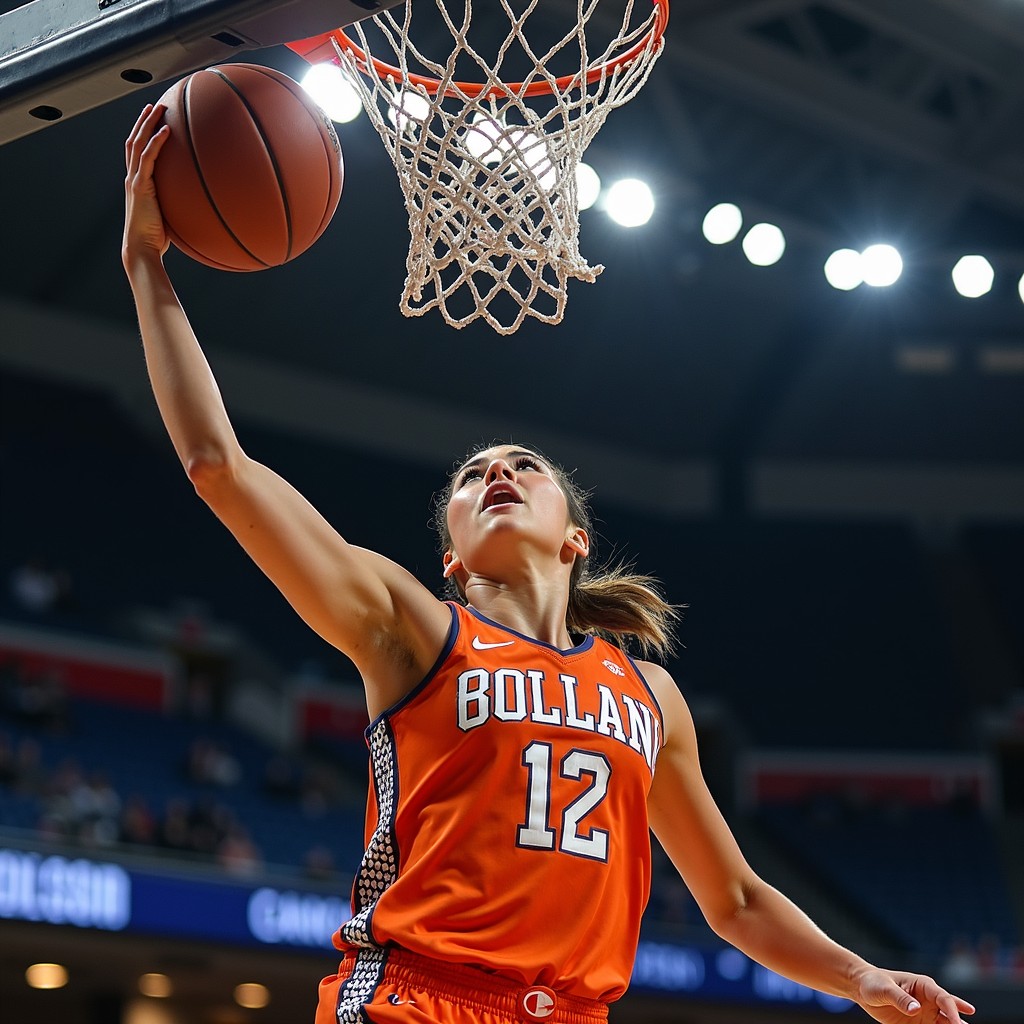}& 
\includegraphics[width=5.6cm]{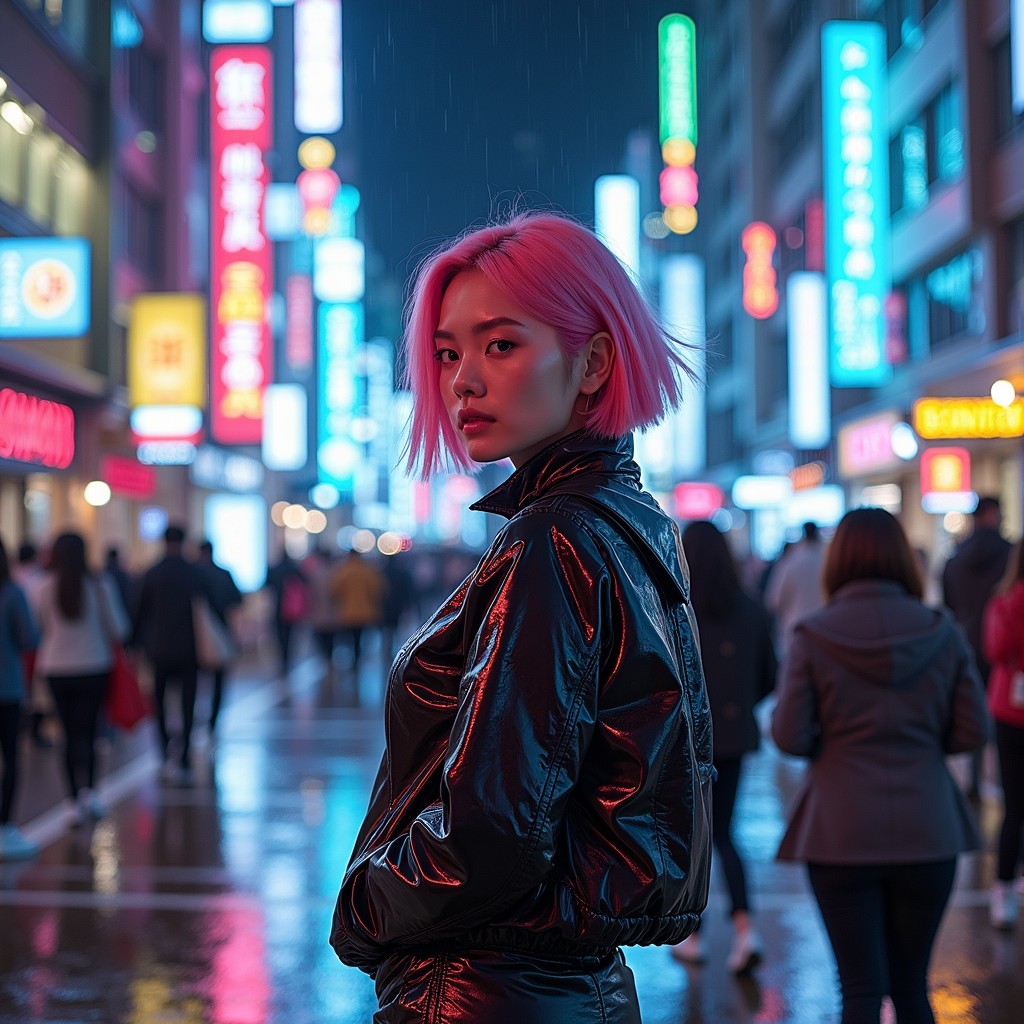}\\ 
Medium shot of a friendly male barista with curly hair and an apron, smiling in a cozy, warm-lit caf & Action shot of a female basketball player mid-dunk, with an intense expression in a brightly lit arena & Full-body shot of a woman with pink hair in a neon-lit, rainy Tokyo street at night, wearing a futuristic jacket \\
\midrule
\includegraphics[width=5.6cm]{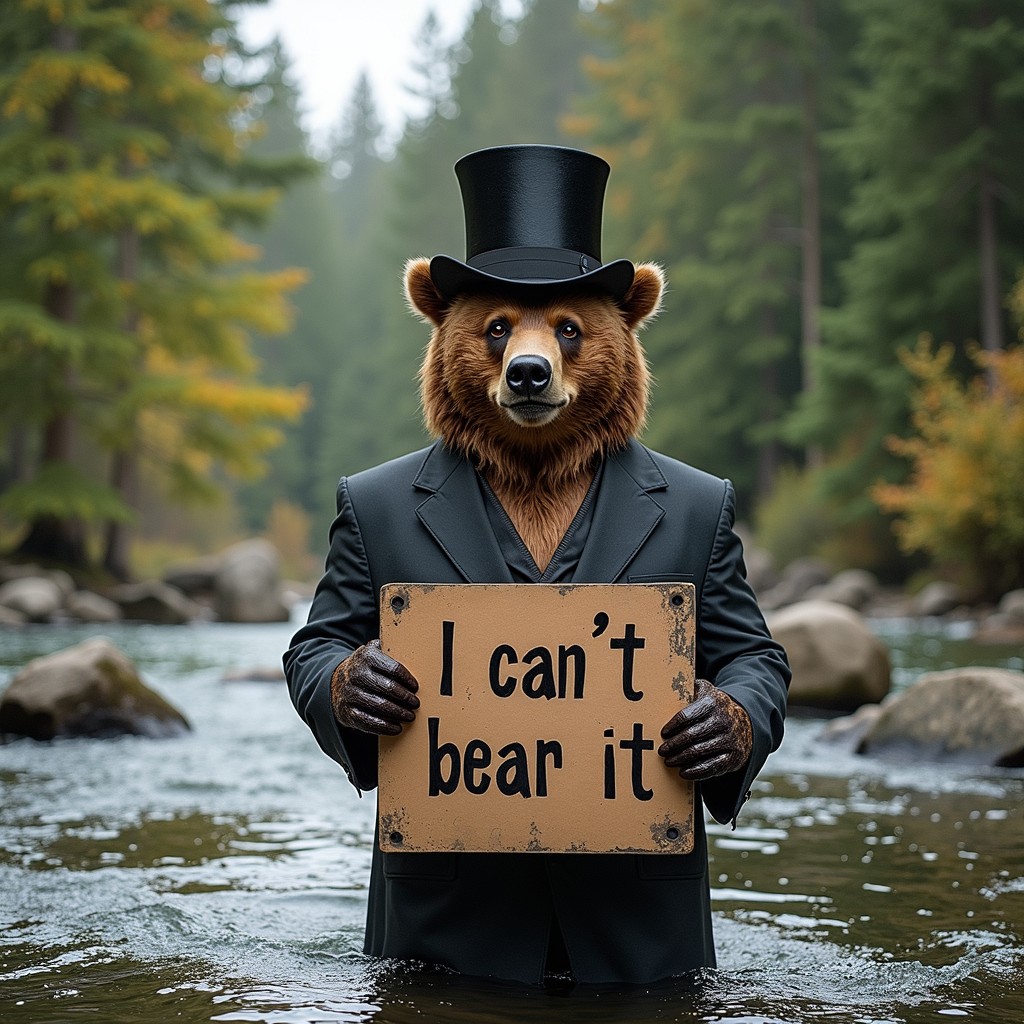}& 
\includegraphics[width=5.6cm]{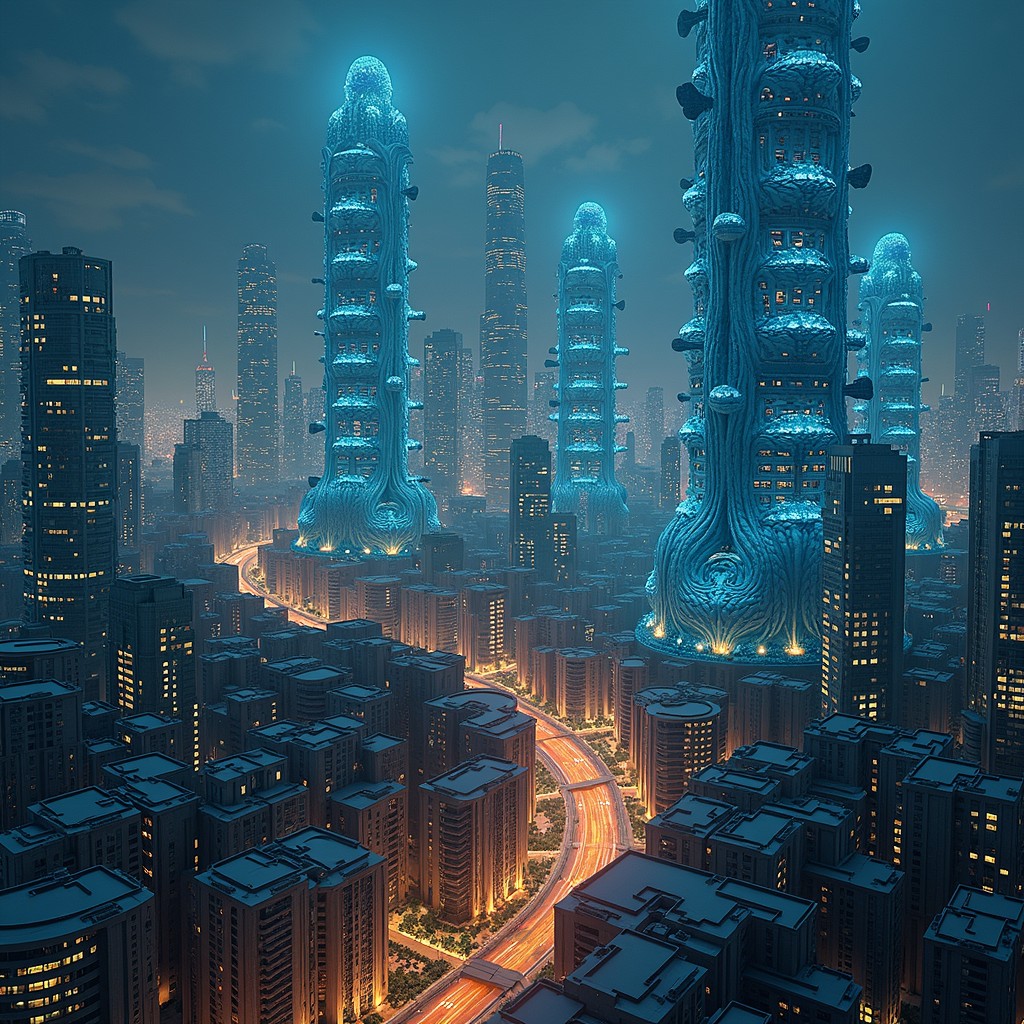}& 
\includegraphics[width=5.6cm]{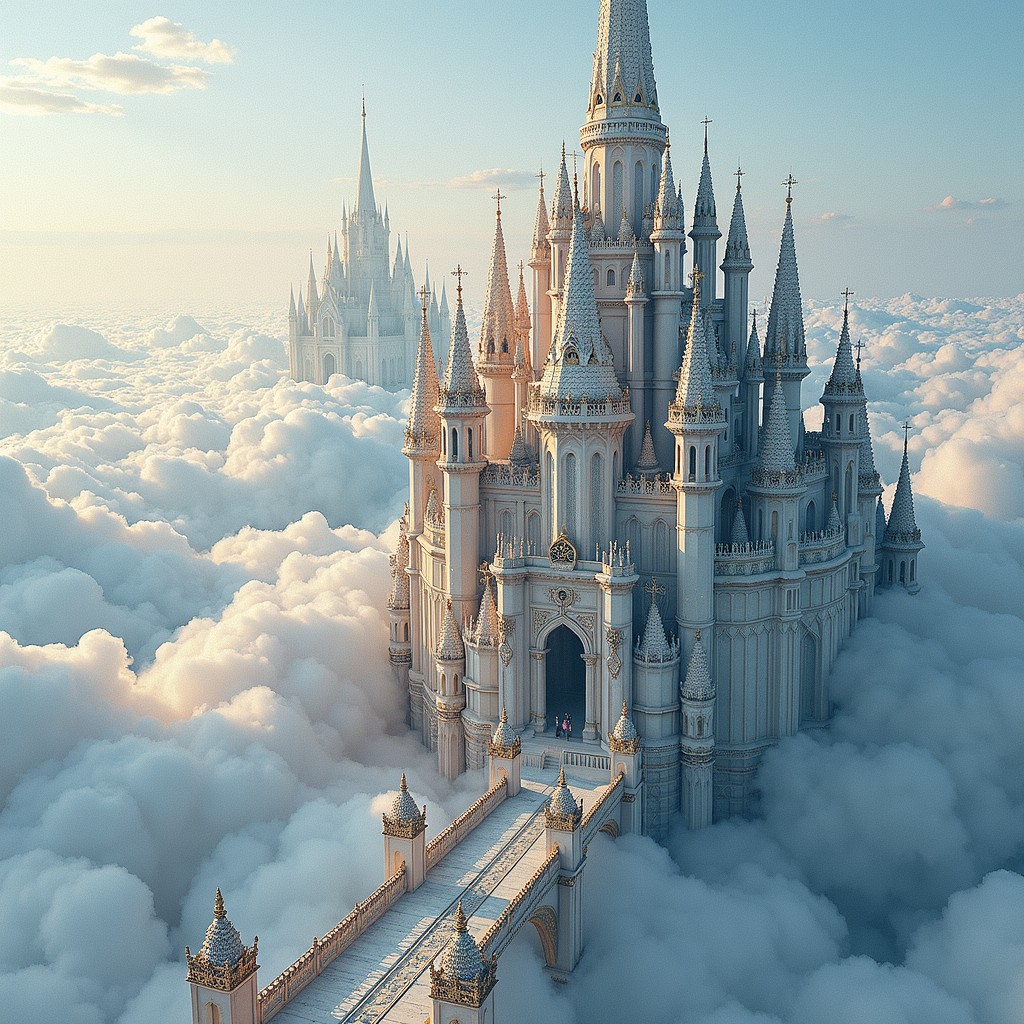}\\
photo of a bear wearing a suit and tophat in a river in the middle of a forest holding a sign that says I cant bear it & A futuristic city at night, where the skyscrapers are built from living, glowing organic material, & An intricate city of marble towers and ornate bridges built upon a dense layer of clouds, \\
\bottomrule
\end{tabular*}
\vspace{-6pt}
\caption{More generated samples of {\LSSGen} on FLUX.1-dev~{\cite{blackforest2024flux}} at $1024^2$ resolution.}
\label{fig:comparison:LSS-FLUX-dev-1024:appe}
\end{figure*}
\begin{figure*}[t]
\centering
\footnotesize
\setlength{\tabcolsep}{0pt}
\begin{tabular*}{\textwidth}{@{\extracolsep{\fill}}p{5.6cm} p{5.6cm} p{5.6cm}}
\toprule
\includegraphics[width=5.6cm]{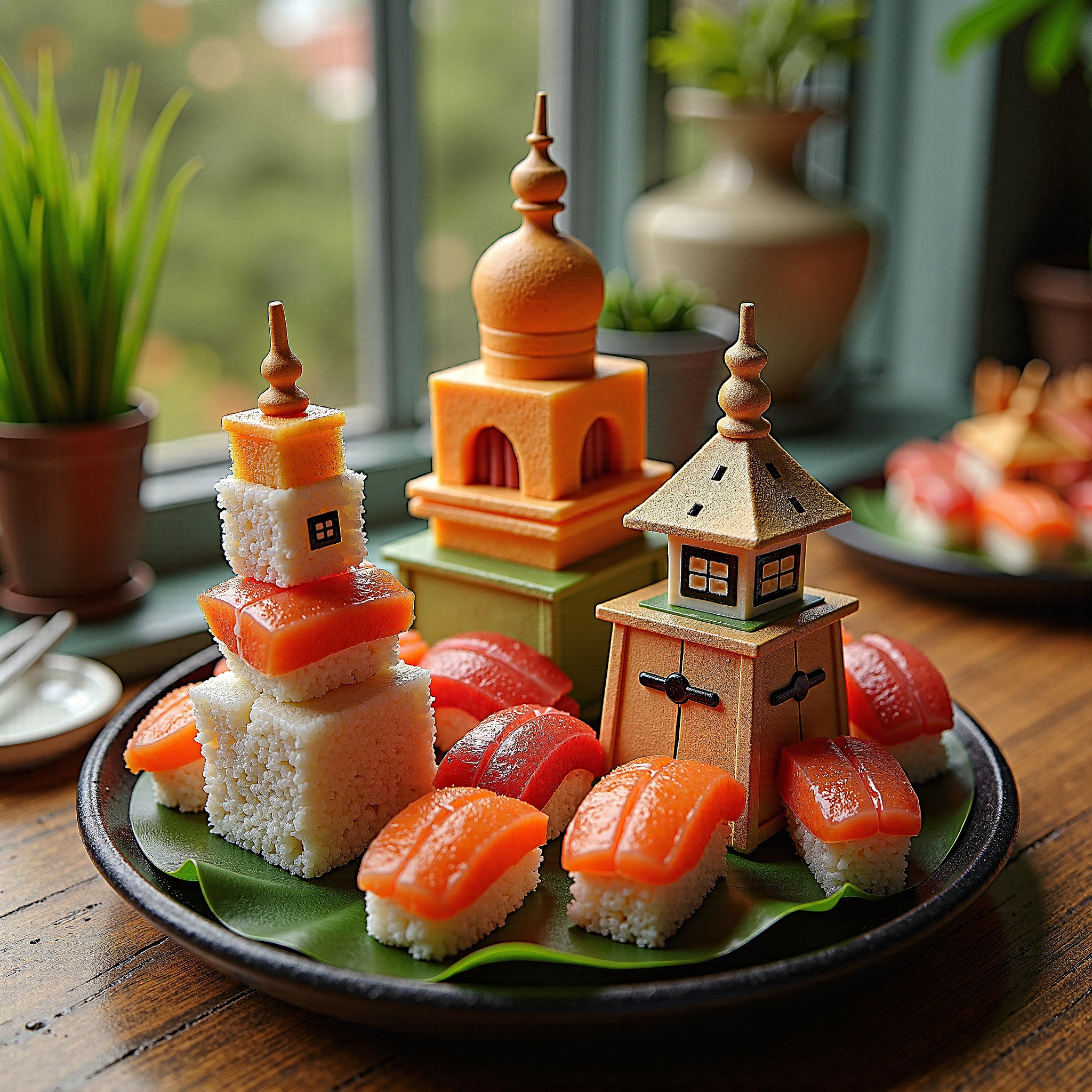}& 
\includegraphics[width=5.6cm]{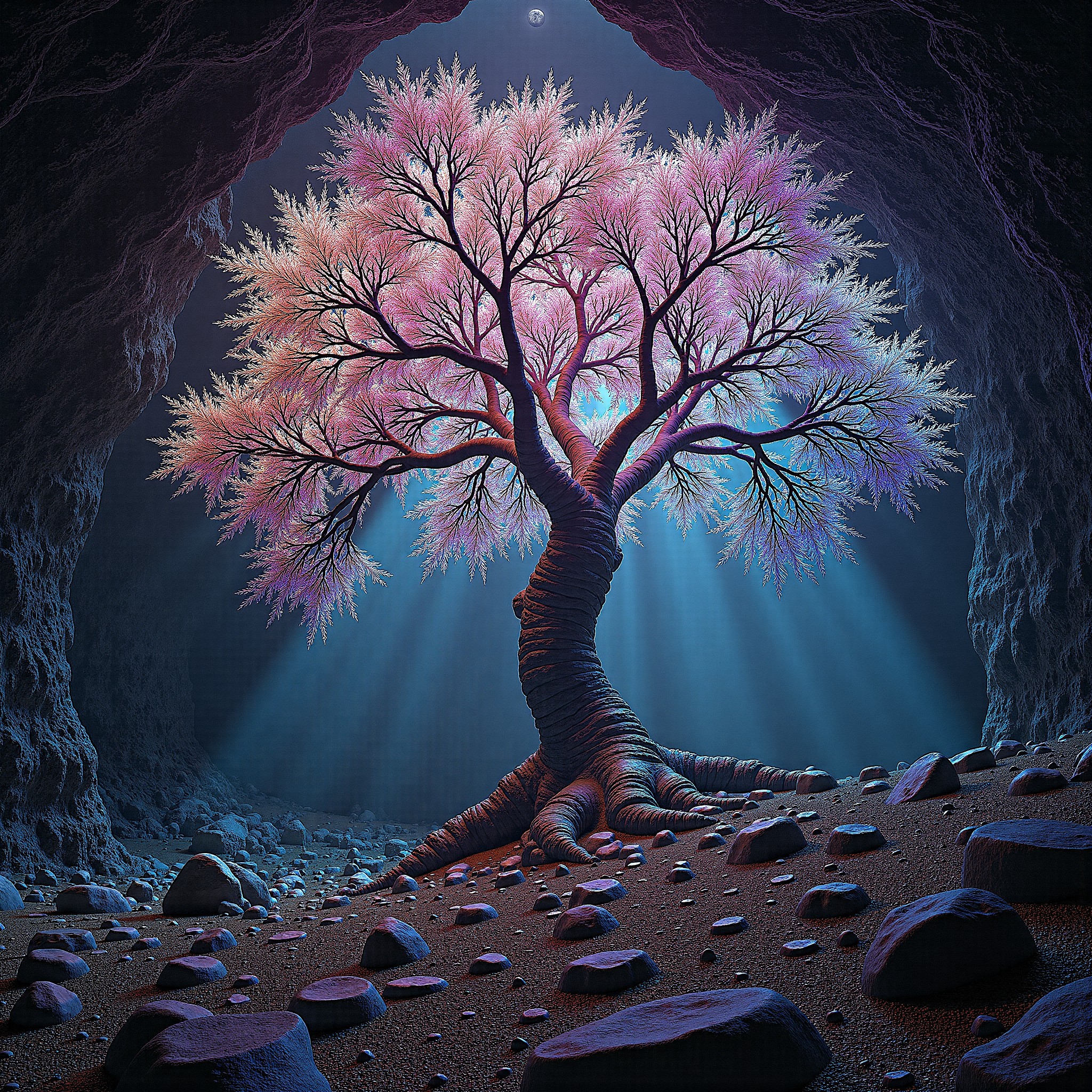}& 
\includegraphics[width=5.6cm]{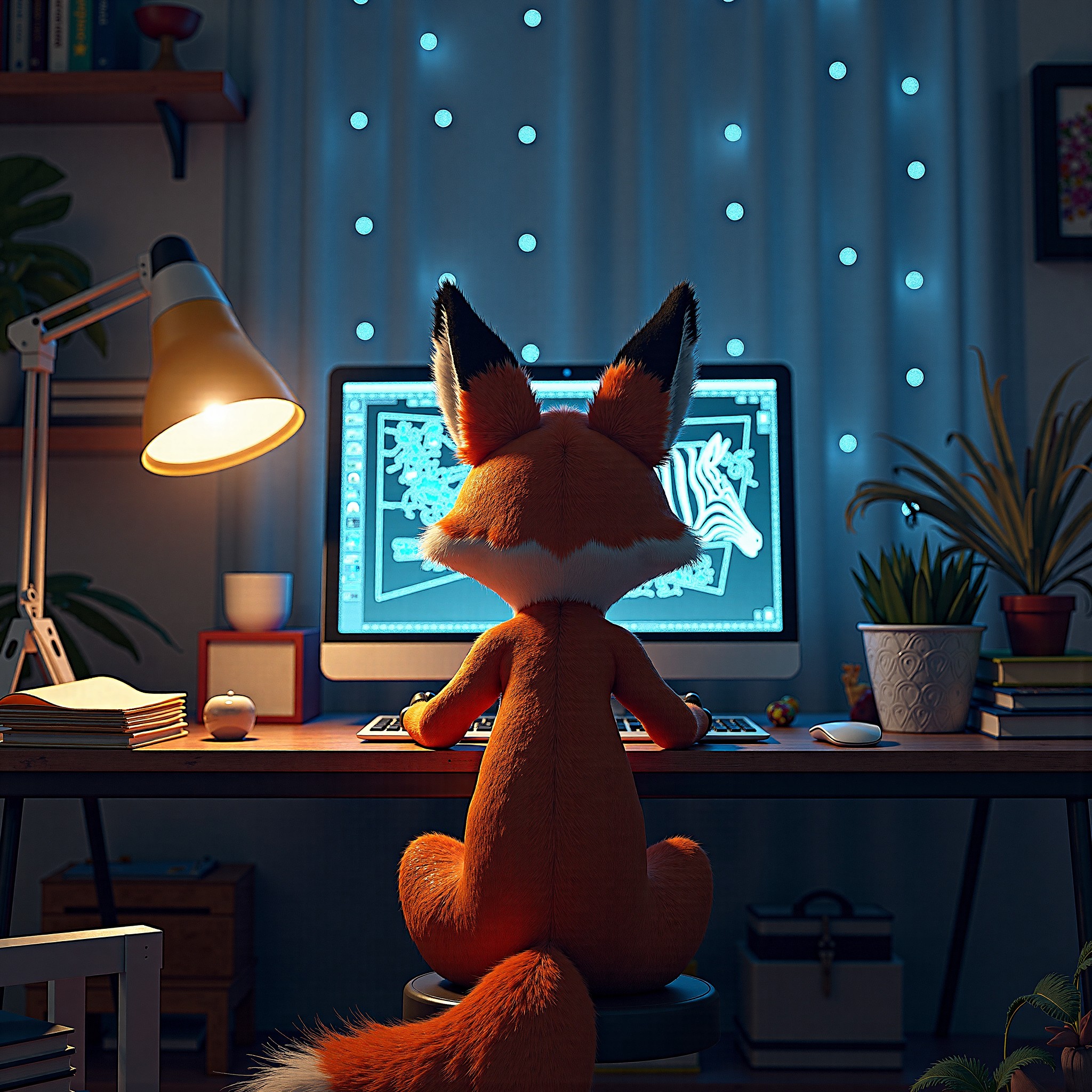}\\ 
tilt shift aerial photo of a cute city made of sushi on a wooden table in the evening & dark high contrast render of a psychedelic tree of life illuminating dust in a mystical cave & fox sitting in front of a computer in a messy room at night. On the screen is a 3d modeling program with a line render of a zebra \\
\midrule
\includegraphics[width=5.6cm]{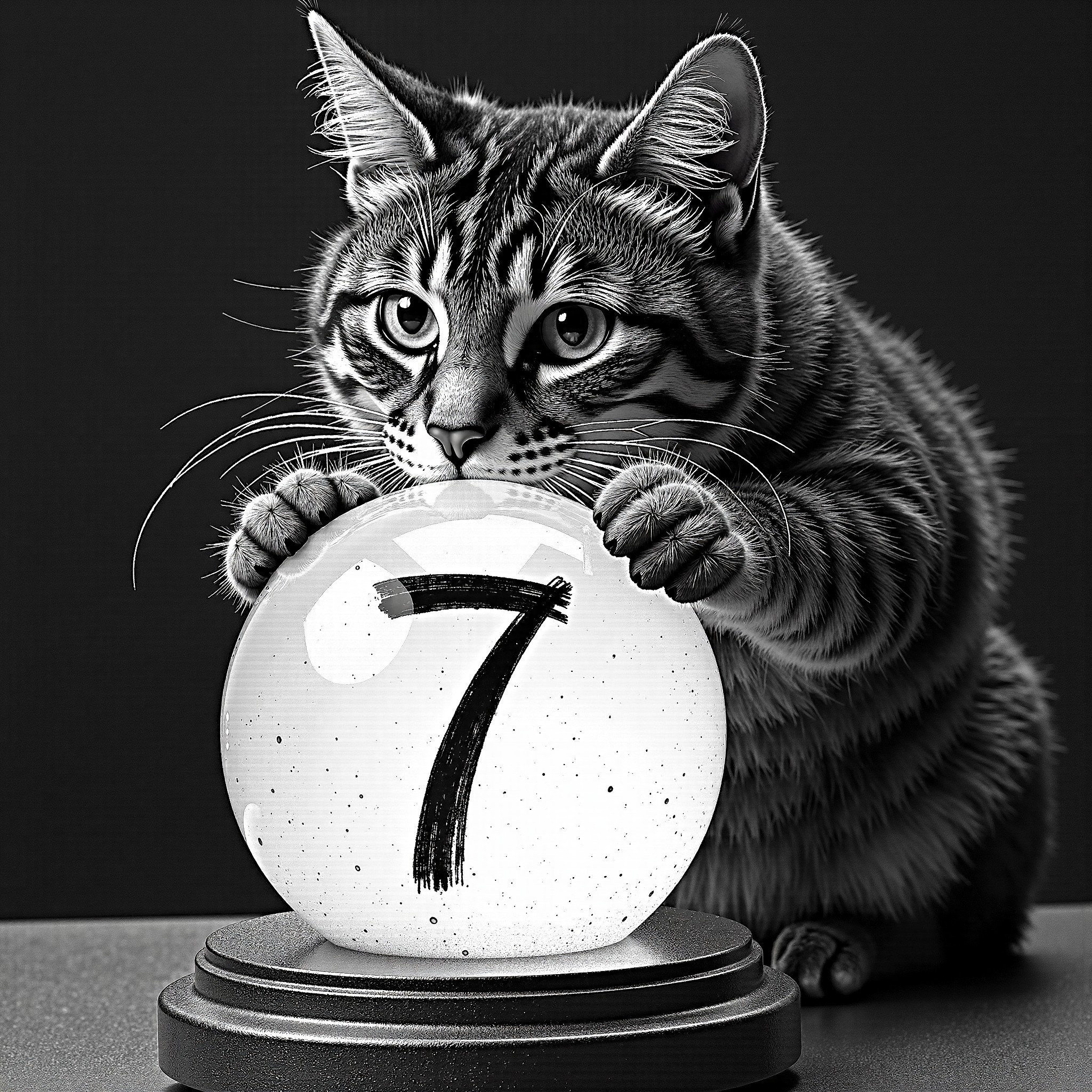}& 
\includegraphics[width=5.6cm]{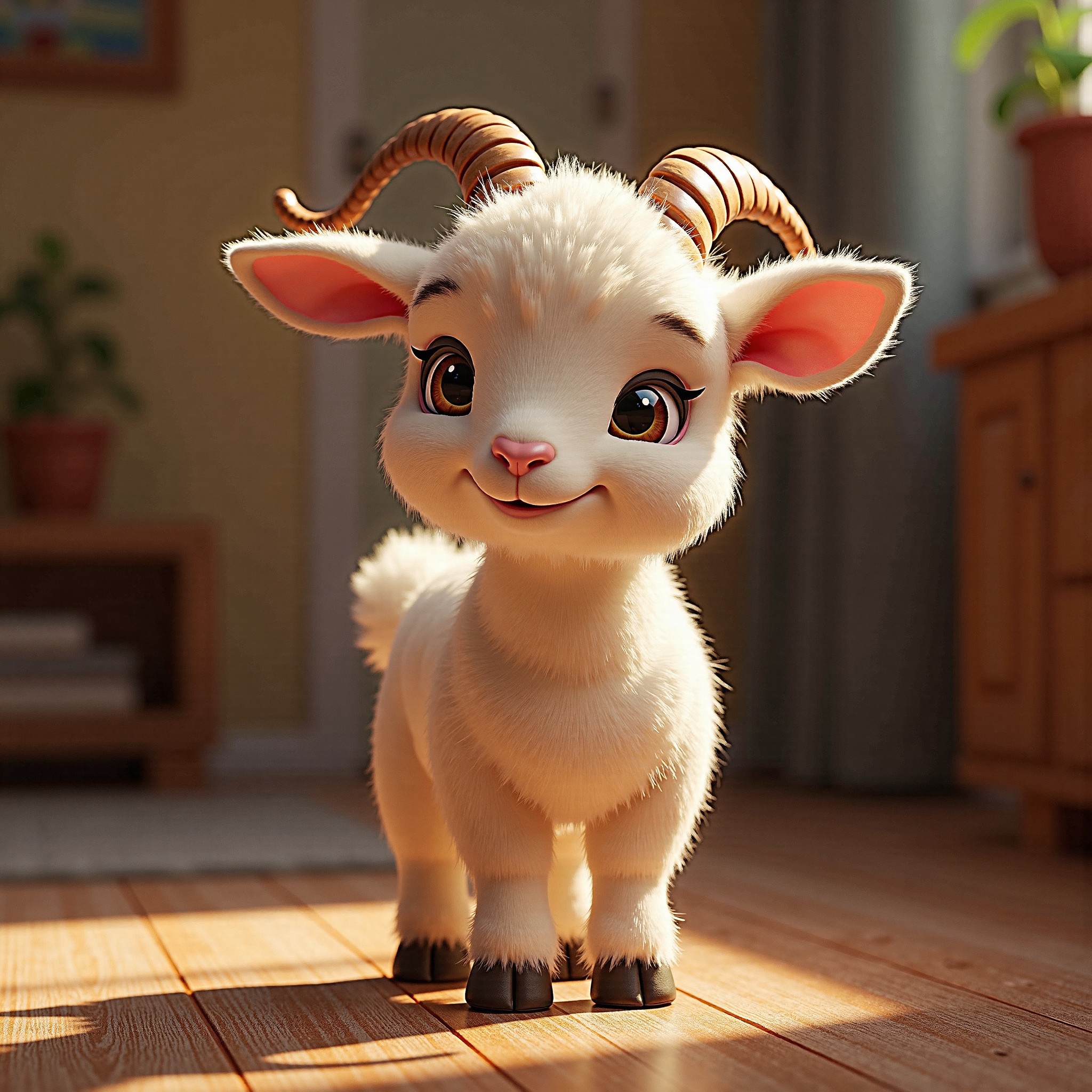}& 
\includegraphics[width=5.6cm]{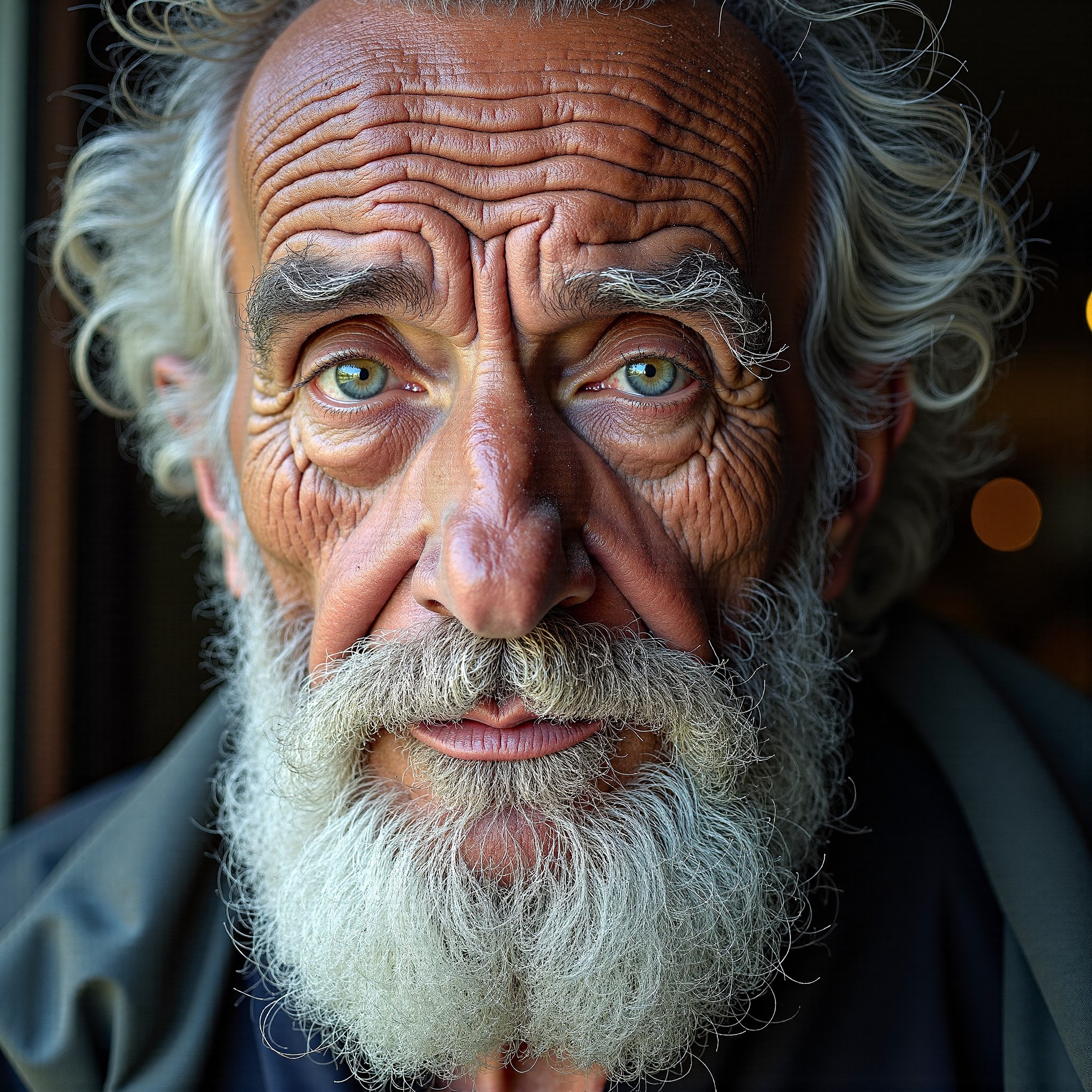}\\ 
cat patting a crystal ball with the number 7 written it in black marker & Cute adorable little goat, unreal engine, cozy interior lighting, art station, detailed digital painting, cinematic, octane rendering & Close-up portrait of an old fisherman with a kind, wrinkled face and white beard, soft window lighting \\
\midrule
\includegraphics[width=5.6cm]{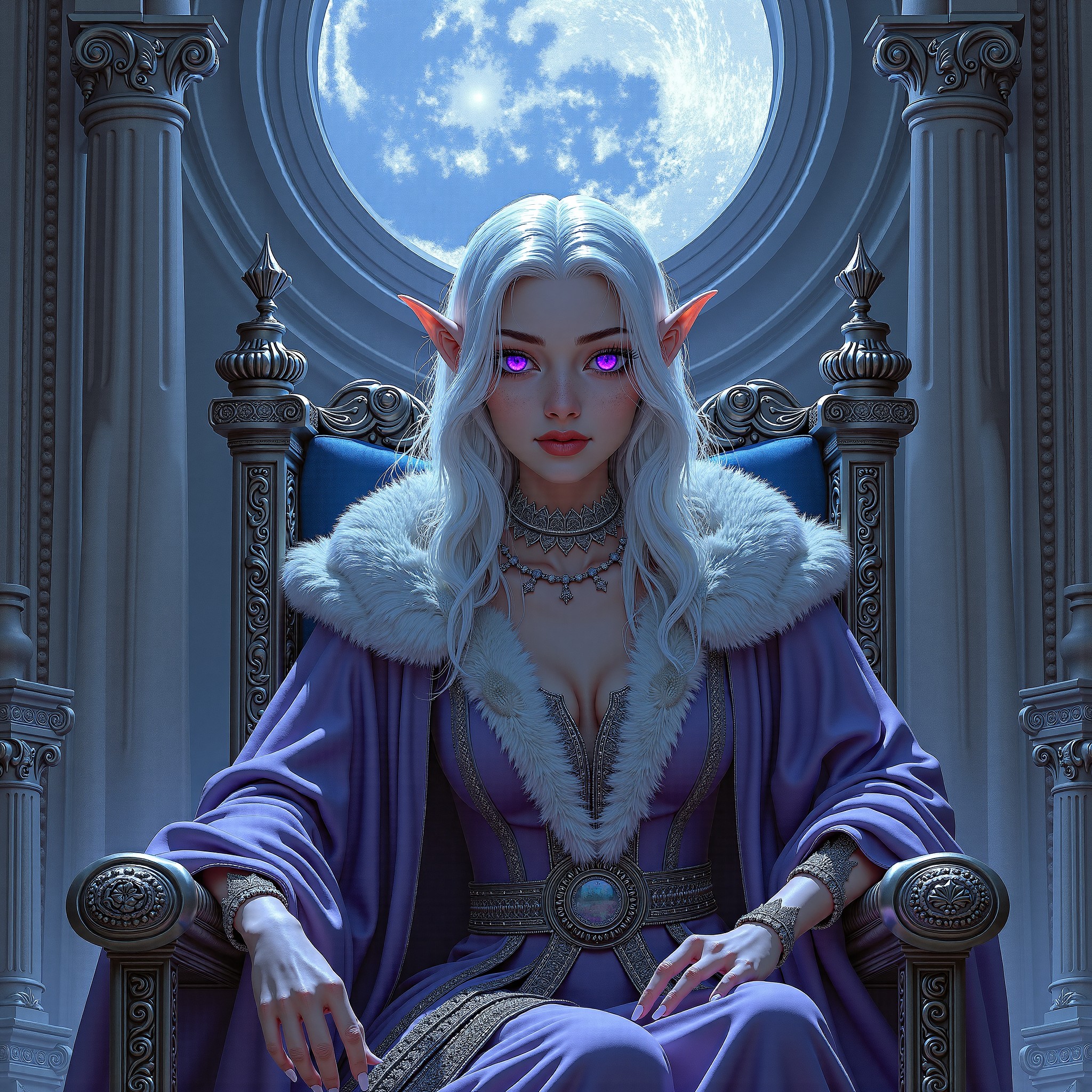}& 
\includegraphics[width=5.6cm]{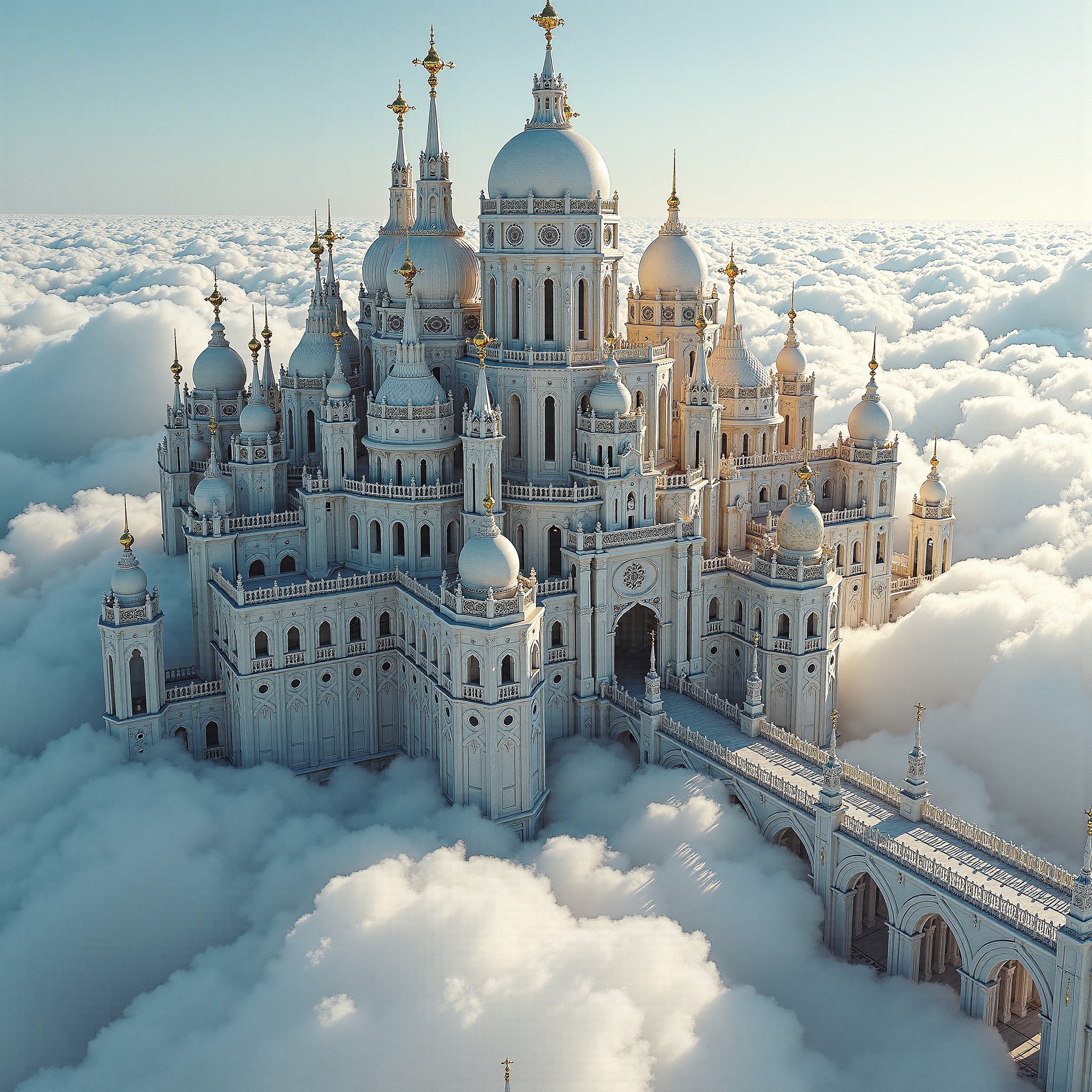}& 
\includegraphics[width=5.6cm]{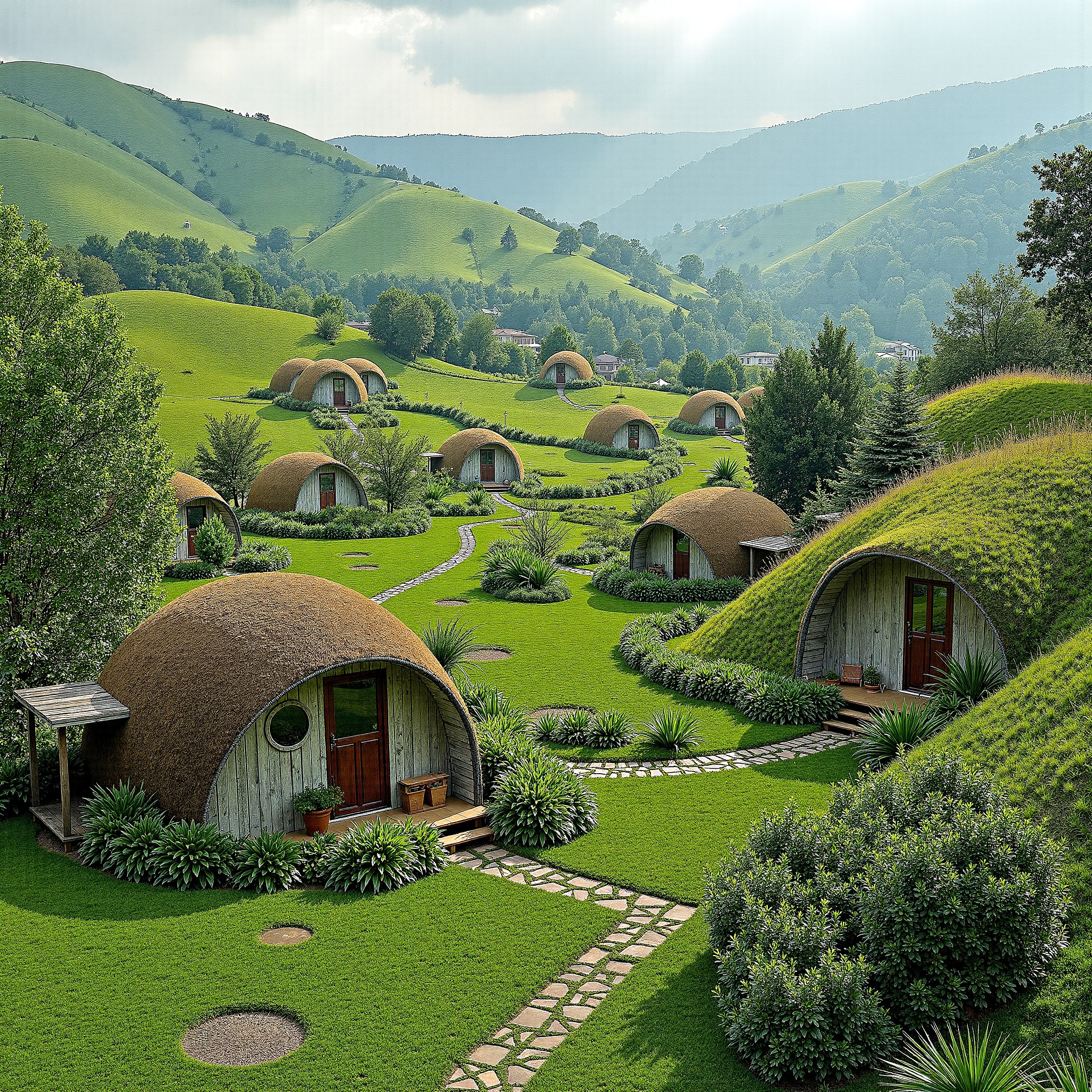}\\
Fantasy portrait of a non-binary elf with silver hair and violet eyes on a throne, in a grand hall with moonlight & An intricate city of marble towers and ornate bridges built upon a dense layer of clouds, & A cozy village of round-doored homes built into lush green hills \\
\bottomrule
\end{tabular*}
\vspace{-1pt}
\caption{More generated samples of {\LSSGen} on FLUX.1-dev~{\cite{blackforest2024flux}} at $2048^2$ resolution.}
\label{fig:comparison:LSS-FLUX-dev-2048:appe}
\end{figure*}